\documentclass[10pt,twocolumn,letterpaper]{article}

\usepackage{zhanggroup}
\usepackage{tikz}
\usepackage{xspace} 
\usepackage{mathrsfs}
\usepackage{bm}
\usepackage{subcaption}
\usepackage{booktabs}
\usepackage{multirow}
\usepackage{diagbox}
\usepackage{graphicx}
\usepackage{url}
\usepackage{caption}
\usepackage{dsfont}
\usepackage{colortbl}
\usepackage{array}
\usepackage{makecell}
\usepackage{algorithm2e}
\usepackage{amsmath,amssymb,amsfonts}
\usepackage{hyperref}
\hypersetup{
  colorlinks,
  linkcolor={blue!70!green},
  citecolor={green!70!blue},
  urlcolor={orange!70!red}
}
\usepackage{tcolorbox}
\usepackage[absolute]{textpos}
\newcommand{\mybox}[1]{
\begin{tcolorbox}[boxsep=1pt,left=2pt,right=2pt,top=0.5 pt,bottom=0.5pt, frame empty]
#1
\end{tcolorbox}
}
\newcommand{\mypara}[1]{\noindent{\bf {#1}.}\xspace}
\newcommand{\red}[1]{\textcolor{red}{#1}}
\def\input{{\texttt{[INPUT]}}}
\def\mask{{\texttt{[MASK]}}}
\definecolor{babypink}{rgb}{0.96, 0.76, 0.76}
\definecolor{darktangerine}{rgb}{1.0, 0.66, 0.07}
\definecolor{emerald}{rgb}{0.31, 0.78, 0.47}

\begin{document}

\begin{textblock}{13}(1.5,1)
\centering
To Appear in the 45th IEEE Symposium on Security and Privacy, May 20-23, 2024.
\end{textblock}

\title{\Large \bf You Only Prompt Once: On the Capabilities of Prompt Learning on Large Language Models to Tackle Toxic Content}

\date{}

\author{
Xinlei He\textsuperscript{1}\ \ \
Savvas Zannettou\textsuperscript{2}\ \ \
Yun Shen\textsuperscript{3}\ \ \
Yang Zhang\textsuperscript{1}
\\
\\
\textsuperscript{1}{CISPA Helmholtz Center for Information Security}\ \ \ 
\textsuperscript{2}{Delft University of Technology}\ \ \ 
\textsuperscript{3}{NetApp}
}

\maketitle

\begin{abstract}

The spread of toxic content online is an important problem that has adverse effects on user experience online and in our society at large.
Motivated by the importance and impact of the problem, research focuses on developing solutions to detect toxic content, usually leveraging machine learning (ML) models trained on human-annotated datasets.
While these efforts are important, these models usually do not generalize well and they can not cope with new trends (e.g., the emergence of new toxic terms).
Currently, we are witnessing a shift in the approach to tackling societal issues online, particularly leveraging large language models (LLMs) like GPT-3 or T5 that are trained on vast corpora and have strong generalizability.
In this work, we investigate how we can use LLMs and prompt learning to tackle the problem of toxic content, particularly focusing on three tasks; 1) Toxicity Classification, 2) Toxic Span Detection, and 3) Detoxification.
We perform an extensive evaluation over five model architectures and eight datasets demonstrating that LLMs with prompt learning can achieve similar or even better performance compared to models trained on these specific tasks.
We find that prompt learning achieves around 10\% improvement in the toxicity classification task compared to the baselines, while for the toxic span detection task we find better performance to the best baseline (0.643 vs. 0.640 in terms of $F_1$-score).
Finally, for the detoxification task, we find that prompt learning can successfully reduce the average toxicity score (from 0.775 to 0.213) while preserving semantic meaning.\footnote{Our code is available at \url{https://github.com/xinleihe/toxic-prompt}.}

\end{abstract}

\smallskip
\noindent{\bf\red{Disclaimer.
This paper contains uncensored toxic content that might be offensive or disturbing to the readers.}}

\section{Introduction}

In online platforms, toxic content can be defined as rude, disrespectful, or unreasonable content that may result in users leaving the conversation~\cite{BDSTV19}.
It has been a long-standing problem affecting our society~\cite{APP19,DWMW17,SMCBW16,ENNVB18}.
To tackle this problem, researchers and companies leverage large-scale labeled datasets to train powerful machine learning (ML) models for toxicity detection and mitigation~\cite{Perspective,DWMW17,ZEBNS20,ZKF18,ZRT18,OWBLM21}.

One major obstacle in the development of accurate and generalizable toxic content classifiers is the lack of a comprehensive labeled dataset that contains different types of toxic content.
This is mainly because the data collection and labeling process for the creation of such datasets is costly, which hinders the development of effective methods for detecting toxic content.
Also, previous work~\cite{APP19,ZEBNS20} has shown that the toxicity detection model trained on one dataset is less effective when applied to other datasets.
Moreover, due to the fast evolution of language (new phrases, words, style, etc.),
it is crucial to develop a toxicity detection mechanism that can quickly adapt to different circumstances.

With the success of pre-trained language models (LMs), a dominant way to adapt the model to downstream tasks is fine-tuning, where the whole model or part of the model is optimized to better fit the downstream tasks.
Recently, large language models (LLMs) like GPT-3~\cite{BMRSKDNSSAAHKHCRZWWHCSLGCCBMRSA20} and T5~\cite{RSRLNMZLL20} have shown promising performance in downstream tasks without updating at all the model's parameters by directly querying the model using natural language, an emerging paradigm called \emph{prompt learning}.
With the help of prompt learning, the LLM can generate an output that aims to solve a specific task, all with a natural language task instruction (e.g., using a prompt: ``Translate it from English to French'' for machine translation) and a few samples as the task input.
Besides the handcrafted fixed prompts, recent work~\cite{LAC21,LL21} shows that prompt tuning is an efficient way to achieve more promising performance on various tasks with restricted computational resources, limited datasets, and bounded time.
Concretely, instead of fine-tuning the LLM, prompt tuning freezes the LLM and only optimizes the prompt (e.g., the way that the prompt is written) in such a way that the LLM's performance is optimized for the specific task at hand.
Given that prompt learning is a promising way to use LLM for various tasks, here we aim to use prompt learning to tackle the problem of toxic content and assess how prompt learning-based approaches compare to state-of-the-art methods of tackling toxic content.

\mypara{Our Work}
In this work, we conduct the first systematic analysis focusing on how prompt learning can help tackle the problem of toxic content.
Concretely, we focus on three tasks, i.e., toxicity classification, toxic span detection, and detoxification (see \autoref{table:general_examples} for examples of these tasks).
Specifically, for the first task (toxicity classification), given a sentence, we first map its label into the word ``Yes'' or ``No'' and fine-tune the prompt to better guide the LLM to conduct the task.
For the second task (toxic span detection), with prompt tuning, given a sentence with toxic spans, we aim to first generate the sentence without the toxic spans, then subtract the original sentence with the generated sentence to obtain the spans.
Finally, for the third task (detoxification), we tune the prompt to rephrase the toxic sentence into a non-toxic version while preserving the semantic meaning.

Extensive evaluation of eight datasets and five model architectures shows that prompt tuning has comparable or even better performance than the baselines.
For instance, for the toxicity classification task, prompt tuning
gains more than 10\% $F_1$-score improvement on average (see \autoref{table:task1_performance_f1}).
For the toxic span detection task, our method achieves 0.643 $F_1$-score, which is better than the best result provided by SPAN-BERT (0.640), but with much less training time.
Regarding the detoxification task, we find that our method can successfully detoxify the text (e.g.,  the average toxicity score drops from 0.775 to 0.213 on ParaDetox) while preserving the semantic information to a large extent.
In general, one major advantage of prompt tuning is that it can adapt to different tasks with fewer training samples/steps.
For online services such as social media, these improvements and cost reductions are significant (given billions of posts per day).
This also fits the purpose of green AI~\cite{GreenAICloud,SDSE20} for making AI research more environmentally friendly and inclusive.

\begin{table}[t]
\centering
\caption{Examples of the three tasks we consider in our work.
Toxicity classification aims to classify whether the given sentence is toxic or not.
Toxic span detection aims to detect \colorbox{emerald}{the exact span} that makes the sentence toxic.
Detoxification aims to reduce the toxicity of the given sentence while preserving its semantic information.}
\scalebox{0.9}{
\begin{tabular}{p{3.6cm} p{3.7cm}}
\toprule
\textbf{Toxicity Classification} & \textbf{Answer} \\
\toprule
your reading comprehension is more fucked up than a football bat.  & Toxic \\
\toprule
\textbf{Toxic Span Detection} & \textbf{Answer} \\
\toprule
keep hiring imbeciles like this jerk and you will end up with a no firearms for rent-a-cops bill next session. &  keep hiring\colorbox{emerald}{imbeciles}like this\colorbox{emerald}{jerk}and you will end up with a no firearms for rent-a-cops bill next session. \\
\toprule
\textbf{Detoxification} & \textbf{Answer} \\
\toprule
 what a chicken crap excuse for a reason. & what a bad excuse for a reason. \\
\bottomrule
\end{tabular}
}
\label{table:general_examples}
\end{table}

In summary, we make the following contributions:

\begin{itemize}
    \item To the best of our knowledge, we perform the first systematic evaluation using prompt tuning to tackle the problem of toxic content.
    \item We leverage prompt tuning to solve the three most representative tasks in this domain, i.e., toxicity classification, toxic span detection, and detoxification.
    \item Extensive evaluations show that our prompt tuning methods can achieve comparable or even better performance than the SOTA methods.
    Also, we observe that prompt tuning has promising performance on fast adaptation to different tasks, i.e., with fewer training samples/epochs.
\end{itemize}

\mypara{Implications}
Our work has important implications for various stakeholders involved in understanding and mitigating online abuse, hate, and harassment.
First, we make our code and annotated dataset available, enabling social media operators to implement solutions to detect and moderate toxic content.
Our approach is superior to previous efforts when considering the annotated data requirements, the performance, the time cost, and the robustness/transferability of the proposed solution.
Additionally, our work can be used to build explainable toxic detection/moderation tools, given our method's outstanding performance on the toxic span detection and detoxification tasks.
Third, we argue that our work can assist and motivate the research community in leveraging the prompt tuning approach for solving other emerging socio-technical issues, such as the spread of misinformation online.
Overall, our work is an important step towards understanding the power and generalizability of LLM in solving hard tasks (e.g., online toxicity), which is an important and timely issue, given the extensive popularity of LLM and chatbots powered by LLM (e.g., ChatGPT).

\mypara{Ethical Considerations} 
We emphasize that in this work we work exclusively with publicly available datasets focusing on toxicity classification, toxic span detection, and detoxification tasks.
Also, we use publicly available large language models to assess their performance on these tasks and how our work compares to previous efforts.
We acknowledge that since we model all three tasks as generation tasks, the model may generate toxic content, however, we took the following steps to minimize harm:
1) we do not share the generated content with people or online users; and
2) all annotations required for our work were done by the authors of this study.
Finally, in this work, we show that using prompt-tuning, large language models can detoxify content with acceptable performance.
At the same time, however, adversaries might use large language models and prompt tuning to do the opposite task (i.e., toxifying content).
We believe that this potential abuse is outside of the scope of this work.
Yet, it highlights the need for the implementation and use of appropriate safeguards (e.g., similar to Stable Diffusion's Safety Filter\footnote{\url{https://stability.ai/blog/stable-diffusion-public-release}.}), to ensure that large language models and prompt tuning can not be used for malicious purposes (e.g., generation and dissemination of toxic content).

\section{Preliminary}

\mypara{Prompt Learning}
With the advance of pre-trained LLM such as GPT-2/3, the previous ``pre-train, fine-tune'' procedure is replaced by the ``pre-train, prompt, and predict'' paradigm~\cite{LYFJHN21}.
Concretely, given a downstream task, fine-tuning requires the training objective to be specified beforehand and the model needs to be updated.
In contrast, prompt learning~\cite{BMRSKDNSSAAHKHCRZWWHCSLGCCBMRSA20} uses a \emph{prompt} that contains the task-specific description and text examples in a natural language way as the input to the model.
In this way, the downstream task can be formulated as a \mask language modeling problem (i.e., predict masked text pieces based on the context) and does not need to update the parameters in the underlying model.
Prompt learning is especially suitable for few-shot downstream tasks when limited training examples are available and fine-tuning the pre-trained model is costly.
In general, prompt learning can be broadly grouped into two categories - manual prompt and learnable prompt (soft prompt).

\mypara{Manual Prompt}
The natural way to create prompts is to manually design intuitive textual templates based on human/domain knowledge~\cite{BMRSKDNSSAAHKHCRZWWHCSLGCCBMRSA20}.
For example, if the task is to classify the sentiment of a movie review ``Absolutely terrible writing and dragged-out unnecessary dialogue", we can append a prompt ``The review is'' to the content and get ``Absolutely terrible writing and dragged-out unnecessary dialogue.
The review is \mask''.
We expect the language model to \emph{generate} ``horrible" than ``great" to replace \mask.
Manual prompts have been proven to solve various tasks with decent accuracy~\cite{LYFJHN21}.
However, handcrafted prompts need to be customized based on the downstream tasks, inevitably introducing artificial bias and leading to sub-optimal results.

\mypara{Learnable Prompt}
In contrast to the manual prompts, learnable prompt methods automatically learn to prompt from  a larger searching space for the candidate prompts to better fit the downstream tasks.
Prefix tuning~\cite{LL21} is one of the most promising techniques for prompt tuning.
Concretely, it adds a prefix (i.e., a sequence of continuous task-specific vectors) before the input, which can be considered as a set of ``virtual tokens''.
Given the downstream task, the prefix will be optimized while the parameters $\theta$ of LM are frozen.
This is extremely efficient compared to fine-tuning the whole model as for different downstream tasks, only different prefixes instead of different models will be updated.
Formally, the prefix matrix $M_{\phi}$ parameterized by $\mathcal{\phi}$ can be updated via the following log-likelihood objective:

\begin{equation}
\label{equation:prefix-tuning}
    \max_{\phi} log P(\bm{y}|\bm{x}; \theta; \phi) = \max_{\phi} \sum_{y_i} log P(y_i | h_{< i}; \theta; \phi)
\end{equation}

\noindent where $h_{<i} = [h_{<i}^{(1)}; \cdots; h_{<i}^{(n)}]$ is a function of the trainable parameters at time step $i$.
It is directly copied from $M_{\phi}$ if the time step is within the prefix ($h_i$ is $M_{\phi}[i]$), otherwise it is computed with the LM.
Similarly, Lester et al.~\cite{LAC21} propose a more efficient method that adds several tunable tokens as the prefix and optimizes the embeddings of those tunable tokens directly.
It has fewer tunable parameters as it does not involve additional tunable parameters in each network layer.
Note that the learnable prompt (prefix matrix) is the embedding of a set of ``virtual words'' which can be optimized.
The embeddings have mathematical meanings but cannot be mapped into real words.

\section{Tasks}

In this work, we consider three tasks that are related to toxicity: 1)~toxicity classification (detect whether the text is toxic), 2)~toxic span detection (detect which parts of the text are toxic), and 3)~detoxification (eliminate toxicity in the text while preserving its semantics).
The three tasks handle toxicity in different levels: toxicity classification only detects whether the whole text is toxic or not; toxic span detection aims to detect the exact character offset of the spans that make the text to be toxic, and detoxification's goal is to eliminate the toxic content from the text while preserving its semantic meaning.

\subsection{Task1: Toxicity Classification}

\mypara{Goal}
We frame this task as a binary classification task, where the input is a piece of text and the output is \emph{whether the given text is toxic or not}.
An example of toxicity classification is shown in \autoref{table:general_examples}.

\mypara{Existing Methods}
Existing toxicity classification methods usually leverage a labeled dataset (a text is annotated as toxic or not) to train classifiers or fine-tune an LM.
Early efforts widely use feature engineering (e.g., dictionaries, bag-of-words, etc.) to extract features from text and detect toxic language or phrases~\cite{DCLT19}.
With the advance of deep neural networks (DNNs), recent efforts have been focusing on training toxicity classification models based on recurrent neural networks (RNNs)~\cite{PMA17}, convolutional neural networks (CNNs)~\cite{GTVP18}, and transformers (e.g., BERT)~\cite{Detoxify}.
The very latest trend of toxicity classification is using LLMs that are pre-trained on large unlabeled corpora and then fine-tuning them to tailor them for the toxicity classification task~\cite{ZZH21}.
The drawback of these methods is that they require a large annotated corpus to train or fine-tune an LM and their detection effectiveness is limited by either the size of the labeled dataset or the time to fine-tune the pre-trained LMs.

\mypara{Our Method}
Given the language model parameterized by $\theta$, a set of texts $\{ \bm{x}|\bm{x} \in X\}$ and the corresponding label $\{ \bm{y} \in Y\}$, we aim to learn the prefix matrix $M_{\phi}$ so that the prompt consist with $M_{\phi}$ (parameterized by $\mathcal{\phi}$) and $\bm{x}$ can successfully retrieve label $\bm{y}$ from the language model $\theta$.
Our optimization goal is summarized in \autoref{eq:task1_loss_function}.

\begin{eqnarray}
\label{eq:task1_loss_function} 
{\phi}^* & = & \underset{{\phi}} {\arg\,\min} ~~~~\mathcal{L}(f(X, {\phi}, \theta), Y)
\end{eqnarray}
\noindent where $\mathcal{L}$ is our loss function (e.g., binary cross-entropy loss) and $f$ is our toxicity classification model.
It is important to note that our model does not fine-tune the language model parameterized by $\theta$.

\subsection{Task2: Toxic Span Detection}

\mypara{Goal}
The toxic span detection aims to identify the specific spans (i.e., the character offsets) that make the text toxic.
For instance, in the example shown in \autoref{table:general_examples},
the toxic span detection task should return two spans - one for ``imbeciles'' (starting at 13 and ending at 21) and one for ``jerk'' (starting at 33 and ending at 36).
It is another important task as it can assist users in better understanding how the toxicity is reflected in the text (e.g., the highlighted toxic span can assist annotators to support their decisions).
Formally, given an input text $t$, our goal is to determine the exact toxic spans $\{S^t\}$ in the text.

\mypara{Existing Methods}
Toxic span detection can be seen as a case of attribution or rationale extraction\cite{PSLA21}.
Most of previous work~\cite{DCLT19,JCLWZL20,HS97} frame this task as a sequence labeling task.
Concretely, given the labeled toxic span corpus, an LM can be trained to label each word as toxic or not.
Once the model is trained and given a text the model will give a toxicity prediction label for each word.
Existing methods have been widely using transformers (e.g., BERT+CRF~\cite{DCLT19}, SPAN-BERT~\cite{JCLWZL20}) or recurrent neural networks (e.g., BiLSTM~\cite{HS97}) to attain the goal.
Some research also experimented with custom loss~\cite{WFL21} and data augmentation~\cite{SJ21} to boost the performance of toxic span detection.

\mypara{Our Method}
Our method is fundamentally different from the existing methods.
Instead of considering the toxic span detection as a sequence labeling task, we treat it directly as a generation task.
Concretely, the input of our model is the original text that contains the toxic content.
We aim to leverage the prompt and the (frozen) LLM to generate text without the toxic span while keeping the rest the same as the input text.
Note that, with the prompt, the LLM does not attempt to replace the toxic span in the generated text, rather it generates a, usually, incomplete text that does not have any toxic spans.
Then, to detect the toxic span, we run a mapping algorithm to ``subtract'' the input text from the generated text and consider the rest as the toxic spans (i.e., character-level offsets).
Our optimization goal, given the input $T=\{t\}$ and $\tilde{T} = \{ t \setminus \{S^t\}  \}$, is summarized in \autoref{eq:task2_loss_function}.

\begin{eqnarray}
\label{eq:task2_loss_function}
{\phi}^*  & = & \underset{{\phi}} {\arg\,\min} ~~~~ \mathcal{L} (\tilde{T}, f(T, {\phi}, \theta))
\end{eqnarray}

\noindent It learns $M_{\phi}$ (parameterized by $\mathcal{\phi}$) that nudges the large language model $\theta$ to remove only toxic spans $\{S^t\}$ from $X$.

\subsection{Task3: Text Detoxification}

\mypara{Goal}
Text detoxification, as its name suggests, aims to eliminate toxicity from text and generate a detoxified version of the text while preserving the semantic meaning.
Different from the previous tasks that only focus on the detection of toxicity (e.g., toxicity classification and toxic span identification), text detoxification addresses the toxic content by proactively rewriting it.
An example of toxicity detoxification is shown in \autoref{table:general_examples}.
Formally, for this task, the input is a toxic text $t$ and our goal is to generate a detoxified version of the text $\hat{t}$.

\mypara{Existing Methods}
Text detoxification can be viewed as a style transfer task.
That is, toxicity can be treated as the style of a text.
The style transfer methods are applied to rewrite the text with similar semantic meaning without the toxicity style.
In previous work~\cite{PLXSA22,LDUMDKSP22}, both supervised and unsupervised methods are proposed to solve this task in a style transfer manner.
Logacheva et al.~\cite{LDUMDKSP22} propose DetoxBART, which fine-tunes the Transformer-based generation model BART~\cite{LLGGMLSZ20} on the ParaDetox dataset.
Such fune-tuning process makes DetoxBART yield the best performance in terms of detoxification and semantic preservation.
The other end-to-end approaches include DualRL~\cite{LLZYCSS19}, Deep Latent Sequence Model (DLSM)~\cite{HWNB20},  Stable Style Transformer (SST)~\cite{L202}, Style
Transfer as Paraphrase (STRAP)~\cite{KWI20}, Paraphrasing GeDi (ParaGeDi)~\cite{DVDLKSP21}, etc.

\mypara{Our Method}
The detoxification task is also a generation task.
Given the paired dataset (i.e., the toxic text $T$ and the paraphrased non-toxic counterpart $\hat{T}$), our goal is to learn the prompt $M_\phi$ that can better transfer the input text (toxic) into the output text (non-toxic) text while preserving the semantics.
The optimization goal is similar to \autoref{eq:task2_loss_function} and the only difference is that the label changes from $\tilde{T}$ to $\hat{T}$ where the former is the texts without toxic spans (incomplete texts) and the later is the detoxified texts (complete texts).

\section{Datasets and Models}
\label{section:datasets_and_models}

\subsection{Datasets}

\begin{table}[t]
\centering
\caption{Overview of datasets. 
Note that $\ast$ means the dataset provides the train/test partition.}
\scalebox{0.9}{
\begin{tabular}{lccc}
\toprule
\textbf{Dataset} & \textbf{Task}  & \textbf{\# Train} & \textbf{\# Test} \\
\midrule
\textbf{HateXplain}~\cite{MSYBGM21}             & 1  & 12,578 & 3,050        \\
\textbf{USElectionHate20}~\cite{GK21} $\ast$    & 1  & 586 & 118        \\
\textbf{HateCheck}~\cite{RVNWMP21}              & 1  & 1,998 & 484        \\
\textbf{SBIC}~\cite{SGQJSC20} $\ast$            & 1  & 93,346 & 11,000        \\
\textbf{MHS}~\cite{KBSV20}                      & 1  & 22,700 & 5,762       \\
\textbf{ToxicSpan}~\cite{PSLA21}  $\ast$        & 2   & 7,888 & 1,991        \\
\textbf{Parallel}~\cite{DUDKSPL21}              & 3  & 886 & 222        \\
\textbf{ParaDetox}~\cite{LDUMDKSP22}            & 3  & 9,551 & 2,388        \\
\bottomrule
\end{tabular}
}
\label{table:dataset_statistic}
\end{table}

In this paper, we consider eight datasets for the evaluation of the three tasks.
Note that, in Task 1 (toxicity classification), for each dataset, we generate a balanced version of it by randomly choosing the same number of samples from the larger category to match the smaller category.
We follow the train/test partition of a dataset if they have already been provided.
Otherwise, we randomly sample 80\% of a dataset as the training dataset and the rest 20\% as the testing dataset.
\autoref{table:dataset_statistic} reports some basic statistics about each dataset.
We describe each dataset below.

\mypara{HateXplain~\cite{MSYBGM21}}
It is a benchmark dataset collected from Twitter and Gab for explainable hate speech detection.
The dataset is annotated by Amazon Mechanical Turk (MTurk) workers with three labels: hate, offensive, or normal.
For our work, we consider both hate and offensive posts as toxic and the rest as non-toxic.

\mypara{USElectionHate20~\cite{GK21}}
This dataset is collected from Twitter by selecting tweets that contain election hashtags or politicians' names.
The authors manually label a subset of tweets with different stances as well as whether the tweet is hateful/offensive.
We consider hateful/offensive tweets as toxic and the rest as non-toxic.

\mypara{HateCheck~\cite{RVNWMP21}}
HateCheck contains a suite of functional tests for hate speech detection models.
Each post is labeled by different annotators and we consider the majority votes as the final label of this post.

\mypara{SBIC~\cite{SGQJSC20}}
The Social Bias Inference Corpus (SBIC) is collected from Reddit, Twitter, and fringe Web communities such as Gab, Stormfront, and banned subreddits.
The dataset is labeled by MTurk workers.
We leverage the v2 version of it for our study and we consider posts labeled offensive as toxic posts and the rest as non-toxic posts.

\mypara{MHS~\cite{KBSV20}}
The Measuring Hate Speech (MHS) dataset is collected from comments on social media like YouTube, Twitter, and Reddit.
The corpus is labeled by MTurk workers from the US.
We consider comments with hate speech score $\ge$ 0 as toxic and all others as non-toxic.

\mypara{ToxicSpan~\cite{PSLA21}}
The ToxicSpan dataset contains $\sim$10k English texts filtered from Civil Comments~\cite{BDSTV19} and was formally introduced as SemEval-2021 Task 5~\cite{PSLA21}.
Each text is reviewed by three to seven raters.
Each rater is asked to identify the spans ``that constitute anything that is rude, disrespectful or unreasonable that would make someone want to leave a conversation"~\cite{PLXSA22}.
The lengths of the highlighted spans were decided by the raters.

\mypara{Parallel~\cite{DUDKSPL21}}
The Parallel dataset contains 2,279 pairs of (toxic sentence, detoxified sentence).
There are 1,108 unique toxic sentences after removing duplicates.
Note that for each toxic sentence, the dataset might offer multiple detoxified versions.
We only select the first detoxified version to construct the pair.

\mypara{ParaDetox~\cite{LDUMDKSP22}}
ParaDetox contains 11,939 toxic sentences and 19,766 paraphrased sentences (detoxified sentences).
Similar to the Parallel dataset, each toxic sentence might have multiple detoxified versions.
We only pick the first detoxified version to construct the pair.
The ParaDetox dataset constructed by us has 11,939 pairs in total.

\mypara{Remarks}
All the datasets are annotated by human annotators.
However, the definition of toxicity might vary across different datasets.
For instance, USElectionHate20 targets hateful tweets against politicians, while SBIC focuses on offensive posts from different Web communities.
This may bring challenges for toxicity classifiers such as the Perspective API~\cite{Perspective}.
On the other hand, our approach diminishes this issue, given that we use a learnable prompt that is tailored for each dataset, effectively capturing the toxic definition of the dataset through the lens of the positive and negative samples in each dataset.

\subsection{Models}

In this paper, we consider prompt tuning over two families of LLM including GPT2~\cite{RWCLAS19} and T5~\cite{RSRLNMZLL20}.
Concretely, we use GPT2-medium, GPT2-large, T5-small, T5-base, and T5-large in our experiments.
In Task 1 (Toxicity Classification), the learning rate is set to 0.3, we set the total optimization steps to 2,000 with Adafactor~\cite{SS18} optimizer and the linear learning rate scheduler with 100 warm-up steps.
For all models, the effective batch size is set to 32 (batch size of 4/8 with gradient accumulation steps of 8/4 for GPT2-L/Others).
We follow the prompt tuning method proposed by Lester et al.~\cite{LAC21} in Task 1.
In Task 2 (Toxic Span Detection) and Task 3 (Detoxification), we set the training epoch to 5, the initial learning rate to 5e-5, and the optimizer of AdamW~\cite{LH19} with the linear learning rate scheduler.
Different from Task 1 (Toxicity Classification), we follow the prompt tuning method proposed by Li and Liang~\cite{LL21} instead as it can achieve better performance in Task 2 and Task 3.
We hypothesize that Lester et al.~\cite{LAC21} initializes the prompt with embeddings that enumerate the output classes, which makes the method more suitable for the classification task.
In contrast, the prompt tuning method proposed by Li and Liang~\cite{LL21} has more tunable parameters than the one proposed by Lester et al.~\cite{LAC21}.
This method learns transformer activations that are fixed across examples at every network layer, allowing subsequent tokens to attend to this prefix.
As such, Li and Liang~\cite{LL21} is a better fit for Task 2 (Toxic Span Detection) and Task 3 (Detoxification).

\section{Task 1: Toxicity Classification}

\begin{table*}[ht]
\centering
\caption{$F_1$-score of Task 1. 
The best results of each dataset are highlighted in bold.}
\scalebox{0.9}{
\begin{tabular}{l |ccc|ccccc}
\toprule
\multirow{2}{*}{\textbf{Dataset}}   & \multicolumn{3}{c|}{\textbf{Baselines}} & \multicolumn{5}{c}{\textbf{Prompt Tuning}}  \\
                                    & \textbf{Perspective} & \textbf{ToxicBERT} & \textbf{UnRoBERTa} & \textbf{GPT2-M} & \textbf{GPT2-L} & \textbf{T5-S} & \textbf{T5-B} & \textbf{T5-L} \\
\midrule
\textbf{HateXplain} & 0.703 & 0.657 & 0.648 & 0.016 & \textbf{0.731} & 0.716 & \textbf{0.731} & 0.637 \\
\textbf{USElectionHate20} & 0.506 & 0.488 & 0.425 & 0.709 & 0.741 & 0.673 & \textbf{0.833} & 0.660 \\
\textbf{HateCheck} & 0.784 & 0.670 & 0.671 & 0.758 & 0.892 & 0.860 & 0.841 & \textbf{0.946} \\
\textbf{SBIC.v2} & 0.669 & 0.581 & 0.581 & 0.721 & \textbf{0.854} & 0.820 & 0.844 & 0.841 \\
\textbf{MHS} & \textbf{0.790} & 0.768 & 0.775 & 0.711 & 0.758 & 0.762 & 0.775 & 0.776 \\
\midrule
\textbf{Avg.} & 0.690 & 0.633 & 0.620 & 0.583 & 0.795 & 0.766 & \textbf{0.805} & 0.772 \\
\bottomrule
\end{tabular}
}
\label{table:task1_performance_f1}
\end{table*}

\subsection{Experimental Setup}

\mypara{Baselines}
Regarding the baselines for Task 1, we consider Google's Perspective API~\cite{Perspective} (Perspective), BERT-base trained on toxicity classification corpus~\cite{Detoxify} (ToxicBERT), and RoBERTa-base trained on toxicity classification corpus~\cite{Detoxify} (UnRoBERTa).
For each baseline, given a text, it provides a toxicity score ranging from 0 to 1.
We consider the text with a score larger than 0.5 as toxic otherwise non-toxic.
The results with the best threshold (rather than 0.5) are shown in \autoref{table:dynamic_threshold} in Appendix.
Note that for Perspective API, on each dataset, we select the perspective score (e.g., Severe Toxicity) that achieves the best classification result, and report the corresponding performance.

\mypara{Datasets}
We use five datasets - HateXplain, USElectionHate20, HateCheck, SBIC, and MHS - to evaluate the baselines and our models.
Note that we observe redundant samples on HateXplain, USElectionHate20, and SBIC.v2.
However, they are less than 1\% and have almost no influence on the final performance based on our initial evaluation.

\mypara{Metrics}
We consider accuracy, precision, recall, and $F_1$-score as the evaluation metrics, which are standard metrics for evaluating the performance of classifiers.
Note that we only report the $F_1$-score on the main paper and put the precision, recall, and accuracy results in \autoref{appendix:task1_performance_other_metrics} in Appendix.

\subsection{Results}

\mypara{Overall Performance}
We first show the $F_1$-score of toxicity classification with toxicity classification in \autoref{table:task1_performance_f1}.
The accuracy, precision, and recall are shown in \autoref{table:task1_performance_acc}, \autoref{table:task1_performance_precision}, and \autoref{table:task1_performance_recall} in the Appendix.
We find that, in general, prompt tuning outperforms baselines across different datasets.
For instance, on HateXplain, the prompt tuning with GPT2-L achieves 0.731 $F_1$-score, while the best baseline (Perspective) only achieves 0.703 $F_1$-score.
The statistical test shows that prompt tuning indeed outperforms the best baseline (see \autoref{table:task1_statistical_test} in Appendix).
This indicates that prompt tuning can indeed unleash the power of LLM to perform the toxicity classification task.
Also, we observe that a larger LM usually provides a more promising performance on the task, e.g., GPT2-L usually outperforms GPT2-M and T5-B/L is better than T5-S in general.
For instance, on HateCheck, GPT2-L achieves 0.892 $F_1$-score while GPT2-M only has 0.758 $F_1$-score.
This implies that the larger capacity of LLM would better guide the prompt tuning to achieve better performance.

\begin{figure*}[!t]
\centering
\begin{subfigure}{0.18\linewidth}
\includegraphics[width=\columnwidth]{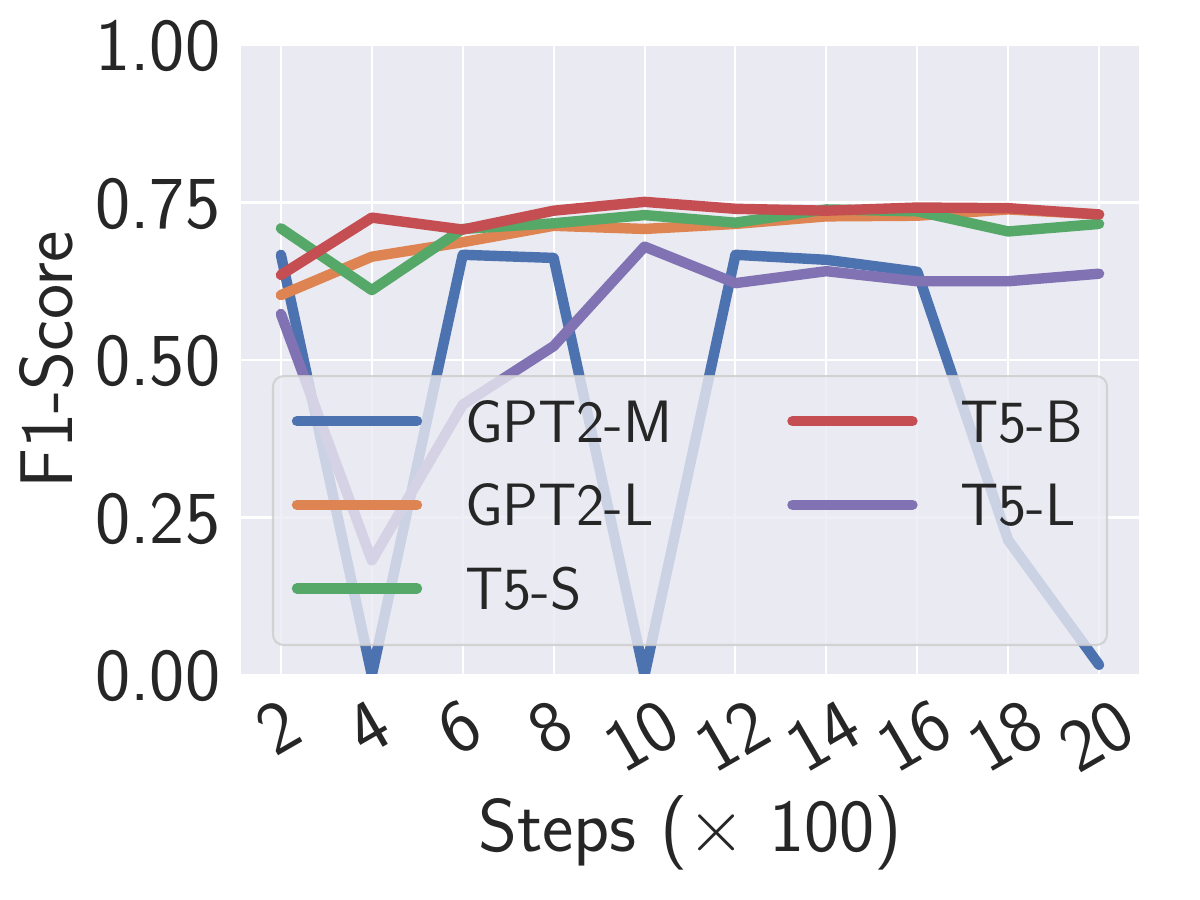}
\caption{HateXplain}
\label{figure:ablation_fewer_steps_HateXplain_f1}
\end{subfigure}
\begin{subfigure}{0.18\linewidth}
\includegraphics[width=\columnwidth]{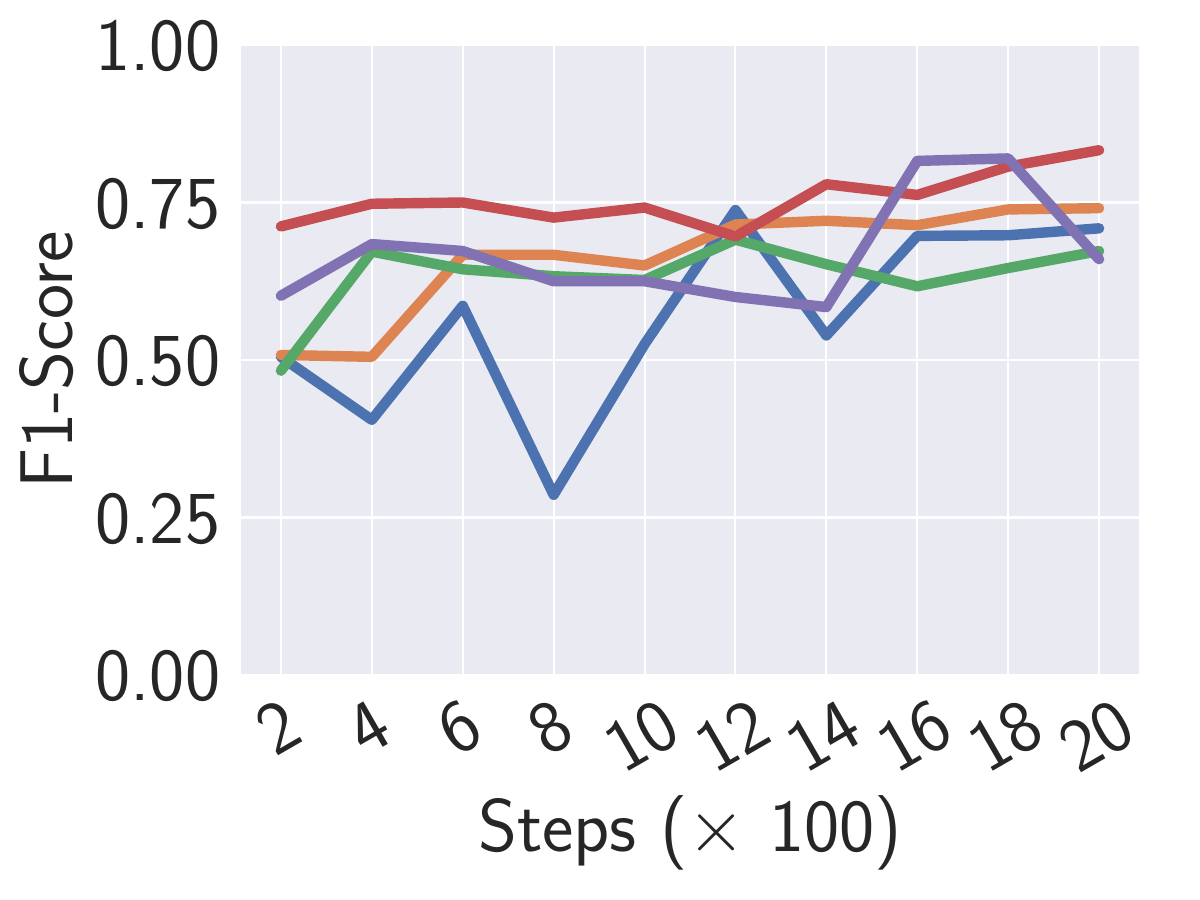}
\caption{USElectionHate20}
\label{figure:ablation_fewer_steps_USElectionHate20_f1}
\end{subfigure}
\begin{subfigure}{0.18\linewidth}
\includegraphics[width=\columnwidth]{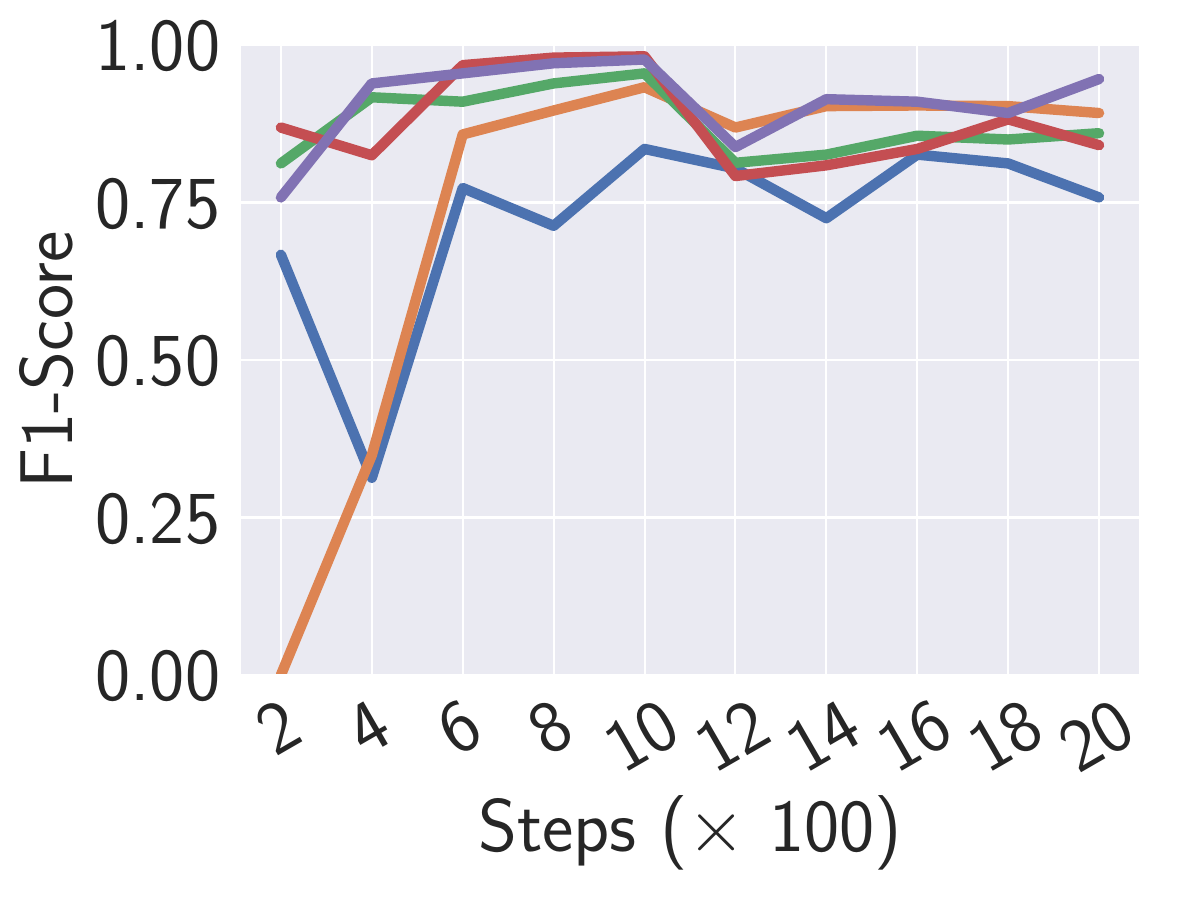}
\caption{HateCheck}
\label{figure:ablation_fewer_steps_HateCheck_f1}
\end{subfigure}
\begin{subfigure}{0.18\linewidth}
\includegraphics[width=\columnwidth]{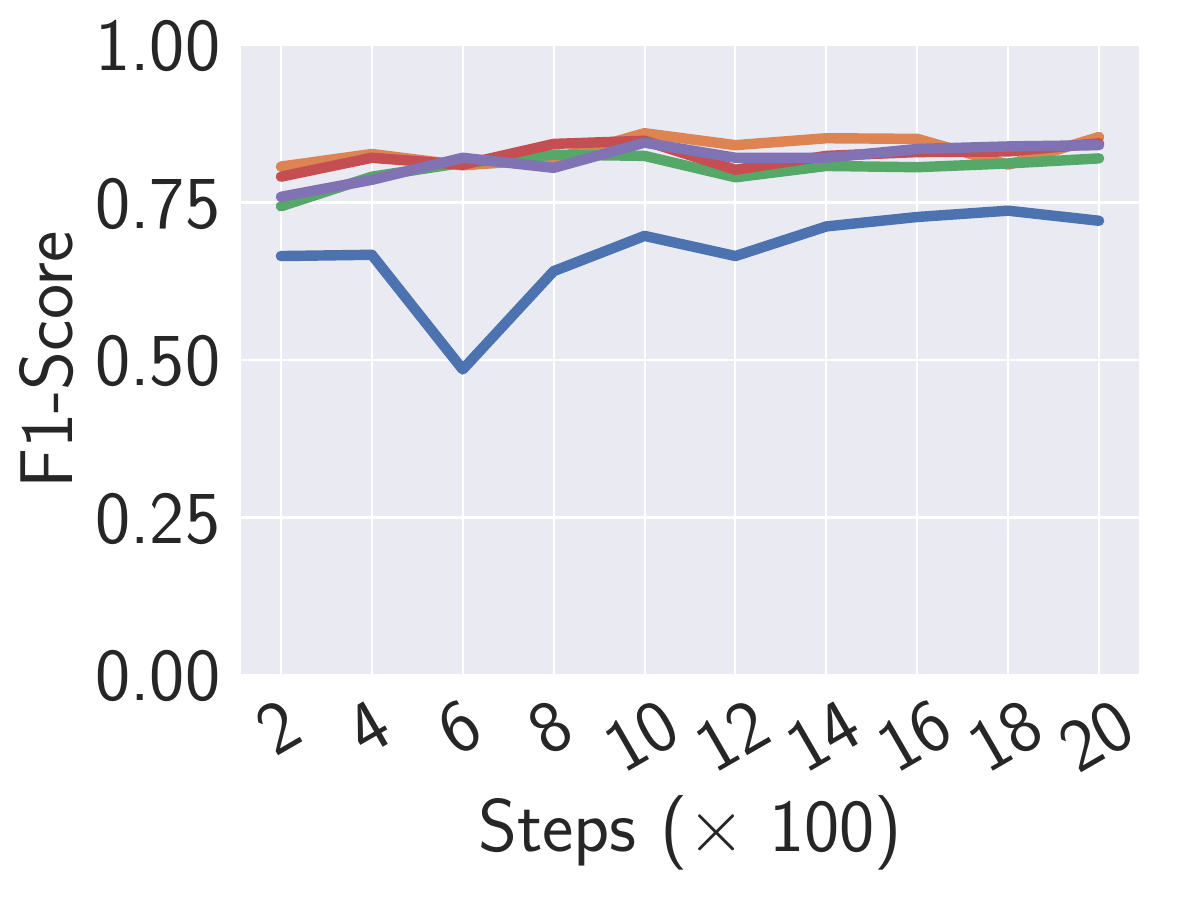}
\caption{SBIC}
\label{figure:ablation_fewer_steps_SBIC_f1}
\end{subfigure}
\begin{subfigure}{0.18\linewidth}
\includegraphics[width=\columnwidth]{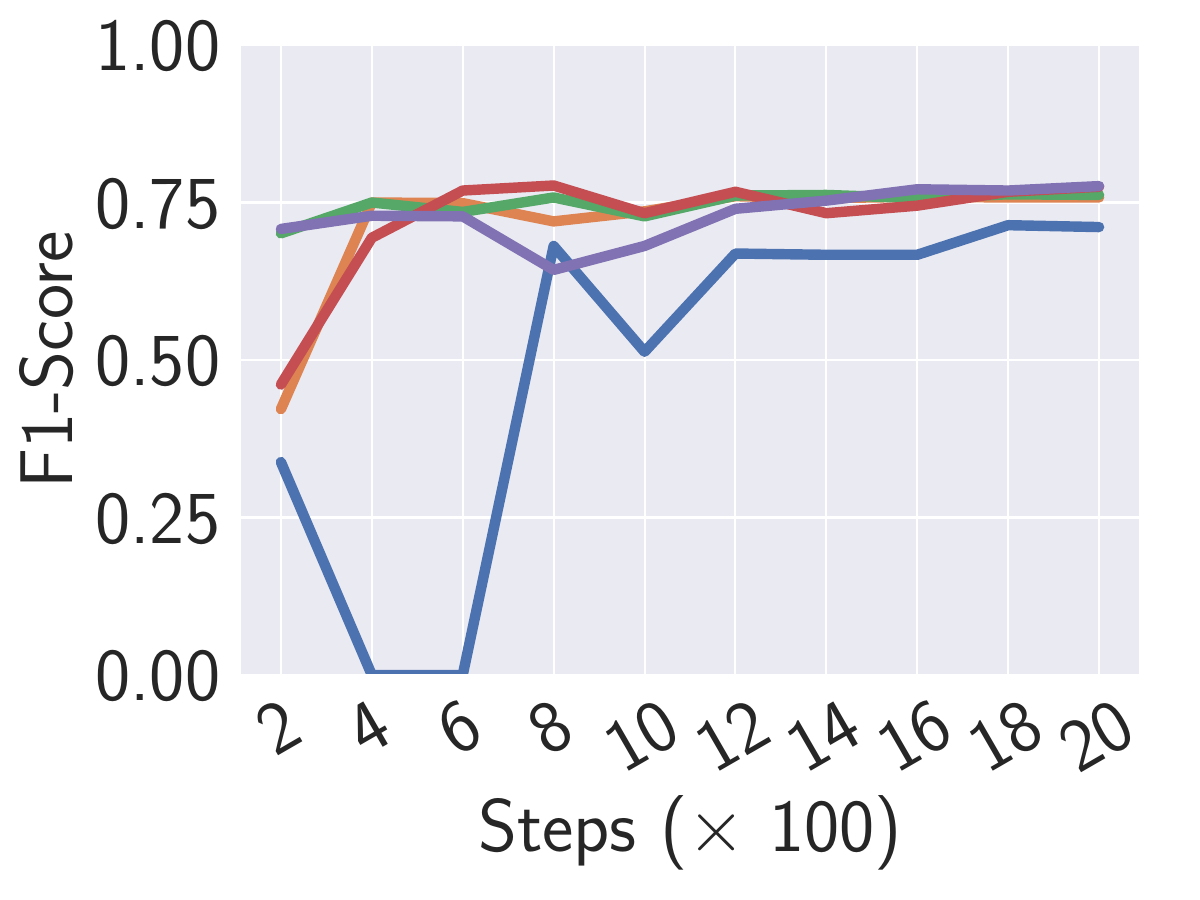}
\caption{MHS}
\label{figure:ablation_fewer_steps_MHS_f1}
\end{subfigure}
\caption{$F_1$-score of Task 1 with different training steps.}
\label{figure:task1_fewer_steps_f1}
\end{figure*}

\begin{table}[ht]
\centering
\caption{$F_1$-score of Task 1 with manual prompt.}
\scalebox{0.9}{
\begin{tabular}{l | ccccc}
\toprule
{\textbf{Dataset}} & \textbf{GPT2-M} & \textbf{GPT2-L} & \textbf{T5-S} & \textbf{T5-B} & \textbf{T5-L} \\
\midrule
\textbf{HateXplain} & 0.080 & 0.111 & 0.041 & 0.276 & 0.667 \\
\textbf{USElectionHate20} & 0.030 & 0.275 & 0.213 & 0.289 & 0.632 \\
\textbf{HateCheck} & 0.056 & 0.032 & 0.256 & 0.008 & 0.569 \\
\textbf{SBIC.v2} & 0.255 & 0.151 & 0.110 & 0.086 & 0.619 \\
\textbf{MHS} & 0.134 & 0.157 & 0.205 & 0.188 & 0.664 \\
\midrule
\textbf{Avg.} & 0.111 & 0.145 & 0.165 & 0.169 & 0.630 \\
\bottomrule
\end{tabular}
}
\label{table:task1_ablation_manual_prompt_f1}
\end{table}

\mypara{Comparison to Manual Prompt}
As previous work~\cite{SUS21,CA21} shows that manual prompts can achieve decent performance in the toxicity classification task, we investigate how the performance of prompt tuning compares to manual prompts.
We use the below manual prompt following Schick et al.~\cite{SUS21} for our comparison study.
Note that we also investigate using the descriptions of 6 different scores in Perspective API as the manual prompt and the results show similar trends.

\mybox{
\noindent \input \\
Question: Does the above text contain rude, disrespectful, or unreasonable language?\\
Answer:\mask
}

\noindent Here, \input is the placeholder for the text to be classified, ``Question: Does the above text contain rude, disrespectful, or unreasonable language? Answer:'' is our manual prompt and \mask is the classification output by the LLM.
The performance is shown in \autoref{table:task1_ablation_manual_prompt_f1}.
We observe that the $F_1$-score of the manual prompt is substantially lower than the prompt tuning approach (see \autoref{table:task1_performance_f1}).
For instance, for the average results, with T5-S, prompt tuning achieves 0.766 $F_1$-score while manual prompt only reaches 0.165.
These results highlight the effectiveness and performance gains when using prompt tuning instead of manual prompts.

\mypara{Fewer Training Steps}
In our previous experiments, we use 2,000 training steps during the prompt tuning procedure.
Here, we investigate how the selection of the value for the training steps affects performance.
\autoref{figure:task1_fewer_steps_f1} summarizes the $F_1$-score for different language models wrt. different training steps.
We observe that the $F_1$-score increases during the initial steps and at some point, the $F_1$-scoregains diminish.
For instance, on HateXplain (see \autoref{figure:ablation_fewer_steps_HateXplain_f1}), from 200 to 800 steps, the $F_1$-score with GPT2-L increases from 0.603 to 0.713, and it stabilizes at 0.731 with 2,000 steps.
This indicates that the prompt tuning can adapt to the downstream task faster, which is important as it can save both time and computational power.

\mypara{Fewer Training Samples}
In previous experiments, we randomly sample 80\% of the dataset as the training dataset and the rest 20\% as the testing dataset.
Here, we investigate whether the prompt tuning can still work well with fewer training samples.
We use the T5 models as the case study as we observe from \autoref{figure:task1_fewer_steps_f1} that T5 models are relatively stable and can achieve good performance in general.
Concretely, for each dataset, we randomly select 500 training samples to form the new training dataset and optimize the prompt for 1,000 steps only.
Note that we exclude the USElectionHate20 dataset for this assessment as it only has 586 training samples, which is close to 500.
The results are summarized in \autoref{table:task1_ablation_500_training_samples_f1}.
We observe that although the performance is lower than training with full data, it is still comparable to or even better than the baselines.
Take SBIC as an example, with 500 samples, the T5-B model achieves 0.782 $F_1$-score, which is lower than training with full data (0.844) but still higher than the Perspective API  with 0.669 $F_1$-score (see \autoref{table:task1_performance_f1}).
Note that 500 training samples are only around 0.5\% of the SBIC's original training dataset, which is a quite small fraction.
This provides us with a new perspective of view to transfer the toxicity-related research into new datasets: instead of using the mature published APIs like Perspective, leveraging prompt tuning with a small fraction of data being labeled is also a promising way to reach desirable performance.

\begin{table}[t]
\centering
\caption{$F_1$-score of Task 1 with 500 training samples on each dataset.}
\scalebox{0.9}{
\begin{tabular}{l | cccc}
\toprule
{\textbf{Dataset}} & \textbf{T5-S} & \textbf{T5-B} & \textbf{T5-L} \\
\midrule
\textbf{HateXplain} & 0.624 & 0.666 & 0.655 \\
\textbf{HateCheck} & 0.865 & 0.897 & 0.654 \\
\textbf{SBIC.v2} & 0.772 & 0.782 & 0.764 \\
\textbf{MHS} & 0.659 & 0.694 & 0.644 \\
\midrule
\textbf{Avg.} & 0.730 & 0.760 & 0.679 \\
\bottomrule
\end{tabular}
}
\label{table:task1_ablation_500_training_samples_f1}
\end{table}

\begin{table*}[t]
\centering
\caption{$F_1$-score of Task 1 (Toxicity Classification) when the training dataset is different from the transfer dataset.}
\scalebox{0.9}{
\begin{tabular}{l |ccccc}
\toprule
\multirow{2}{*}{\textbf{Training Dataset}}   & \multicolumn{5}{c}{\textbf{Transfer Dataset}} \\
 & \textbf{HateXplain} & \textbf{USElectionHate20} & \textbf{HateCheck} & \textbf{SBIC} & \textbf{MHS} \\
\midrule
\textbf{HateXplain} & - & 0.488 & 0.373 & 0.419 & 0.688 \\
\textbf{USElectionHate20} & 0.650 & - & 0.472 & 0.485 & 0.733 \\
\textbf{HateCheck} & 0.543 & 0.297 & - & 0.534 & 0.579 \\
\textbf{SBIC.v2} & 0.638 & 0.404 & 0.646 & - & 0.655 \\
\textbf{MHS} & 0.694 & 0.581 & 0.610 & 0.518 & - \\
\bottomrule
\end{tabular}
}
\label{table:task1_performance_transfer_t5-base_f1}
\end{table*}

\mypara{Prompt Transferability}
Finally, we assess the generalizability power of prompt tuning by investigating the performance when training a prompt on one dataset and testing on another.
Here we take the T5-base model as the pre-trained LLM for prompt tuning.
\autoref{table:task1_performance_transfer_t5-base_f1} displays the results.
We can observe that in some cases, the prompt can successfully transfer to another dataset.
For instance, the prompts trained on USElectionHate20 can achieve 0.650 $F_1$-score on HateXplain and 0.733 $F_1$-score on MHS, which are about 5\% lower than the baselines (0.703 accuracy on HateXplain and 0.790 accuracy on MHS according to \autoref{table:task1_performance_f1}).
However, the performance is less satisfying in some other cases where the $F_1$-score is below 0.500.
We also notice that the prompt trained on the MHS dataset can better transfer to other datasets.
For instance, after training on MHS, the $F_1$-score is 0.694 on HateXplain and 0.581 on USElectionHate20, which is comparable or even better to the $F_1$-score provided by the Perspective API (0.703 and 0.506).
This can be credited to the fact that MHS covers various kinds of toxicity including insult, humiliation, violence, hate speech, etc.
By fine-tuning with the diverse distributed data, the learned prompt is more general and can better transfer to other datasets.
On the other hand, prompts learned from dataset like HateXplain is less effective to transfer into other datasets.
We suspect this is because these datasets have a relatively narrow definition of toxicity.
In general, the prompt learned from a more diverse dataset with different types of toxicities may have a better generalization ability to other datasets.
Meanwhile, as we have shown before (see \autoref{table:task1_ablation_500_training_samples_f1}), the prompts can better fit different downstream datasets with the help of only a small fraction of labeled samples, which further demonstrates the efficacy of prompt learning.

\mypara{Comparison with Fine-tuning}
Here we take T5-S on USElectionHate20 as an example.
We observe that prompt tuning reaches 0.712 accuracy within 6 minutes, while the best accuracy (evaluated every 200 steps) for fine-tuning the whole model is only 0.619 within 100 minutes.
This is because the LLM is trained with a large corpus and can generate informative representations of the inputs.
Prompt tuning can guide the model better leverage the representation for the downstream tasks with a small number of parameters, which can adapt faster to new tasks compared to finetuning, especially with fewer training samples.

\mypara{Robustness}
Given the misspellings in the training procedure, we do observe that prompt tuning can adapt to the testing posts with misspellings.
E.g., on 100 randomly selected toxic posts on HateCheck, there do exist misspelling words like ``tr4sh,'' ``4ssholes,'' ``Fukc,'' and ``crippl3.''
And prompt tuning with T5-S can correctly identify them (98\% accuracy).
We further perturb these 100 evaluation posts by randomly repeating one character of each toxic word several times or adding extra spaces inside the toxic word, e.g.,  ``sluttttts,'' and ``w h o r e.''
Note that we leverage such perturbations since we also observe them in the toxic texts and such perturbations are also considered by previous work~\cite{HKZP17}.
We observe that, without further prompt tuning, the evaluation accuracy on these modified 100 posts is still 97\%, which remains almost unchanged.
This implies that prompt tuning is robust to adversarial perturbation.

\mypara{Error Analysis}
Although prompt tuning outperforms other baselines in most cases, wrongly predicted texts still exist (20 in total).
We take the USElectionHate20 dataset (with T5-B) as a case study to analyze the wrongly predicted cases.
As shown in \autoref{table:task1_examples}, the main reason that causes the wrong prediction is the wrong label, e.g., in the example, we observe some toxicity against Trump, but the text is labeled as non-toxic.
Also, we observe that some variations of the slur words and toxic hashtags may cause wrong predictions.
Last, prompt tuning is less effective against some texts with implicit toxic content.

\begin{table}[!ht]
\centering
\caption{Failed examples on USElectionHate20. 
Note that we shorten long texts for presentation purposes.}
\scalebox{0.9}{
\begin{tabular}{p{3cm} p{3cm} c}
\toprule
\textbf{Reason} & \textbf{Example} & \textbf{Percentage (\%)} \\
\midrule
\multirow{6}{*}{Wrong ground truth} & @realDonaldTrump Why would the government stop this when our President is among those guiltiest? Money launder much,Trump? ...  & \multirow{6}{*}{50}\\
\midrule
Slur word variation & F@\&amp;CK, BS  & 10\\
\midrule
\multirow{3}{*}{Hashtag hate} & \#FakeNewsMediaClowns, \#LyinSleepyWiredUpJoeBiden & \multirow{3}{*}{5}\\
\midrule
\multirow{4}{*}{Other} & The biggest threat to our nation dwells within the White House. Vote Biden. Pass it on! ...  & \multirow{4}{*}{35}\\
\bottomrule
\end{tabular}
}
\label{table:task1_examples}
\end{table}

\mypara{Takeaways}
Our results show that prompt tuning outperforms baselines in the toxicity classification task with sufficient labeled data.
Also, the detection performance is still promising with fewer training steps/samples.
Another observation is that directly transferring the prompt trained on one dataset into another dataset might be less effective as the two datasets might share different types of toxicity.
However, this can be addressed by adding only a small number of labeled samples from the distribution of the testing dataset.
Our results suggest that prompt tuning can also serve as an alternative tool to assist the annotation process, especially for the newly emerging toxicity.

\section{Task 2: Toxic Span Detection}

\subsection{Experimental Setup}

As we observed from Task 1 (Toxicity Classification), T5 models and GPT2 models share similar performance.
In the following evaluation, we mainly leverage T5 models as our pre-trained LLMs.

\mypara{Baselines}
We consider three baselines, i.e., BiLSTM~\cite{HS97}, BERT~\cite{DCLT19}, and SPAN-BERT~\cite{JCLWZL20}.
Concretely, we follow the default hyper-parameters setting of Pavlopoulos et al.~\cite{PLXSA22}.
We train/fine-tune the models for 100 epochs on the training partition of the ToxicSpan dataset and evaluate it on its test partition.

\mypara{Datasets}
We use the ToxicSpan dataset to evaluate the baselines and our models.

\mypara{Metrics}
We follow previous work~\cite{PLXSA22} and leverage $F_1$-score as the main evaluation metric.
Note that the $F_1$-score in Task 2 is different from Task 1.
Concretely, for the $i$-th sample, we consider its ground truth span (i.e., the character offsets) as $S_g^i$ and the predicted span as $S_p^i$.
The sample-level precision $P^t$ , recall $P^t$, and $F_1$-score $F_1^t$ are defined as the following:

\begin{align}
    & P^t(S_g^i, S_p^i) = \frac{|S_g^i \cap S_p^i|}{|S_p^i|} \\
    & R^t(S_g^i, S_p^i) = \frac{|S_g^i \cap S_p^i|}{|S_g^i|} \\
   & F_1^t(S_g^i, S_p^i) = \frac{2 \cdot P^t(S_g^i, S_p^i) \cdot R^t(S_g^i, S_p^i)}{P^t(S_g^i, S_p^i) + R^t(S_g^i, S_p^i)}
\end{align}
Note that if the ground truth span $S_g^i$ and the predicted span $S_p^i$ are both empty, we consider $F_1^t(S_g^i, S_p^i)=1$ ($F_1^t(S_g^i, S_p^i)=0$ if one of them is empty).
Then, we average the $F_1$-score for all samples to obtain a single $F_1$-score.

\subsection{Results}

\begin{table}[ht]
\centering
\caption{Performance of Task 2 (Toxic Span Detection).}
\scalebox{0.9}{
\begin{tabular}{l c c}
\toprule
\textbf{Method} & {\textbf{$F_1$}} & Time Cost (Second)\\
\midrule
\textbf{BiLSTM}     & 0.566 & 94 \\
\textbf{BERT}       & 0.629 & 1,828 \\
\textbf{SPAN-BERT} & 0.640 & 3,334 \\
\midrule
\textbf{PT (T5-S)}       & 0.571 & 175 \\
\textbf{PT (T5-B)}       & 0.615 & 363 \\
\textbf{PT (T5-L)}       & 0.643 & 838 \\
\bottomrule
\end{tabular}
}
\label{table:task2_performance}
\end{table}

\begin{figure}[ht]
\centering
\includegraphics[width=0.8\columnwidth]{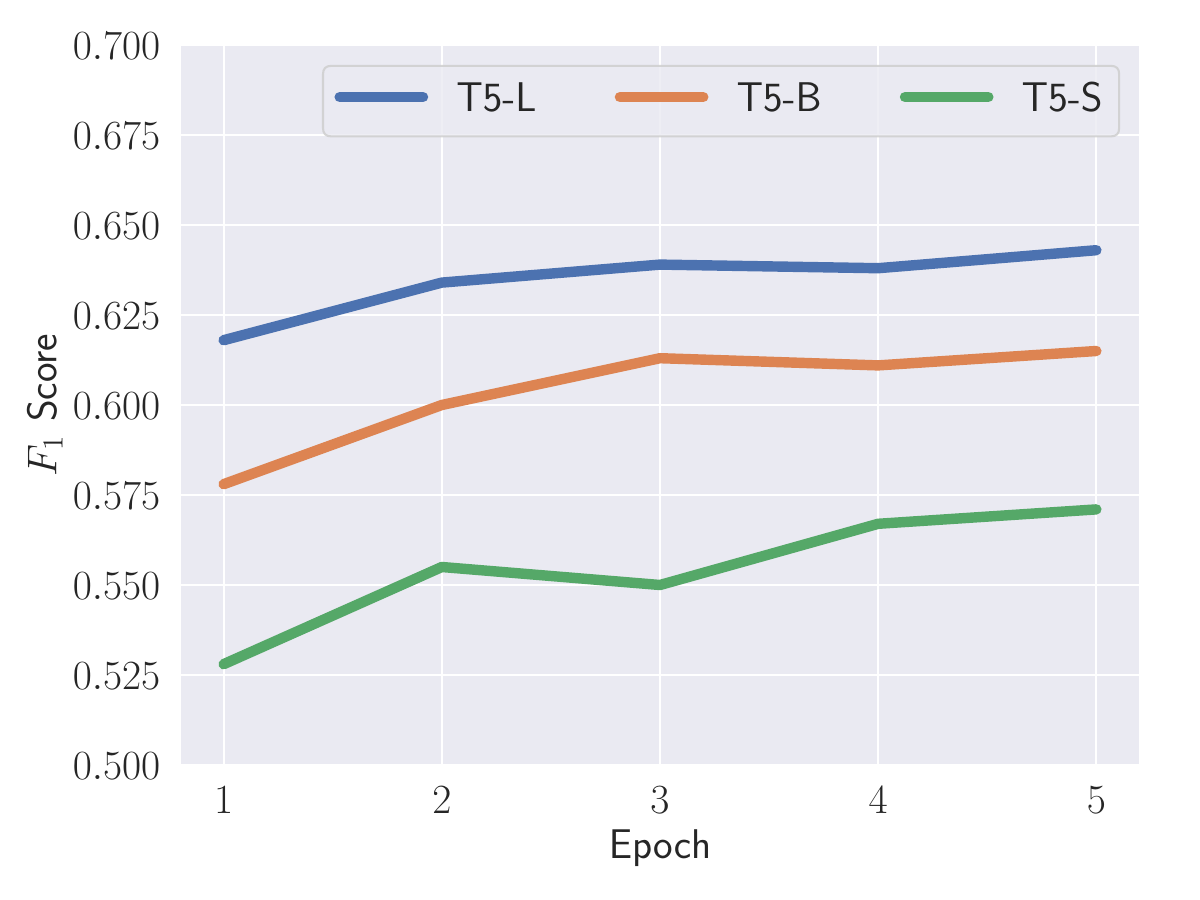}
\caption{$F_1$-score of Task 2 (Toxic Span Detection) with different training epochs.}
\label{figure:task2_fewer_epoch}
\end{figure}

As shown in \autoref{table:task2_performance}, prompt tuning achieves comparable or even better performance than the baselines.
For instance, prompt tuning on T5-L reaches an $F_1$-score of 0.643, which is higher than BiLSTM (0.566), BERT (0.629), and SPAN-BERT (0.640).
On the other hand, prompt tuning achieves this outstanding performance with much less time.
For example, prompt tuning on T5-L only takes 838 seconds, while the SPAN-BERT needs 3,334 seconds for the fine-tuning process.
This is because prompt tuning has fewer parameters to be updated compared to those of fine-tuning the LM.
Another observation is that the prompt tuning achieves better performance with a larger LM, e.g., the $F_1$-score is 0.571, 0.615, and 0.643 on T5-S, T5-B, and T5-L, respectively.
This suggests that a larger capacity of LMs would facilitate the span detection process as well.

\mypara{Effects of Training Epochs}
We then investigate whether prompt tuning is still effective with fewer training epochs.
As shown in \autoref{figure:task2_fewer_epoch}, prompt tuning already achieves remarkable performance even in the first epoch.
For instance, with only 1 epoch on the T5-L model, prompt tuning can achieve 0.618 $F_1$-score, which is close to 0.643 with 5 epochs.
This further demonstrates the efficacy of prompt tuning in adapting to new tasks.

\mypara{Prompt Transferability}
As we only have one dataset for Task 2, to investigate the prompt transferability, we manually label the toxic spans of 100 randomly sampled posts from the Parallel dataset (we used in Task 3) and form a new testing dataset.
Given the prompt trained with T5-L on ToxicSpan, we observe that our method can correctly identify the toxic spans on 85\% of posts.
We then dive deeper into the failed cases and find that most of them belong to Categories 1 and 8 as shown in \autoref{table:task2_examples}.
In general, this case study demonstrates that prompt tuning can indeed transfer to out-of-distribution data.

\mypara{Comparison with Fine-tuning}
For Task 2, we also compare the performance of prompt tuning with fine-tuning.
Taking T5-L model as an example, we observe that, with the same training epochs, prompt tuning yields slightly better performance (0.643 $F_1$-score) than fine-tuning (0.628 $F_1$-score) and costs less time.
This indicates that prompt tuning can unleash the power of LLM with only limited effort.

\mypara{Robustness}
Following the perturbation strategy in Task 1, we perturb 100 randomly selected posts from TSD and compare the performance with the original posts.
We observe that prompt tuning reports the same toxic span for 57 perturbed posts.
For 38 perturbed posts, prompt tuning failed to detect or can only detect part of the toxic spans.
For the rest 5 perturbed posts, prompt tuning can obtain even better toxic spans than their original version.
Compared to Task 1, prompt tuning is less robust in Task 2.
This can be credited to the lack of perturbed toxic spans in the training dataset, which may be mitigated by introducing perturbation during the training phase as well.

\mypara{Error Analysis}
We conduct a case study regarding the wrongly detected spans.
Concretely, we randomly select 100 test samples with wrongly predicted spans and manually verify the possible reasons.
Then, we categorize the reasons into 9 categories (see \autoref{table:task2_examples}).
Note that each test sample is manually verified by three annotators to put into a category with full agreement.
We find that a substantial percentage of wrong span predictions in categories 2, 3, 4, and 5 (47\%) are caused by the problematic ground truth label.
For instance, in category 2, the ground truth span contains both toxic and non-toxic text.
Note that the ground truth inconsistency is caused by the fact that the lengths of the toxic spans were decided by the raters~\cite{PSLA21}.
The ToxicSpan dataset accepts character offsets that at least two raters have included each character offset in their spans.
Category 2 actually covers the corner cases relating to such human errors/bias when building the ToxicSpan dataset.
Nevertheless, our method successfully detects the real toxic span ``cowards'' from this example.
Also, in category 3, the toxic span is not labeled by the ground truth.
However, they are accurately detected by our method.
We also observe that prompt tuning may fail to identify some ambiguous toxic spans such as the ``embarrassment'' example shown in category 4 (\autoref{table:task2_examples}).
A more interesting case (category 5) shows that our method can dig out the missing toxic span from the text.
For instance, the ground truth span only contains ``stupid'', while our method discovers ``idiots'' as well.
This case demonstrates the potential of prompt tuning to become an effective tool to improve the annotation quality of toxic spans.
We also notice that the cases in categories 1, 6, 7, 8, and 9 (53\%) are caused (or partially caused) by our method.
For category 1, we observe that our method repeats the original sentence without any change.
We then diver deeper into those samples and find that they are mainly short sentences or contain less toxic spans, which may lead the prompt to become less sensitive to these cases.
For category 6, we observe that our method successfully generates the sentence without toxic spans, but the mapping algorithm fails to provide an exact span area as the ground truth span, e.g., prompt tuning includes the quota into the toxic span as well since it serves as an emphasize to the toxic expression.
In category 9, we observe that our method overlooks the ground truth span, but surprisingly detects a new span like the ``crap'' example.
Those wrong cases show that toxic span detection from the view of prompt tuning is not perfect, but prompt tuning shows its great potential in facilitating and correcting the toxic span detection process.
For instance, it can serve as an assistant tool for better annotation quality.

\begin{table*}[!ht]
\centering
\caption{Failed examples on Task 2.
\colorbox{emerald}{Green} denotes the ground truth span is correctly predicted by the algorithm. 
\colorbox{babypink}{Pink} denotes the ground truth span is not detected by the algorithm. 
\colorbox{darktangerine}{Orange} denotes the span is not in ground truth but is detected by the algorithm.}
\scalebox{0.9}{
\begin{tabular}{c p{4cm} p{8cm} c}
\toprule
\textbf{Category} & \textbf{Reason} & \textbf{Text Example} & \textbf{Percentage (\%)} \\
\midrule
1 & Labeled by ground truth (GT) but not by our method.  & they're not patriots. they're \colorbox{babypink}{vandals}, \colorbox{babypink}{thieves}, and \colorbox{babypink}{bullies}. they've plastered a facade of patriotism over their outrage at being expected to obey the law. &  17 \\
\midrule
2 & GT contains both toxic and non-toxic spans.   & adn is endorsing, without officially endorsing.  \colorbox{babypink}{bunch} \colorbox{babypink}{of} \colorbox{emerald}{cowards}!!!  &   9 \\
\midrule
3 & GT is none, but our method labels more toxic spans.    &  he's as \colorbox{darktangerine}{stupid} as those commie propagandists here who tried to attribute poor potato harvests to potato beetle supposedly being dropped from cia airplanes over gdr, czechoslovakia or poland. this was so \colorbox{darktangerine}{stupid} and out of sync with real world that it was subject of snickering among local populations. obviously you will not read that in books written by last marxists in the world, that is, western academics. &  15 \\
\midrule
4 & GT span is non-toxic.    & justin is an \colorbox{babypink}{embarrassment} to canada. he needs a muzzle. he needs a brain.  &  6 \\
\midrule
5 & GT only contains parts of toxic spans, and our method detect more.     &  the money you \colorbox{darktangerine}{idiots} give these people are why they are here. stop feeding the fire. unbelievable how \colorbox{emerald}{stupid} people can be....drops mic &  17 \\
\midrule
6 & Error caused by the matching algorithm & i'll ignore your \colorbox{darktangerine}{``}\colorbox{emerald}{stupid}\colorbox{darktangerine}{"} insult and reply anyway...   &  12 \\
\midrule
7 & Our method marks extra non-toxic span as toxic. & why don't you call yourself \colorbox{emerald}{dickhead} instead of \colorbox{babypink}{pubic} ... good grief.  &  12 \\
\midrule
8 & All GT are toxic, but our method ignores some of them.   & when you consider the source - he writes like the trump we've all come to know -  "i could stand in the middle of 5th avenue and shoot somebody and i wouldn't lose voters", a \colorbox{emerald}{racist}, \colorbox{babypink}{misgynistic}, liar who only brings hate to the table.  &  8 \\
\midrule
9 & GT is toxic, but our method instead finds other toxic.   & uh-no, keep voting for failed liberal \colorbox{babypink}{idiocy} that guarantees results ala detroit, chicago, etc.  you'll wish your body had only some \colorbox{darktangerine}{crap} rather than gangbanger gunfire.  &  4 \\
\bottomrule
\end{tabular}
}
\label{table:task2_examples}
\end{table*}

\mypara{Takeaways}
We observe that prompt tuning can achieve comparable performance with the best conventional method, i.e., SPAN-BERT, but with much less time cost.
Also, the performance is relatively stable even with fewer training epochs.
This further demonstrates the potential of leveraging prompt tuning to tackle the toxic span detection tasks and provides evidence for better span labeling.
We also show that prompt tuning, in some cases, can identify additional toxic spans not labeled by the ground truth (i.e., human annotators).

\section{Task 3: Detoxification}

Different from previous tasks that only focus on toxicity detection, this task aims to detoxify the given text while preserving the corresponding semantic meaning.

\subsection{Experimental Setup}

\mypara{Baselines}
We use the vanilla version of BART~\cite{LLGGMLSZ20} and the DetoxBART~\cite{LDUMDKSP22} as the baselines.
Note that the DetoxBART is also trained on the ParaDetox dataset for 10,000 epochs according to Logacheva et al.~\cite{LDUMDKSP22}.

\mypara{Datasets}
We use Parallel and ParaDetox datasets to evaluate the performance of baselines and prompt tuning.

\mypara{Metrics}
To quantify the quality of the detoxification, we consider two aspects, i.e., the detoxification effectiveness and the utility of the generated sentences.
For detoxification effectiveness, we leverage the Perspective API to quantify the toxicity level change since it offers the best performance among all baselines and is robust on different datasets.
Specifically, we first measure the average toxicity score change and then quantify the percentage of texts that has high toxicity score (0.7 or 0.9), following the guidelines of Perspective API.\footnote{\url{https://developers.perspectiveapi.com/s/about-the-api-score}.}
Note that we use $\mathbf{T_{avg}}$, $\mathbf{T_{0.7}}$, and $\mathbf{T_{0.9}}$ to denote the average toxicity score of texts, the ratio of texts that have toxicity score over 0.7, and the ratio of texts that has toxicity score over 0.9, respectively.
Regarding the utility, we consider five different metrics.
We first consider BLEU score as the utility evaluation metric, which is also widely used in previous work~\cite{SUM19,LDUMDKSP22}.
Then we quantify the semantic preservation by comparing the text embeddings similarity between the original text and the detoxification text.
Concretely, we consider two types of embedding following~\cite{LDUMDKSP22}, i.e., contextual string embeddings~\cite{ABV18} from flairNLP~\cite{flairNLP}, which is denoted as SIM (F),  and SIMILE proposed by Wieting et al. ~\cite{WBGN19}, which is denoted as SIM (W).
We denote the two types of embedding similarities as SIM (F) and SIM (W), respectively.
Besides, we also use the token-level perplexity~\cite{RWCLAS19} to measure the fluency of the text, where lower perplexity denotes better fluency.

\subsection{Results}

\begin{table*}[t]
\centering
\caption{Performance of Task 3. 
The arrow denotes which direction is for better results.}
\scalebox{0.9}{
\begin{tabular}{ll| ccc | cccc}
\toprule
\textbf{Dataset} & \textbf{Method} & $\mathbf{T_{avg}}$ $\downarrow$ & $\mathbf{T_{0.7}}$ $\downarrow$ & $\mathbf{T_{0.9}}$ $\downarrow$ &  \textbf{BLEU} $\uparrow$ & \textbf{SIM (W)} $\uparrow$ & \textbf{SIM (F)} $\uparrow$ & \textbf{TokenPPL} $\downarrow$ \\
\midrule
\multirow{7}{*}{\textbf{Parallel}}  & None & 0.755 & 0.676 & 0.135 & 1.000   & 1.000 & 1.000 & 227.834 \\
  & GroundTruth & 0.178 & 0.009 & 0.000 & 0.491 & 0.757   & 0.669 & 550.725 \\
  & BART  & 0.754 & 0.676 & 0.135 & 0.999   & 0.999 & 0.998     & 227.904 \\
  & DetoxBART  & 0.242 & 0.036 & 0.000 & 0.708   & 0.879 & 0.843 & 236.654 \\
  & PT (T5-S) & 0.573 & 0.463 & 0.077 & 0.835   & 0.927 & 0.939 & 326.696 \\
  & PT (T5-B) & 0.408 & 0.256 & 0.032 & 0.770   & 0.898 & 0.909 & 301.597 \\
  & PT (T5-L) & 0.396 & 0.329 & 0.031 & 0.754   & 0.881 & 0.889 & 284.861 \\
\midrule
\multirow{7}{*}{\textbf{ParaDetox}}  & None          & 0.775 & 0.778 & 0.134 & 1.000   & 1.000 & 1.000 & 330.829 \\
  & GroundTruth  & 0.166 & 0.000 & 0.000 & 0.633   & 0.828 & 0.778 & 393.800 \\
  & BART         & 0.774 & 0.777 & 0.133 & 0.999   & 0.999 & 0.998 & 331.250 \\
  & DetoxBART    & 0.180 & 0.013 & 0.000 & 0.688   & 0.862 & 0.832 & 438.242 \\
  & PT (T5-S) & 0.253 & 0.081 & 0.007 & 0.760  & 0.910 & 0.905 & 593.442  \\
  & PT (T5-B) & 0.224 & 0.051 & 0.005 & 0.754  & 0.920 & 0.897 & 499.851 \\
 & PT (T5-L) & 0.213 & 0.037 & 0.003 & 0.743  & 0.916 & 0.886 & 404.565 \\
\bottomrule
\end{tabular}
}
\label{table:task3_performance}
\end{table*}

The detoxification performance on different datasets is shown in \autoref{table:task3_performance}.
We observe that DetoxBART performs slightly better in detoxifying the text than prompt tuning.
For instance, on ParaDetox, DetoxBART reduces the $\mathbf{T_{avg}}$, $\mathbf{T_{0.7}}$, and $\mathbf{T_{0.9}}$ to 0.180, 0.013 and 0, respectively while prompt tuning on T5-L can reduce them into 0.213, 0.037, and 0.003 respectively.
This means that ParaDetox has better detoxification effectiveness than prompt tuning.
However, we also observe that the text quality generated with prompt tuning is better than the DetoxBART.
For instance, on ParaDetox, compared to DetoxBART, the PT (T5-B) has a higher BLEU score, SIM (W), SIM (F), while attaining a smaller TokenPPL.
This indicates the text generated by prompt tuning has better fluency and can better preserve the semantic meaning of the original text.
In general, we consider both DetoxBART and prompt tuning as successful methods as they can largely reduce the toxicity level while preserving the semantic meaning and fluency of the original text.

\begin{figure*}[ht]
\centering
\begin{subfigure}{0.2\linewidth}
\includegraphics[width=\columnwidth]{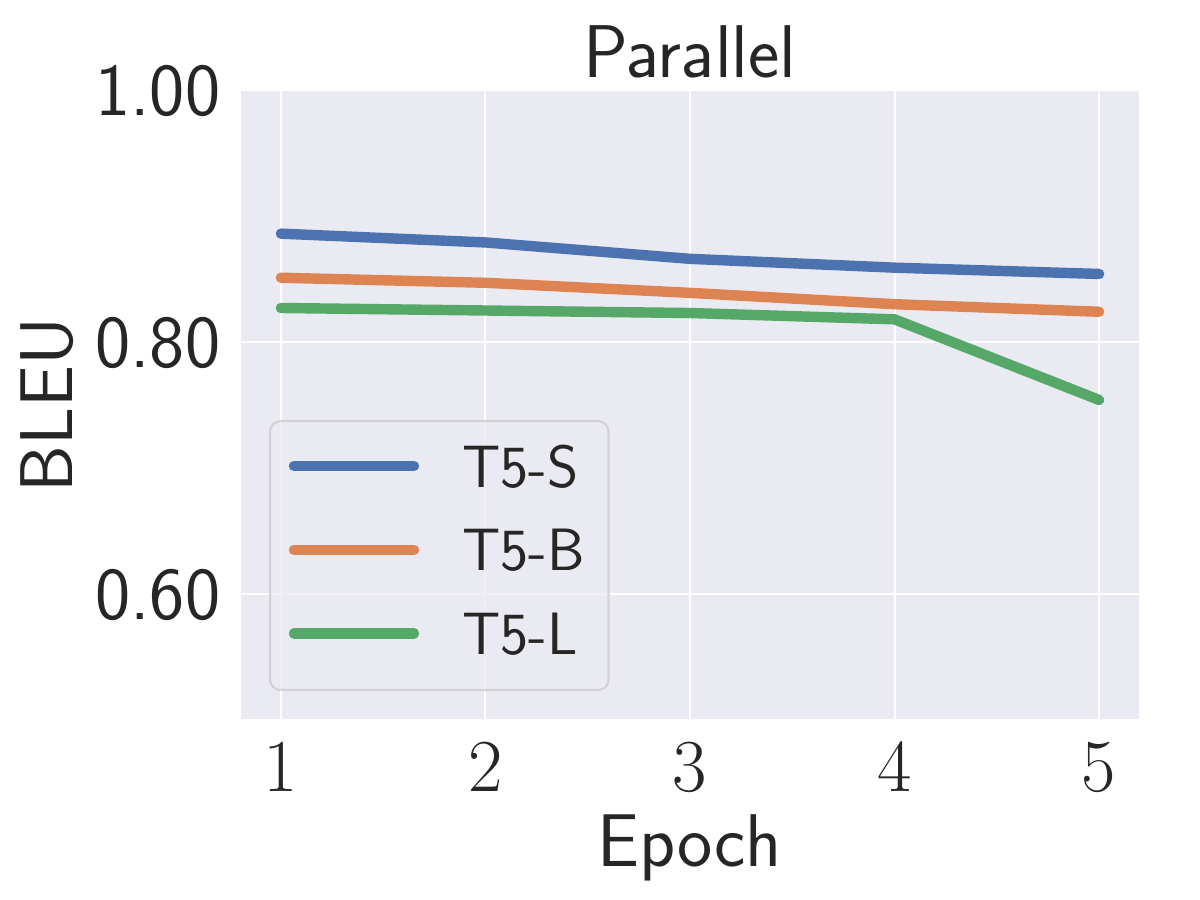}
\label{figure:task3_ablation_fewer_epoch_utility_bleu_Parallel}
\end{subfigure}
\begin{subfigure}{0.2\linewidth}
\includegraphics[width=\columnwidth]{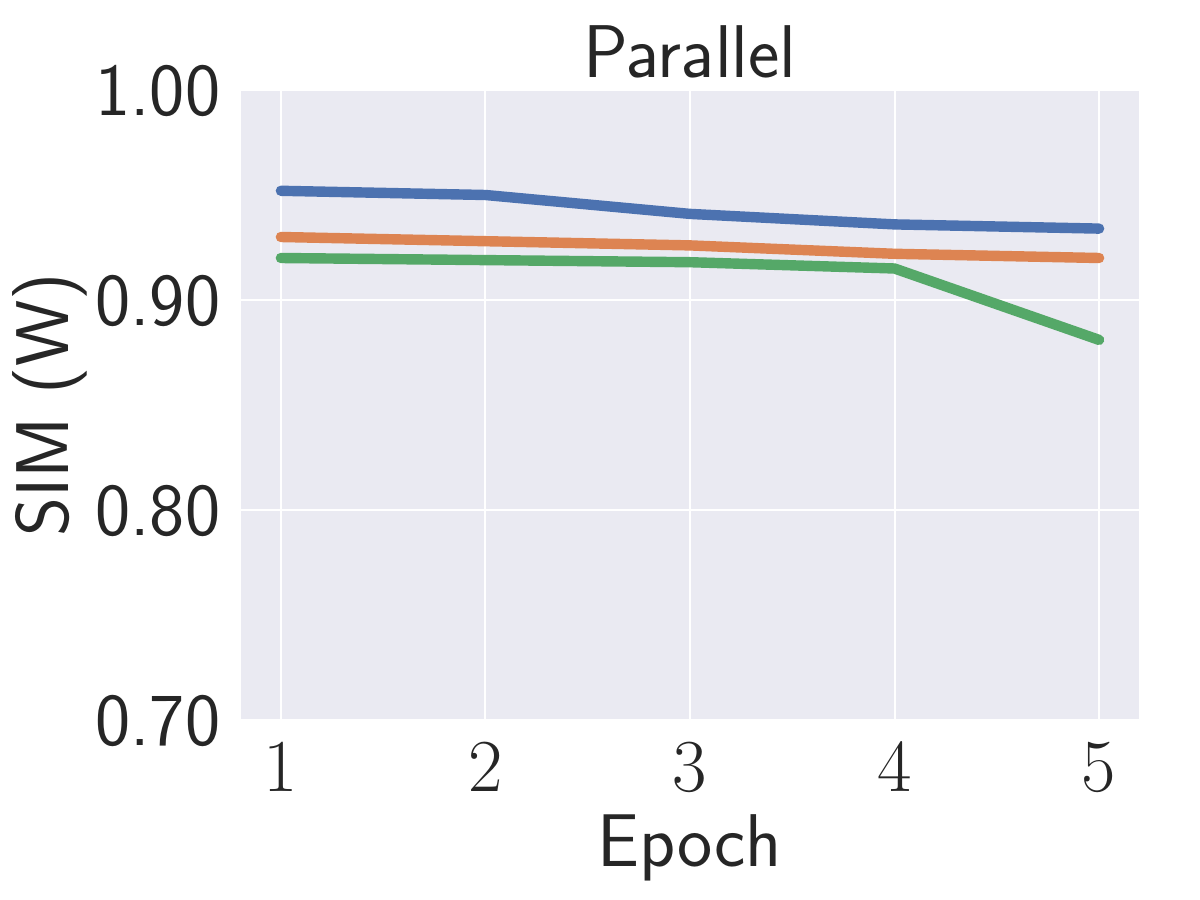}
\label{figure:task3_ablation_fewer_epoch_utility_sim1_Parallel}
\end{subfigure}
\begin{subfigure}{0.2\linewidth}
\includegraphics[width=\columnwidth]{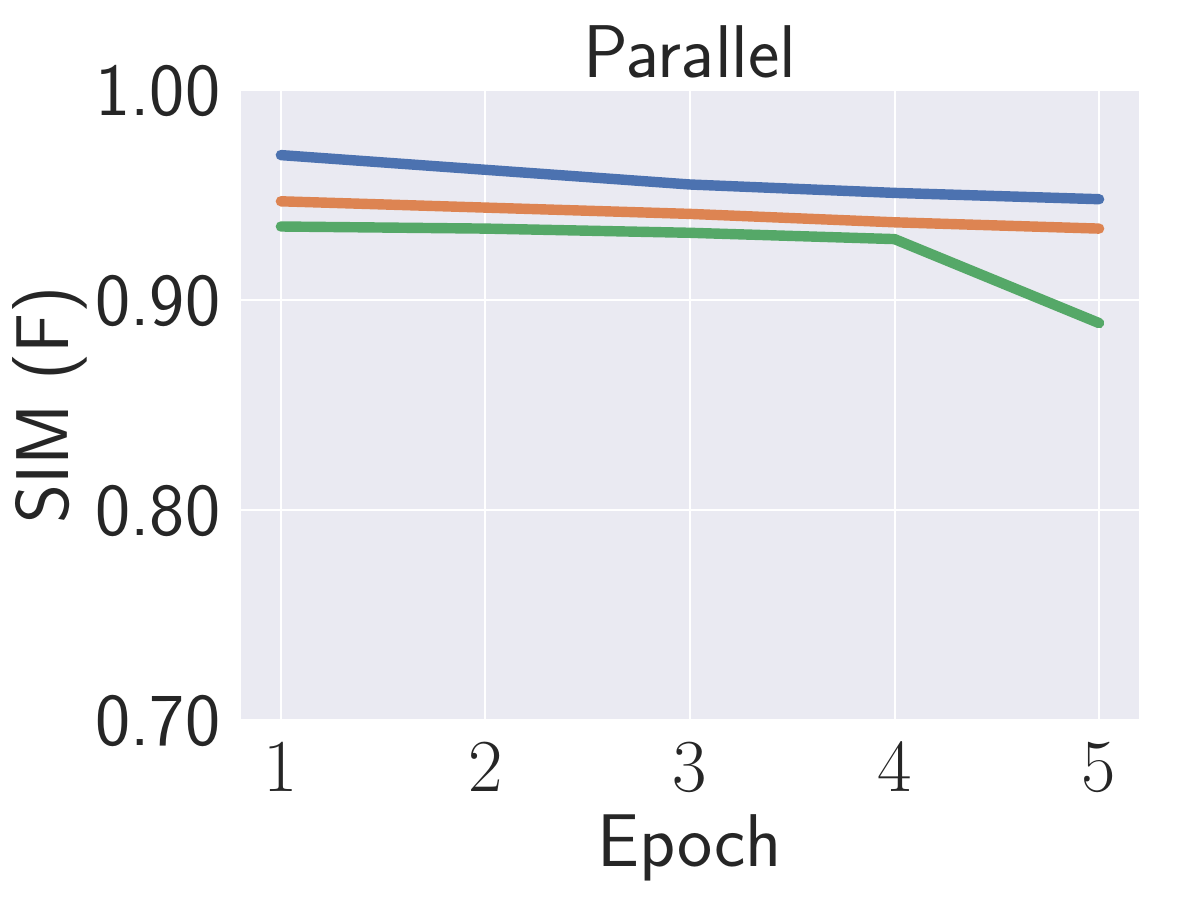}
\label{figure:task3_ablation_fewer_epoch_utility_sim2_Parallel}
\end{subfigure}
\begin{subfigure}{0.2\linewidth}
\includegraphics[width=\columnwidth]{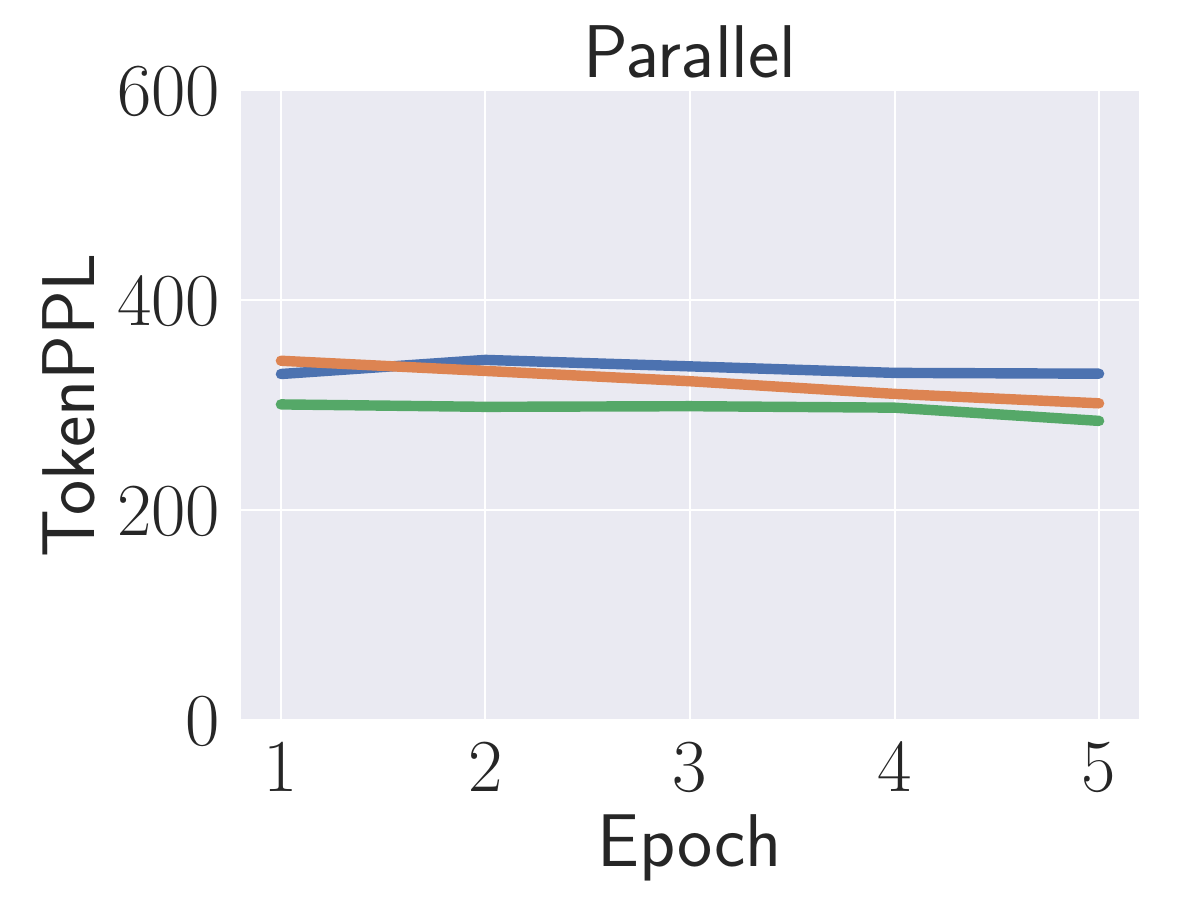}
\label{figure:task3_ablation_fewer_epoch_utility_ppl_Parallel}
\end{subfigure}
\begin{subfigure}{0.2\linewidth}
\includegraphics[width=\columnwidth]{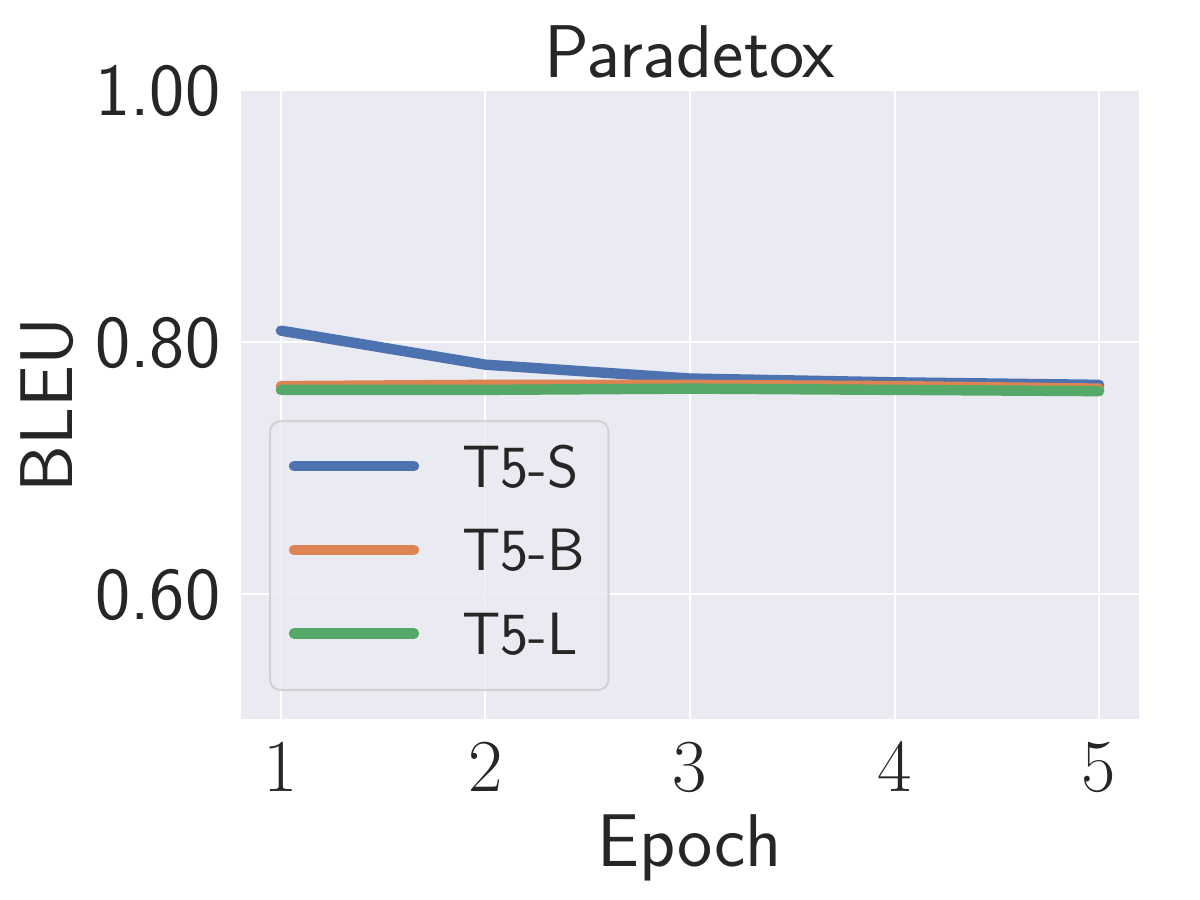}
\label{figure:task3_ablation_fewer_epoch_utility_bleu_Paradetox}
\end{subfigure}
\begin{subfigure}{0.2\linewidth}
\includegraphics[width=\columnwidth]{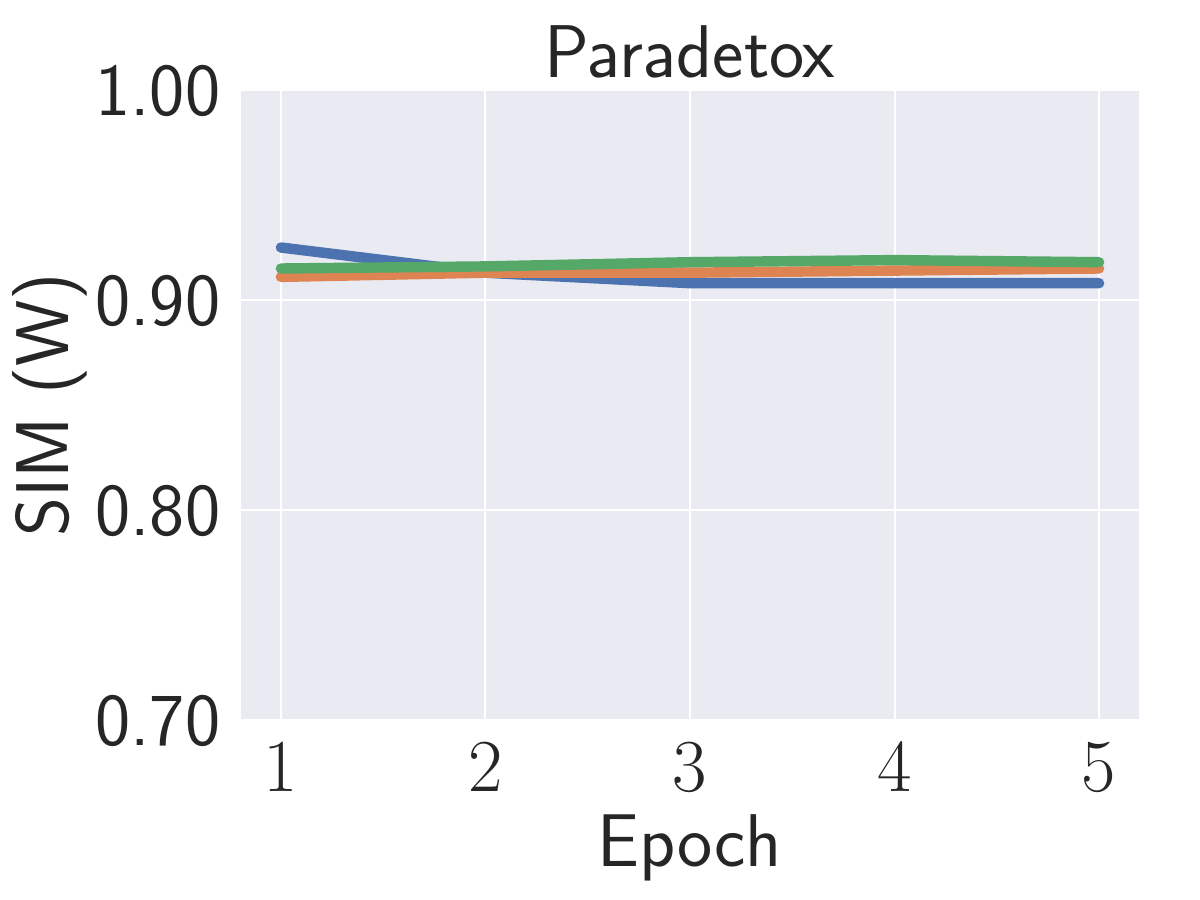}
\label{figure:task3_ablation_fewer_epoch_utility_sim1_Paradetox}
\end{subfigure}
\begin{subfigure}{0.2\linewidth}
\includegraphics[width=\columnwidth]{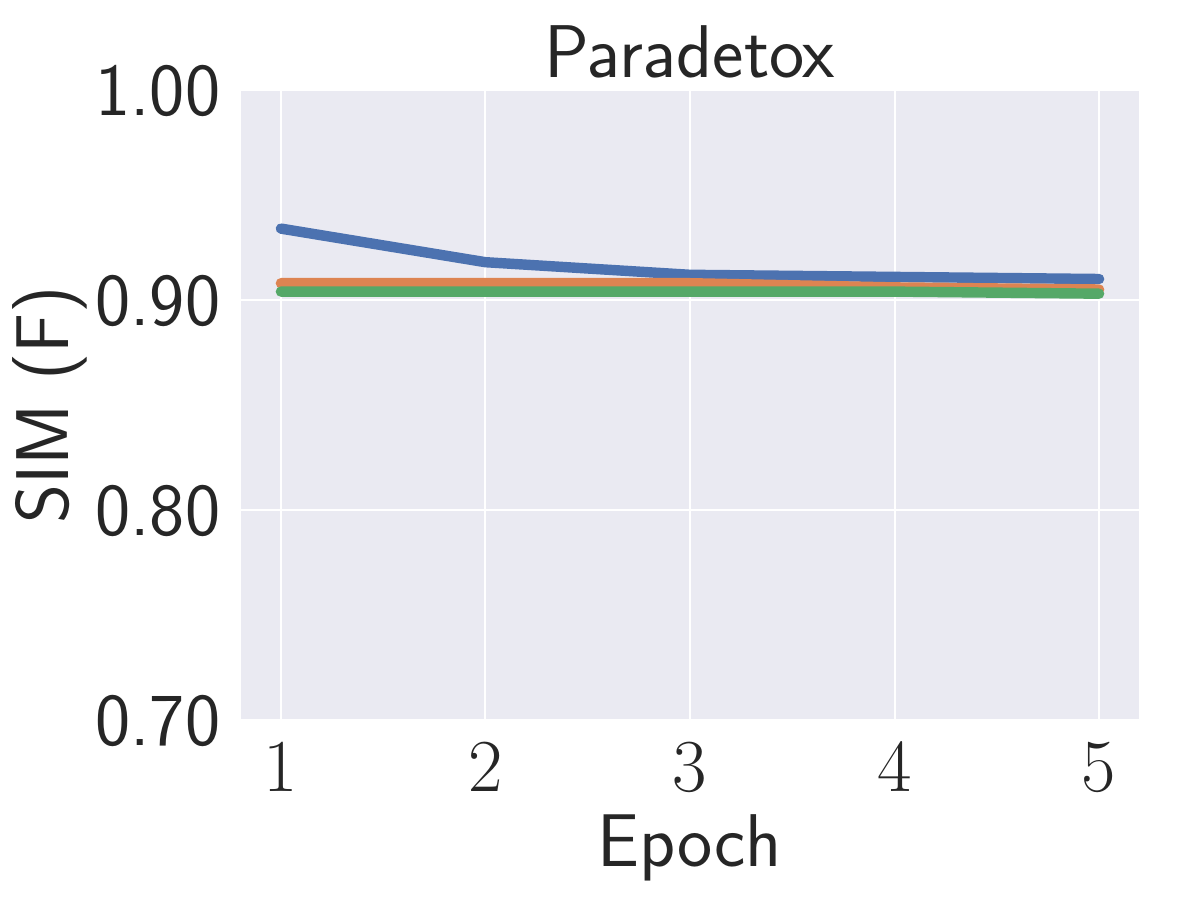}
\label{figure:task3_ablation_fewer_epoch_utility_sim2_Paradetox}
\end{subfigure}
\begin{subfigure}{0.2\linewidth}
\includegraphics[width=\columnwidth]{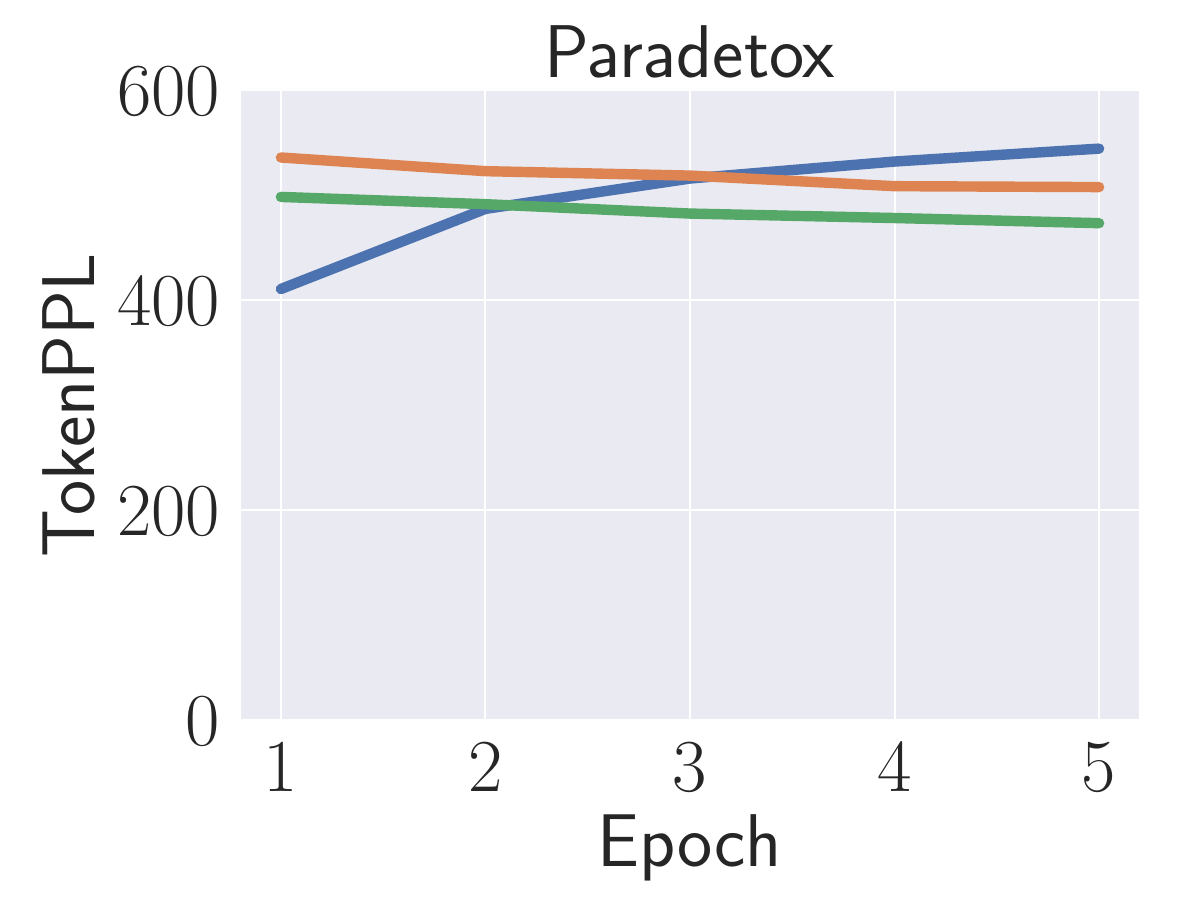}
\label{figure:task3_ablation_fewer_epoch_utility_ppl_Paradetox}
\end{subfigure}
\caption{Utility of Task 3 with different training epochs.}
\label{figure:task3_fewer_epochs_utility}
\end{figure*}

\begin{figure}[!t]
\centering
\begin{subfigure}{0.32\linewidth}
\includegraphics[width=\columnwidth]{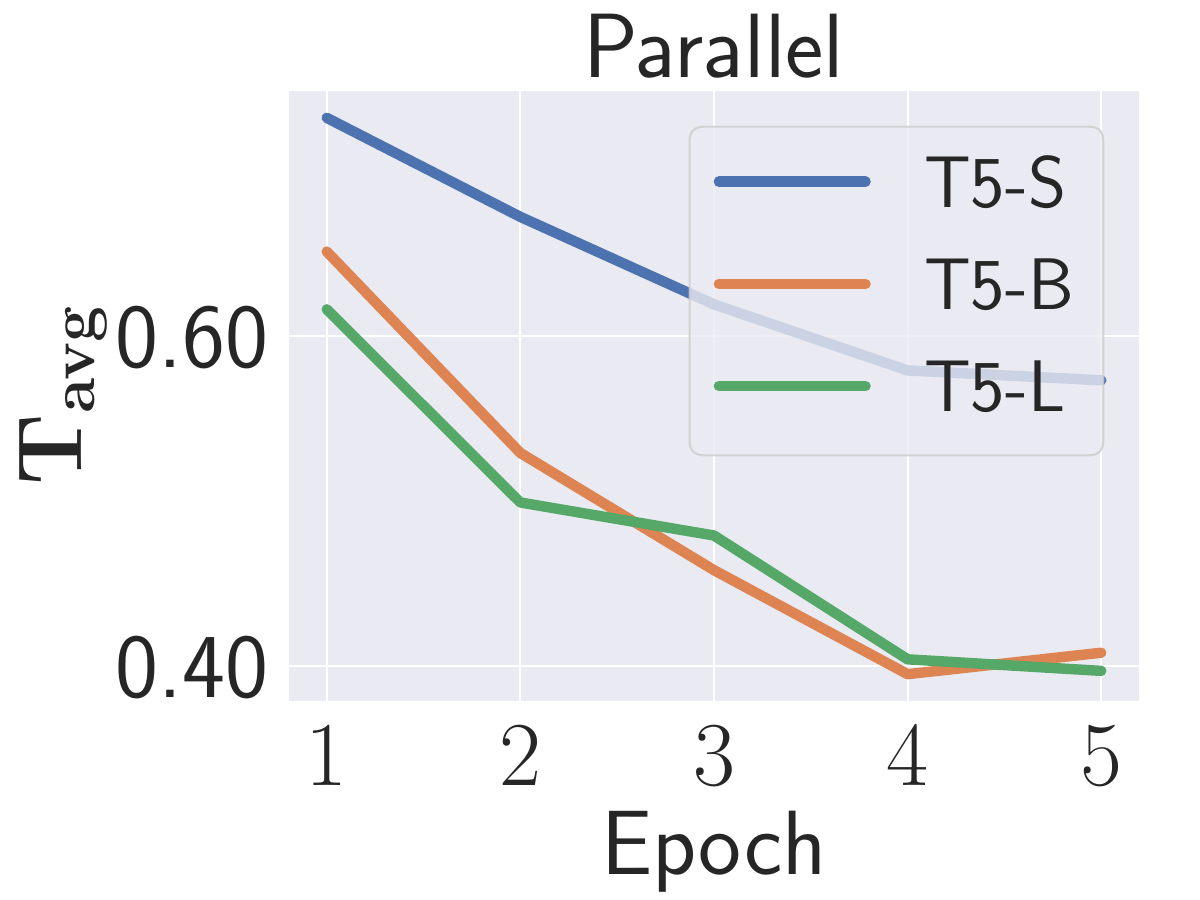}
\label{figure:task3_ablation_fewer_epoch_perspective_acc1_Parallel}
\end{subfigure}
\begin{subfigure}{0.32\linewidth}
\includegraphics[width=\columnwidth]{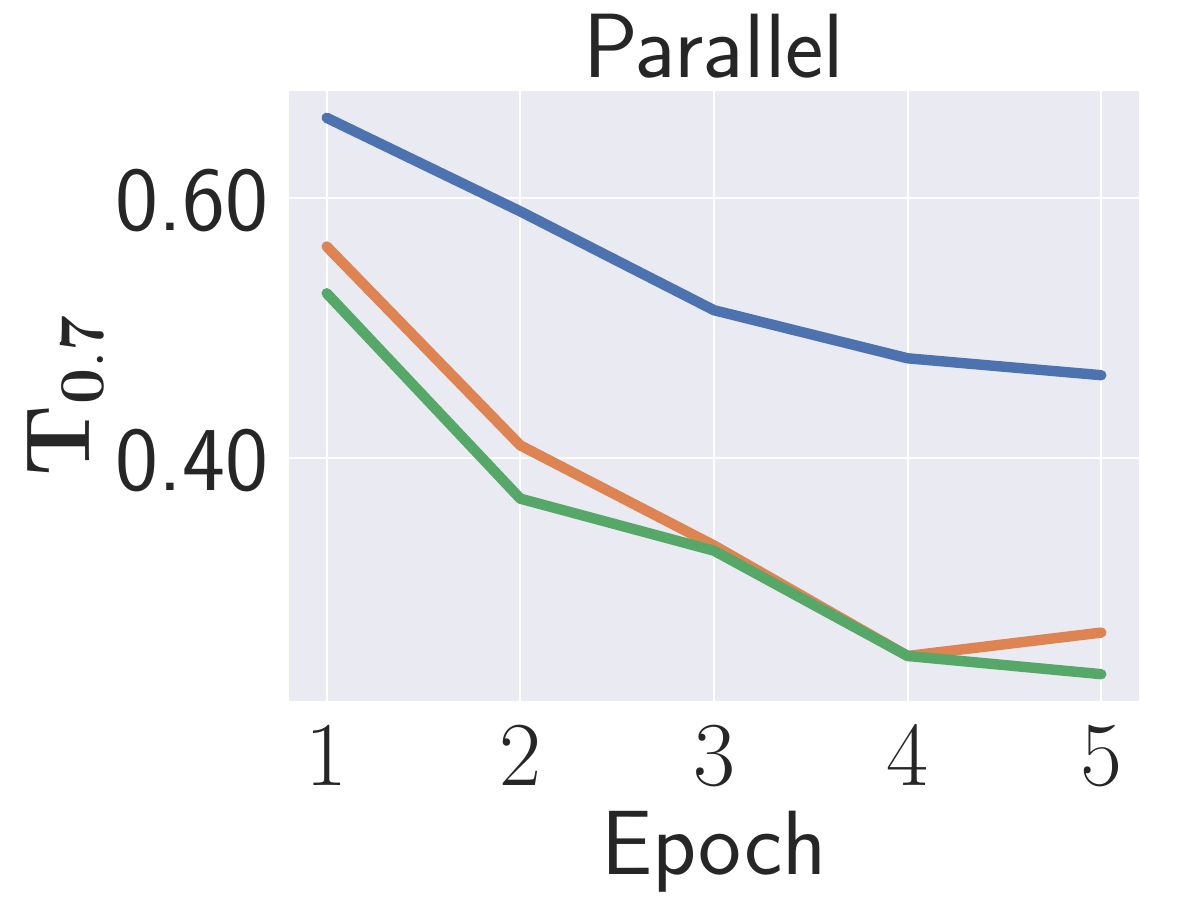}
\label{figure:task3_ablation_fewer_epoch_perspective_acc2_Parallel}
\end{subfigure}
\begin{subfigure}{0.32\linewidth}
\includegraphics[width=\columnwidth]{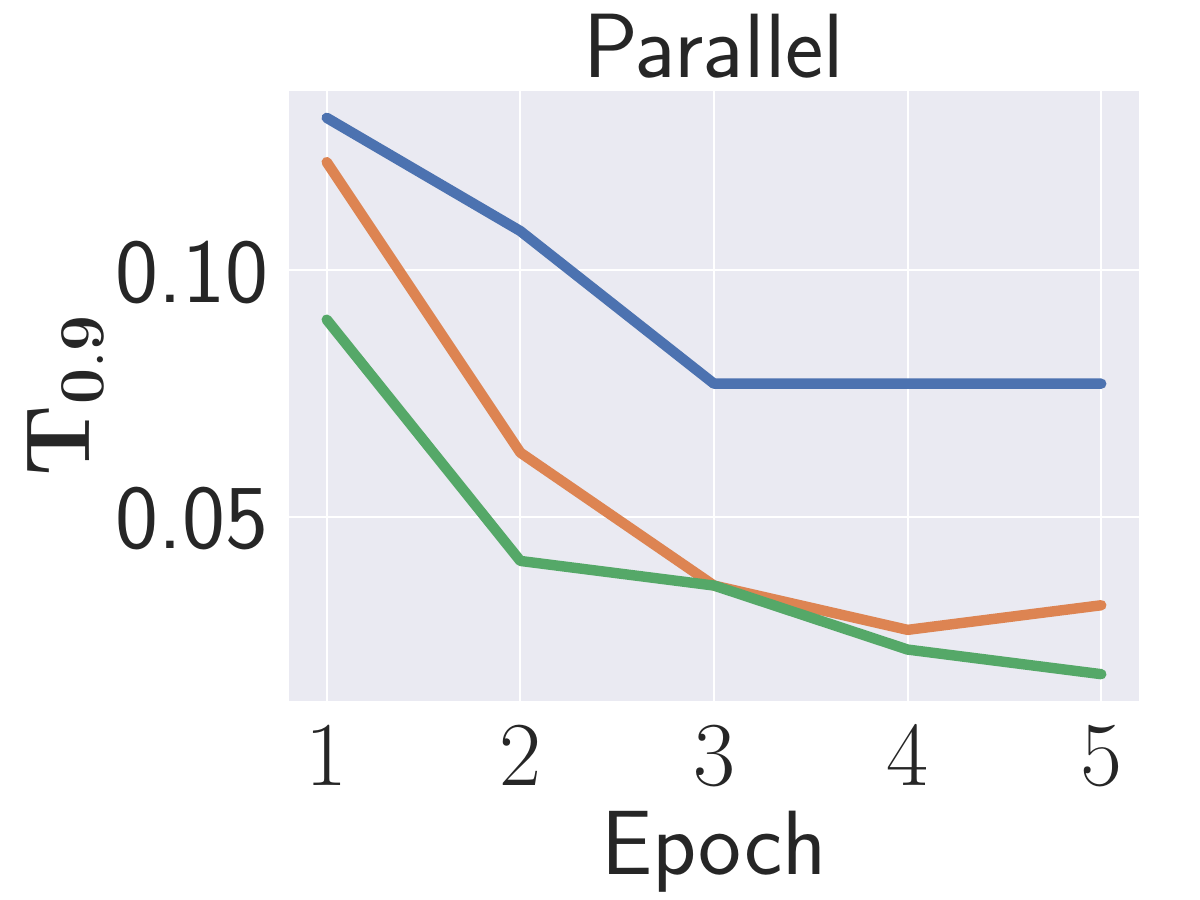}
\label{figure:task3_ablation_fewer_epoch_perspective_acc3_Parallel}
\end{subfigure}
\begin{subfigure}{0.32\linewidth}
\includegraphics[width=\columnwidth]{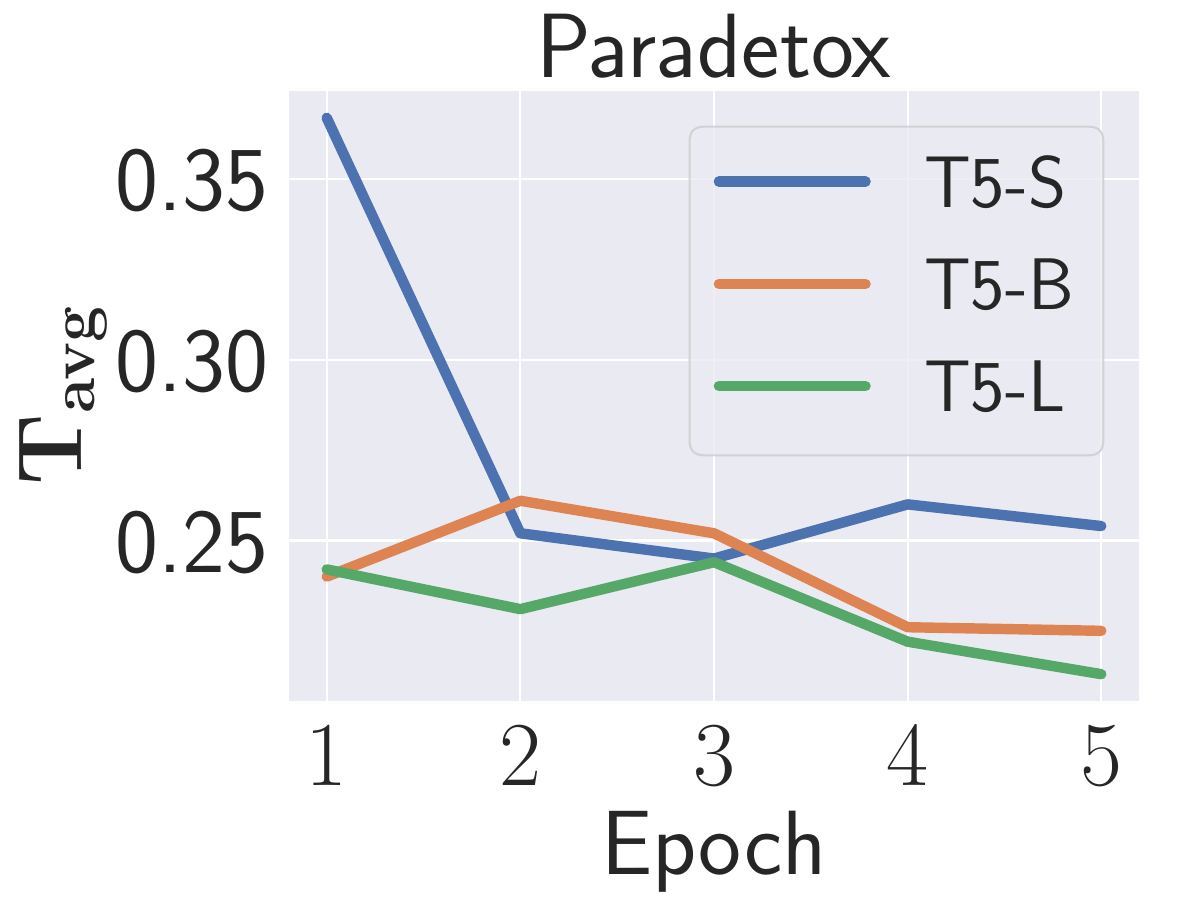}
\label{figure:task3_ablation_fewer_epoch_perspective_acc1_Paradetox}
\end{subfigure}
\begin{subfigure}{0.32\linewidth}
\includegraphics[width=\columnwidth]{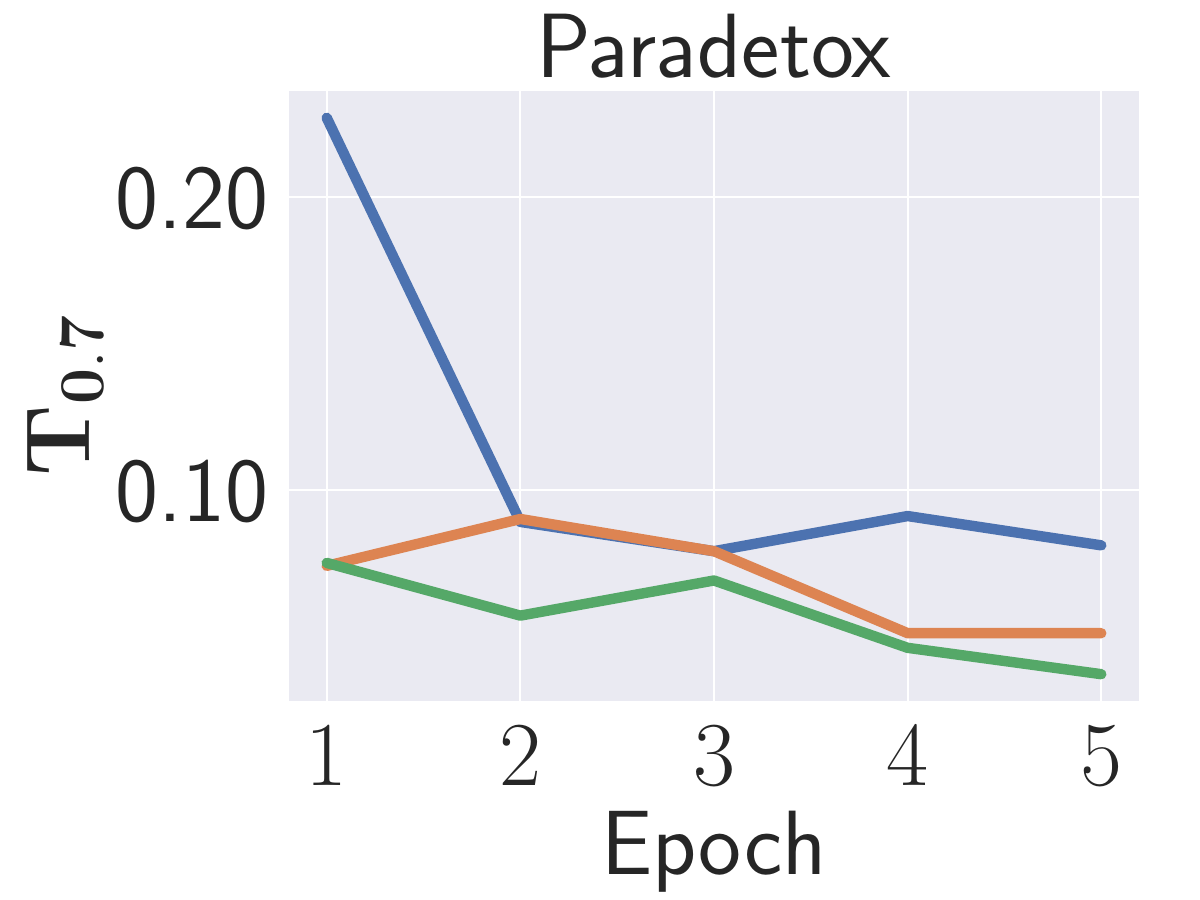}
\label{figure:task3_ablation_fewer_epoch_perspective_acc2_Paradetox}
\end{subfigure}
\begin{subfigure}{0.32\linewidth}
\includegraphics[width=\columnwidth]{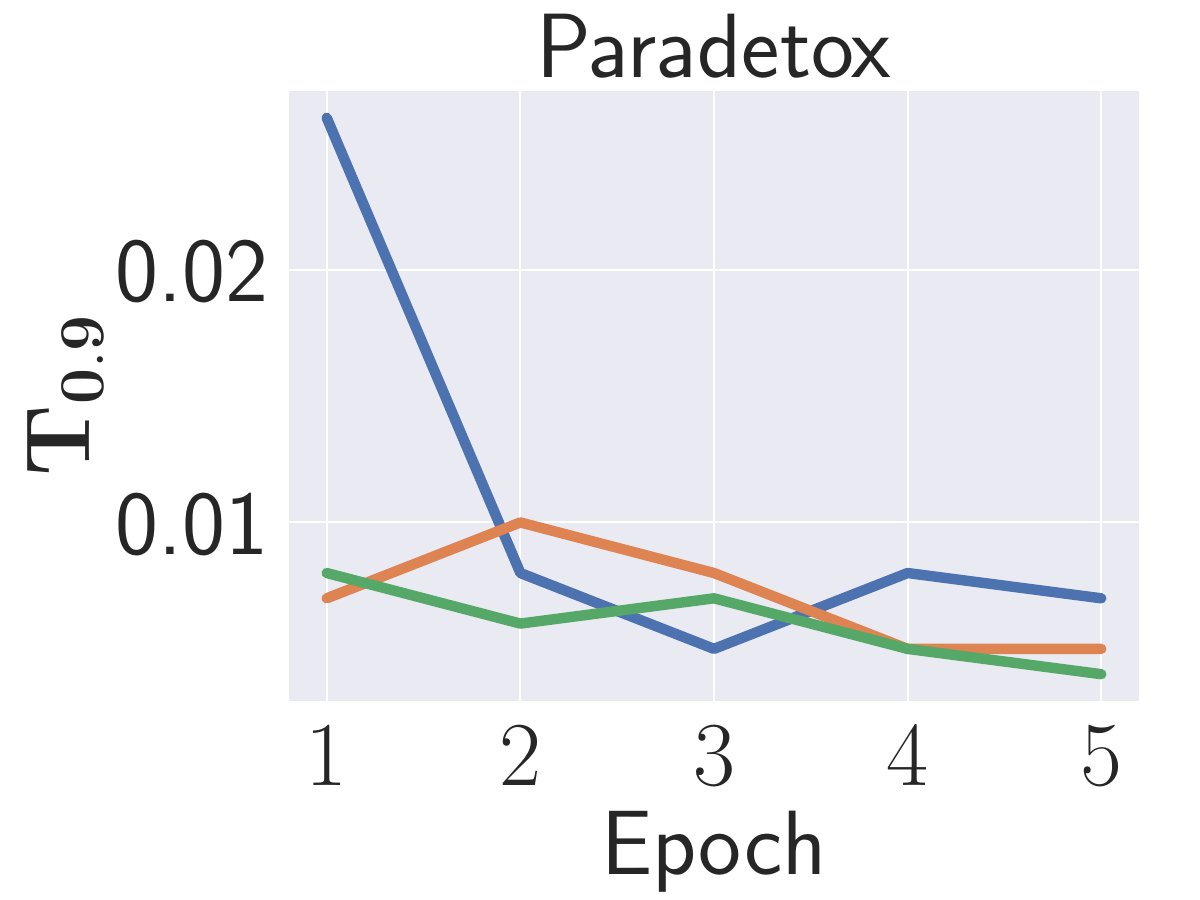}
\label{figure:task3_ablation_fewer_epoch_perspective_acc3_Paradetox}
\end{subfigure}
\caption{Detoxification effectiveness of Task 3 with different training epochs.}
\label{figure:task3_fewer_epochs_perspective}
\end{figure}

\mypara{Different Epochs}
We then investigate how the training epochs affect the detoxification effectiveness and the model's utility regarding semantic preservation.
The results are shown in \autoref{figure:task3_fewer_epochs_perspective} and \autoref{figure:task3_fewer_epochs_utility}, respectively.
From \autoref{figure:task3_fewer_epochs_perspective}, we have three observations, first, we find that more training epochs lead to better detoxification performance.
For instance, on Parallel, prompt tuning on T5-L can reduce the $\mathbf{T_{avg}}$ to 0.616 with 1 epoch, while decreasing to 0.397 with 5 epochs.
Second, prompt tuning on larger models lead to  better detoxification performance, e.g., T5-L performs the best while T5-S performs the worst.
This is expected as a larger model can represent the data in a more informative way thus better guiding the prompt tuning in the direction of detoxification.
Third, in a larger dataset such as Paradox, prompt tuning already achieves good detoxification performance in the early epoch, e.g., the first or second epoch.
Our results further exemplify the effectiveness of prompt tuning as the time cost is much less than training the detoxification model like DetoxBART.

Regarding utility, we find that the utility is relatively stable for different models in different epochs.
This indicates that those LLMs have good generation ability in general.

\mypara{Prompt Transferability}
We then take ParaDetox as the training dataset and Parallel as the testing dataset to investigate the generalizability power of prompt tuning.
With T5-B trained on ParaDetox, the $T_{avg}$, $T_{0.7}$, and $T_{0.9}$ on Parallel drop to 0.251, 0.027, and 0.000, respectively, which are even better than the original results shown in \autoref{table:task3_performance} (0.408, 0.256, and 0.032).
One possible reason is that ParaDetox contains a larger number of training data, which better guides the prompt for the detoxification tasks and makes it more transferrable to other datasets like Parallel.

\mypara{Comparison with Fine-tuning}
For Task 3, we take the T5-L model on Parallel as a case study.
We observe that, prompt tuning can reduce the toxicity of posts to a larger extent, e.g., the $\mathbf{T_{avg}}$ of prompt tuning is 0.396, while the value is 0.437 for fine-tuning.
On the other hand, we find that fine-tuning can generate more fluent sentences, e.g., the BLEU score is 0.795 for fine-tuning, while only 0.754 for prompt tuning.
In general, prompt tuning can still be considered as a lightweight plugin to adapt LLMs to new tasks.

\mypara{Robustness}
We again follow the perturbation strategy in Task 1 to perturb 100 randomly selected posts from the Parallel dataset.
We observe that, for the original version of these 100 posts, prompt tuning (with T5-L) can reduce the $T_{avg}$, $T_{0.7}$, and $T_{0.9}$ from 0.725, 0.590, and 0.130 to 0.357, 0.120, and 0.010, respectively, while the values are 0.402, 0.180, and 0.020 for the perturbed 100 posts, which is close to detoxify the original version.
This indicates that prompt tuning is relatively robust in Task 3.

\mypara{Case Study}
We then dive deeper into the generated text of the ParaDetox dataset and check them manually.
We consider both successful cases (C1 and C2) and failed cases (W1-W5).
\autoref{table:task3_examples} shows the examples of these cases.
In most cases, prompt tuning is powerful in reducing the toxicity level of the sentence while preserving its semantic meaning.
For example, in C1, our method achieves similar detoxification performance (toxicity score decreases from 0.827 to around 0.163).
Also, our method preserves the semantic meaning properly.
In C2, we observe that our method can even detoxify the sentence better than the ground truth.

Among the 2,388 text samples, we observe that there are 88 detoxification samples (3.68\%) that still have a high toxicity score, i.e., larger than 0.7.
We manually check those samples and find that they can be categorized into 5 different wrong categories (W1-W5).
For W1 (6/88), we observe that the sentence is hard to be detoxified, and the ground truth sentence is identical to the original sentence.
For W2 (52/88), prompt tuning just directly repeats the original sentence without any modification.
For W3 (27/88), we observe that prompt tuning indeed preserves the semantic meaning and reduces the toxicity level.
We acknowledge that for some implicit toxic content, as shown in the example, it might be harder for the prompt model to detect and eliminate them perfectly.
For W4 (1/88), prompt tuning actually provides better semantic preservation compared to the ground truth.
For W5 (1/88), we observe that prompt tuning just considers ``i jus clicked tht nasty shit'' as toxic parts and directly removes them.
During the labeling, we notice that there indeed exists a trade-off between detoxification and semantic preservation.
However, in most cases, prompt tuning can do well on both aspects (see also \autoref{table:task3_performance}).
It indicates that prompt tuning can be a good tool for assisting the detoxification task, e.g., providing possible solutions for the annotators to make their decision.
Currently, our current prompt tuning is based on paired datasets, which is similar to machine translation.
However, such datasets are usually small.
One promising direction that we aim to explore in our future work is to combine the paired dataset with the unpaired dataset (i.e., it only contains sets of toxic and non-toxic contents but without the pairs) to jointly fine-tune the prompt.

\begin{table*}[ht]
\centering
\caption{Examples on Task 3. 
Here we consider successfully detoxified cases (with toxicity score less than 0.7) as the correct cases (C), and those cases with a larger than 0.7 toxicity score are considered as wrong cases (W).}
\scalebox{0.9}{
\begin{tabular}{c p{4.8cm} p{4.8cm} p{4.8cm}}
\toprule
\textbf{Category} & \textbf{Original} & \textbf{Ground Truth} & \textbf{Generated}  \\
\midrule
C1 & if assad wasn 't such a dick then involvement would not be necessary. [0.827]  & if assad wasn't such that bad involvement would not be necessary. [0.163]  &  if assad wasn 't such a bad person then involvement would not be necessary. [0.159] \\
\midrule
C2 & your reading comprehension is more fucked up than a football bat. [0.835] & your reading comprehension is worse than a football bat. [0.575] & your reading comprehension is not good [0.355] \\
\midrule
W1 & if you hold a knife against the cops you deserve to be shot! [0.827]  & if you hold a knife against the cops you deserve to be shot! [0.827]  & if you hold a knife against the cops you deserve to be shot ! [0.774] \\
\midrule
W2 & did they bring along a signedxglossy of boy wonder and castro as a token of their stupidity ? [0.770] & did they bring along a signedxglossy of boy wonder and Castro as a token [0.133] & did they bring along a signedxglossy of boy wonder and castro as a token of their stupidity? [0.754] \\
\midrule
W3 & someone should give this idiot a slap on the head. [0.939] & someone should give him a lesson. [0.257] & someone should give him a slap on the head. [0.721] \\
\midrule
W4 & i hate guys with a bitch mouth sit the fuck down nigga [0.985] & I hate bad guys [0.440] & i hate guys with a bad mouth sit down [0.721] \\
\midrule
W5 & wtf ! ! ! ! ! ! ! ! ! ! i jus clicked tht nasty shit [0.979] & I just clicked something. [0.042] & wtf!!!!!!!!!!!!!!!!!!!!!!!!!!! [0.713] \\
\bottomrule
\end{tabular}
}
\label{table:task3_examples}
\end{table*}

\mypara{Takeaways}
We empirically show that prompt tuning can reduce the toxicity level to a large extent and better preserve the semantic meanings.
An interesting observation is that the semantic meaning of the original sentence can be properly preserved even with fewer training epochs due to the strong representation ability of the LLM.
However, with fewer epochs, the detoxification performance might be less satisfying as the process of toxic to non-toxic contents is more difficult than previous tasks and needs more learning steps to better guide the prompt tuning.
The effective detoxification and semantic preserving abilities make prompt tuning a strong competitor to conventional methods in the detoxification task.

\section{Related Work}

\mypara{Prompt Learning}
Prompt learning is a new paradigm in natural language processing (NLP)~\cite{LYFJHN21}.
It allows users to directly specify the task they want in natural language for the pre-trained language model to interpret and complete.
This paradigm paves way for using a single LLM as the \emph{universal solver} for various understanding and generation tasks, such as text classification~\cite{SS21}, machine translation~\cite{RSRLNMZLL20}, semantic parsing~\cite{SLTCRPPKED21}, question answering~\cite{JXAN20}, etc.
To unleash the full potential, research on prompt learning has been investigating automatically inducing the discrete/continuous prompts~\cite{LL21,TMCEVH21}, multi-prompt learning~\cite{QE21,JXAN20}, prompt training, and fine-tuning strategy~\cite{PRS19,DISFHS20}, transferability of prompts~\cite{PKC21}, etc.
Our work is built on top of prompt learning.
We conduct the first systematic hateful language study from the prompt tuning perspective.

\mypara{Toxicity Classification}
The problem of toxic online content is a longstanding and challenging~\cite{APP19} problem affecting our society.
Motivated by the impact that the problem has on both the online and offline world, the research community and the industry devoted substantial resources to developing models to detect toxic content.
One of the most used tools for assessing toxicity online is Perspective API~\cite{Perspective}, a set of machine learning models trained on a human-annotated dataset, released by Google.
The Perspective API, given a piece of text, provides a set of scores that correspond to how likely the text is toxic, attacking specific identities, sexually explicit, etc.
At the same time, Google released its annotated dataset, which enabled other researchers to develop more models aiming to tackle the problem.
One such example is Detoxify~\cite{Detoxify}, which leverages the power of transformer models to detect toxicity in text, across multiple languages.

Davidson et al.~\cite{DWMW17} highlight that there is a distinction between offensive language and hate speech.
Also, the authors release HateSonar, a machine learning model, that identifies whether a piece of text contains offensive language or hate speech.
As previous research notes~\cite{ZEBNS20}, however, the HateSonar classifier performs poorly compared to the Perspective API, when tested on comments left on news articles.
Zimmerman et al.~\cite{ZKF18} highlight that by leveraging deep learning ensembles, we can improve the performance of previous models in detecting hate speech on Twitter.
Other work focuses on identifying the targets of toxic content~\cite{SMCBW16,ENNVB18}, or on identifying specific forms of toxic content such as Antisemitism~\cite{ZFBB20,OWBLM21}, Islamophobia~\cite{VY18}, and Sinophobia~\cite{TSLBSZZ21,ZHSK21}.

All of the above-mentioned efforts in detecting toxic content are based on fine-tuning existing models or developing dedicated classifiers focusing on the specific task of detecting toxic content.
Recently, the pre-train and prompt paradigm is becoming increasingly popular, hence the research community started investigating how prompt learning can be leveraged to tackle the problem of toxic content online.
In particular, Chiu et al.~\cite{CA21} use OpenAI's GPT-3 language model to investigate the performance of prompt learning in the task of detecting racist or sexist content.
They find that by using a pre-defined prompt and a few-shot learning setting, they can identify racist or sexist content with an accuracy of up to 85\%, highlighting that prompt learning can play a role in identifying toxic content.
Similarly, Schick et al.~\cite{SUS21} find that language models can identify toxic content and whether the generated text contains undesirable biases, all using prompt learning techniques.
Also, they propose a de-biasing method, which helps the language model generate less biased content.
Overall, both works~\cite{SUS21,CA21} highlight that large language models and prompt learning can detect toxic content with a decent performance.
While this previous work is essential, it is limited in the sense that it focuses only on the toxicity classification task and, more importantly, relies on manual pre-defined prompts.
In contrast, our work provides a comprehensive evaluation of how large language models and prompt learning can assist in tackling the problem of toxic content by considering multiple tasks (toxicity classification, toxic span detection, and detoxification).
Also, we show that by using prompt tuning techniques, instead of pre-defined prompts, we can substantially increase the performance of the language models in the three tasks.

\mypara{Toxic Span Detection} 
Toxic span detection~\cite{PSLA21} aims to identify the specific span that makes the sentence to be toxic.
Pavlopoulos et al.~\cite{PLXSA22} treat this task as the sequence labeling task to annotate the suspicious span in the sentence.
Three models including  BiLSTM~\cite{HS97}, BERT~\cite{DCLT19}, and SPAN-BERT~\cite{JCLWZL20} are considered.
We instead formalize this task as a generation task and show that prompt-tuning can achieve comparable performance to the SPAN-BERT but with much less computational time.

\mypara{Detoxification} 
Detoxification aims to reduce the toxicity level of the sentence while preserving the semantic meaning to the largest extent.
It is similar to neural style transfer~\cite{JJHVM22}.
Laugier et al.~\cite{LPSD21} propose a self-supervised method named CAE-T5 to learn the transformation from toxic to civil from the unpaired corpus.
Logacheva et al.~\cite{LDUMDKSP22} develop DetoxBART which fine-tunes the BART model on the ParaDetox dataset to achieve better performance.
Our work is substantially different from their work as we do not need to fine-tune the model but just the prompt, which is less costly.
We notice that conventional methods like DetoxBART can achieve better detoxification performance while prompt tuning can better preserve semantic information.

\section{Conclusion}

In this paper, we performed the first extensive evaluation of using prompt learning with tunable prompts (prompt tuning) to tackle the problem of toxic content.
Particularly, we focused on three tasks (toxicity classification, toxic span detection, detoxification) and assessed the performance of prompt tuning and how it compares to state-of-the-art baselines in these tasks.
Among other things, we find that prompt tuning can achieve comparable or even better performance compared to the baselines.
As shown by our evaluation, integrating prompt tuning into toxic content research can better help to improve the dataset quality and the model utility as the toxicity label (Task 1), predicted toxic span (Task 2), and detoxified sentence (Task 3) can be used to assist the labeling procedure.

\mypara{Limitations}
Naturally, our work has some limitations. First, we use GPT2 and T5 as the LLMs to demonstrate the efficacy of prompt tuning.
Our evaluation has demonstrated that prompt tuning can perform well even with these LLMs, and larger models generally perform better (see \autoref{table:task1_performance_f1}).
While we acknowledge that conducting experiments with larger models with billions of parameters would be appealing, our hardware capabilities limit such endeavors.
Also, we use Perspective API as an indicator to quantify the toxicity level (e.g., on Task 3), which is likely to yield some false positives/false negatives.
Nevertheless, detecting toxic content is an open research challenge and the Perspective API is also leveraged by previous work~\cite{SUS21,SHBBZZ22}, indicating that it is a good proxy for assessing toxic content.
Despite these limitations, we believe that our research can pave new ways for the study of toxic content, as researchers with limited computing resources can utilize currently available pre-trained large language models to perform important toxicity-related tasks with acceptable performance.

\mypara{Acknowledgment}
We thank the anonymous reviewers and our shepherd for their invaluable comments and feedback that helped us improve our manuscript.
This work is partially funded by the European Health and Digital Executive Agency (HADEA) within the project ``Understanding the individual host response against Hepatitis D Virus to develop a personalized approach for the management of hepatitis D'' (D-Solve) (grant agreement number 101057917).

\begin{small}
\bibliographystyle{plain}
\bibliography{normal_generated_py3}    
\end{small}

\appendix

\section{Task 1 Performance with Other Metrics}
\label{appendix:task1_performance_other_metrics}

\mypara{General Result}
\autoref{table:task1_performance_acc}, \autoref{table:task1_performance_precision}, and \autoref{table:task1_performance_recall} show the accuracy, precision, and recall of Task 1.

\mypara{Manual Prompt}
\autoref{table:task1_ablation_manual_prompt_accuracy_pr} shows the accuracy, precision, and recall of Task 1 with the manual prompt.

\begin{figure*}[!ht]
\centering
\begin{subfigure}{0.18\linewidth}
\includegraphics[width=\columnwidth]{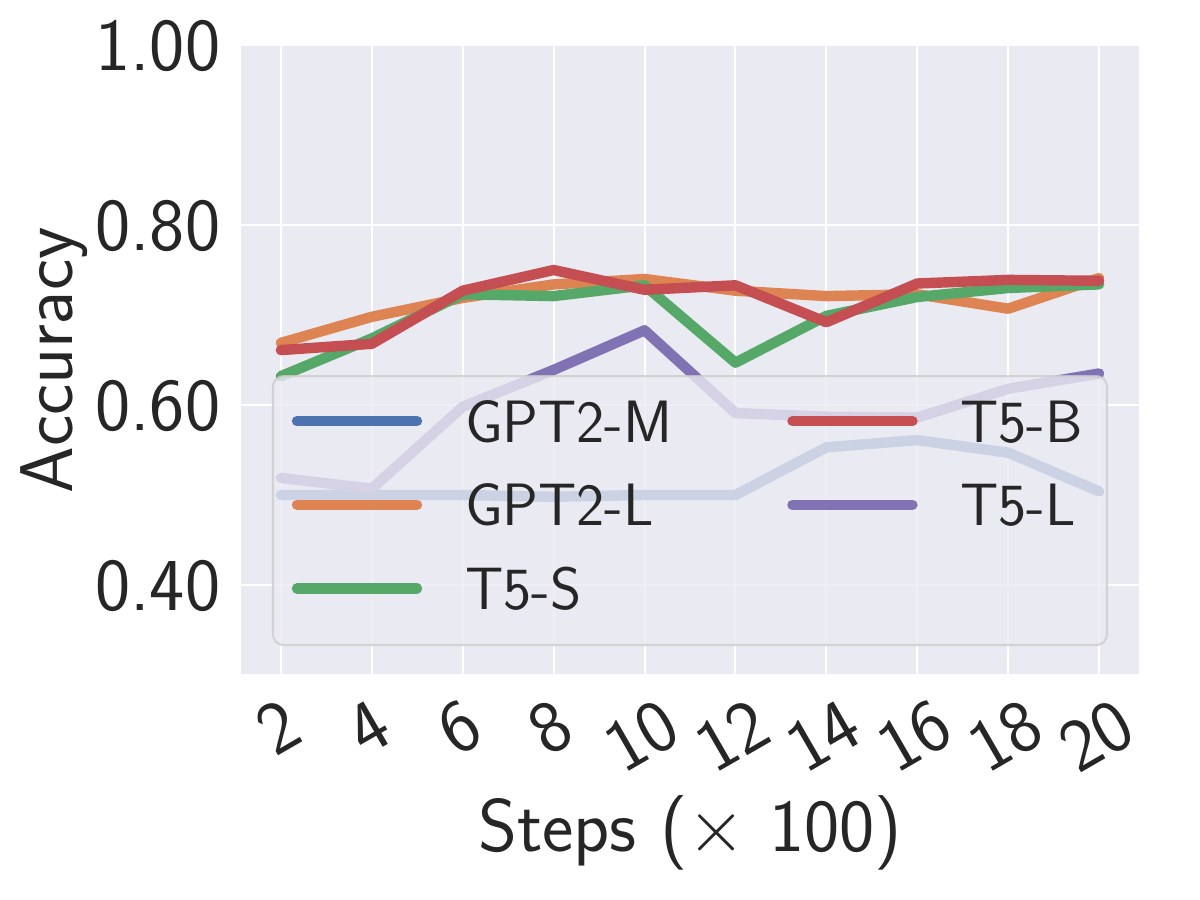}
\caption{HateXplain}
\label{figure:ablation_fewer_steps_HateXplain_test_acc}
\end{subfigure}
\begin{subfigure}{0.18\linewidth}
\includegraphics[width=\columnwidth]{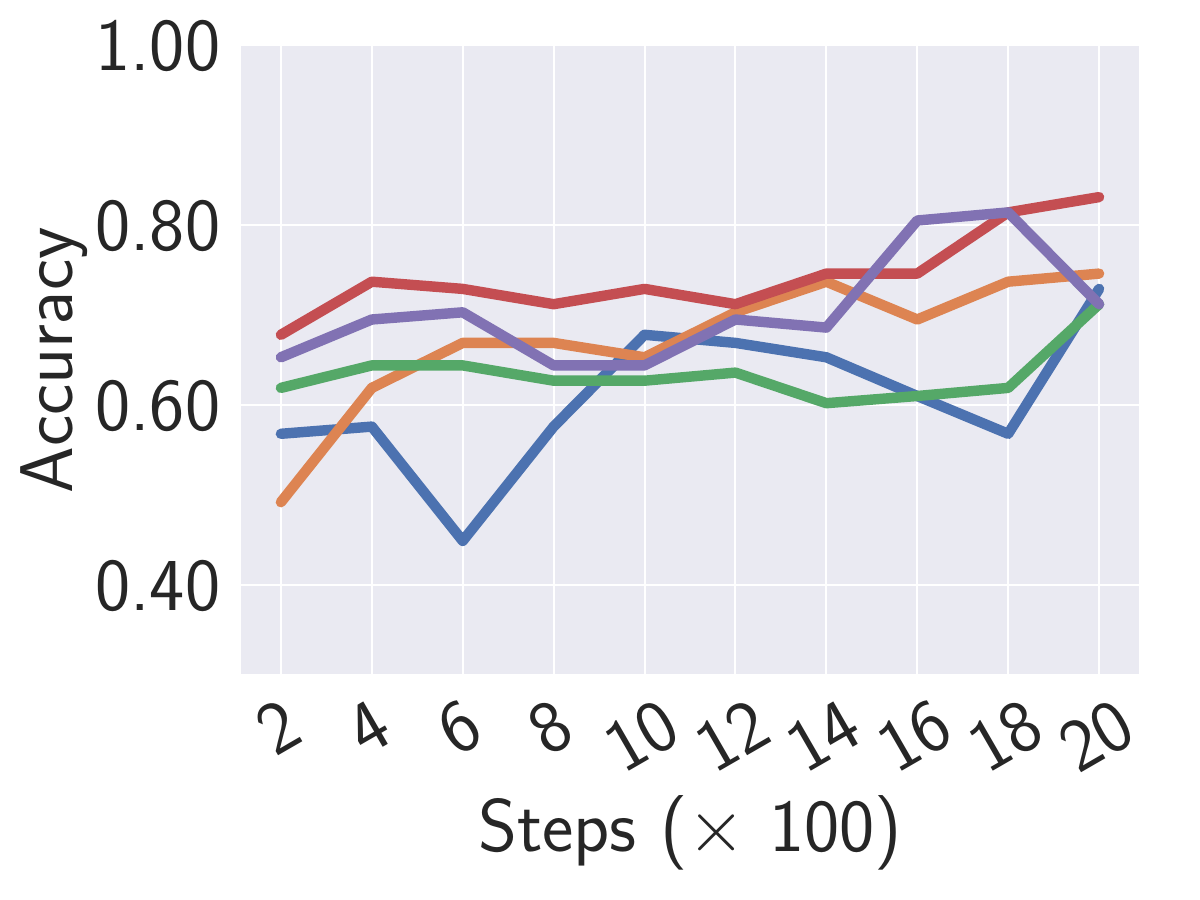}
\caption{USElectionHate20}
\label{figure:ablation_fewer_steps_USElectionHate20_test_acc}
\end{subfigure}
\begin{subfigure}{0.18\linewidth}
\includegraphics[width=\columnwidth]{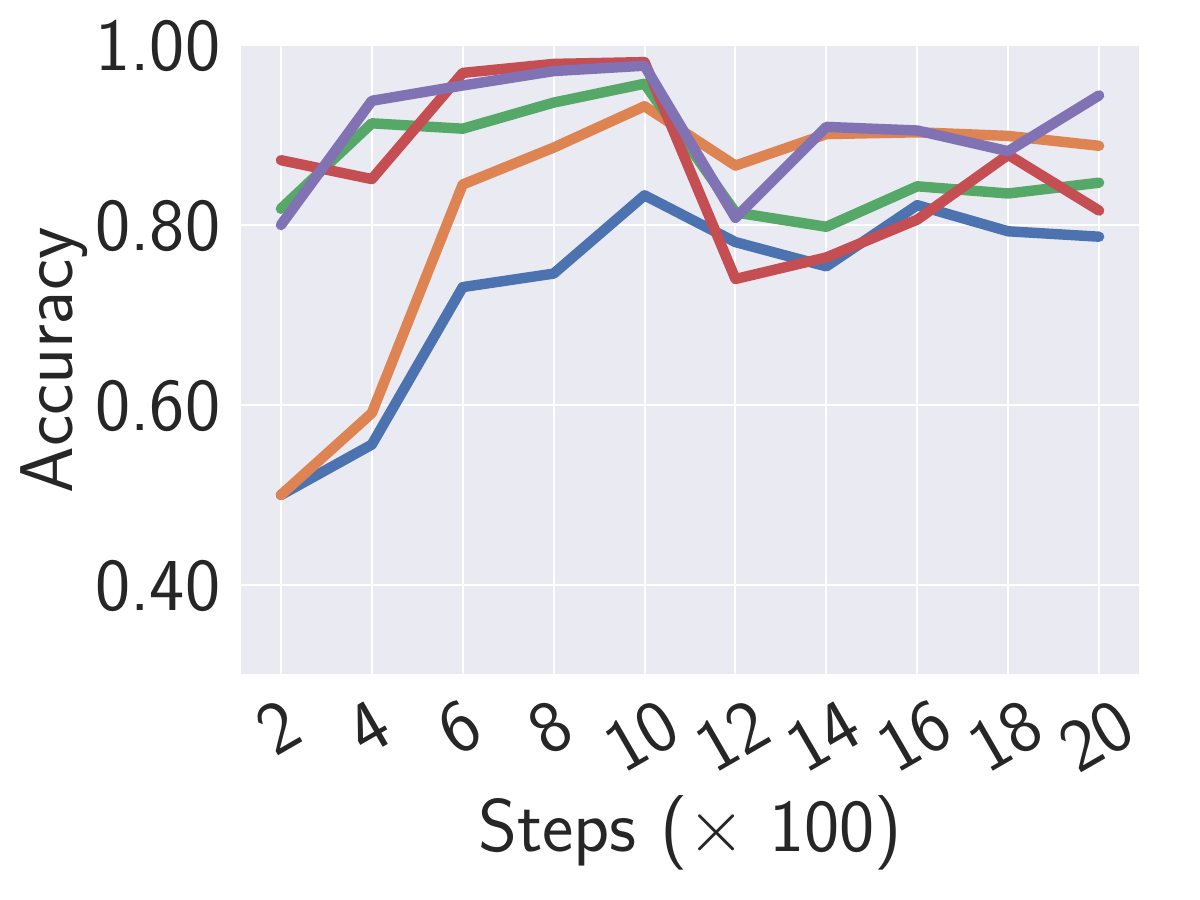}
\caption{HateCheck}
\label{figure:ablation_fewer_steps_HateCheck_test_acc}
\end{subfigure}
\begin{subfigure}{0.18\linewidth}
\includegraphics[width=\columnwidth]{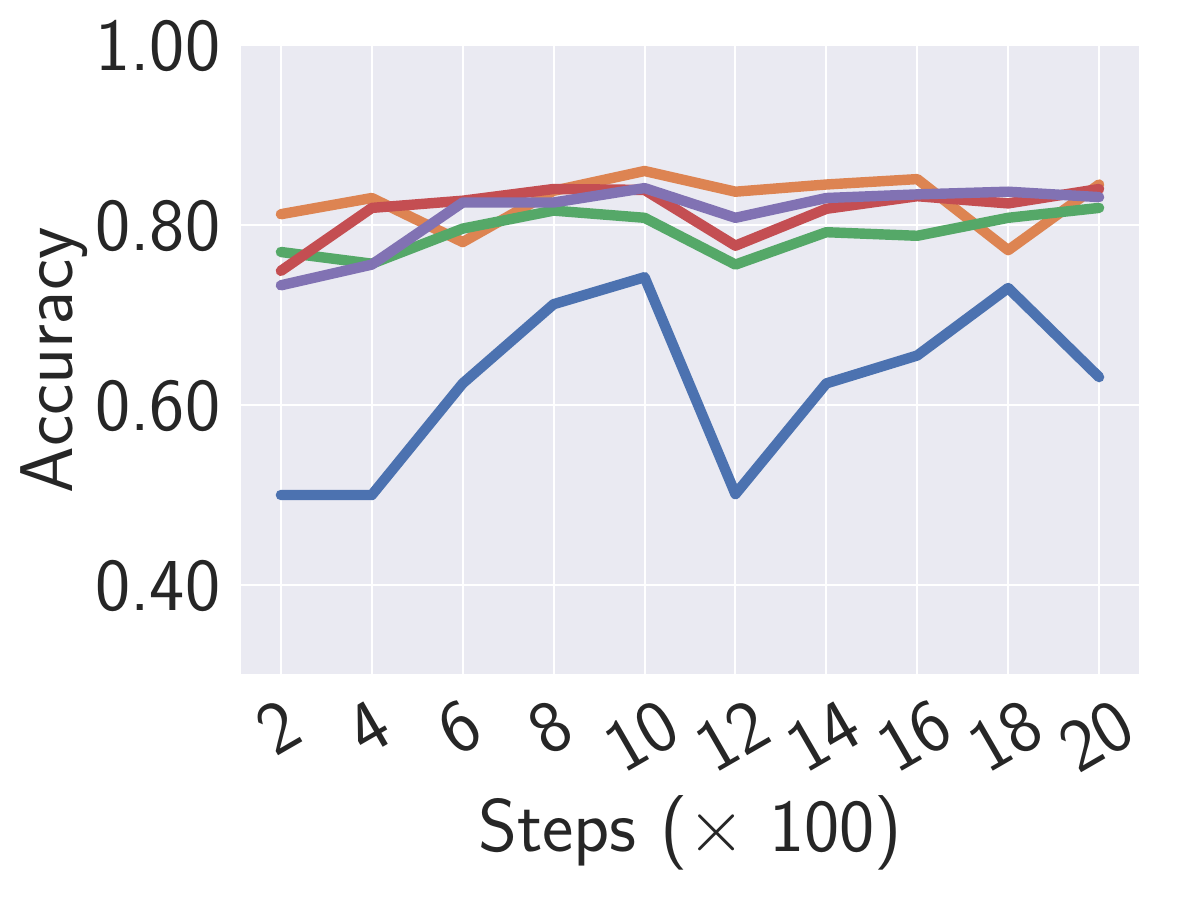}
\caption{SBIC}
\label{figure:ablation_fewer_steps_SBIC_test_acc}
\end{subfigure}
\begin{subfigure}{0.18\linewidth}
\includegraphics[width=\columnwidth]{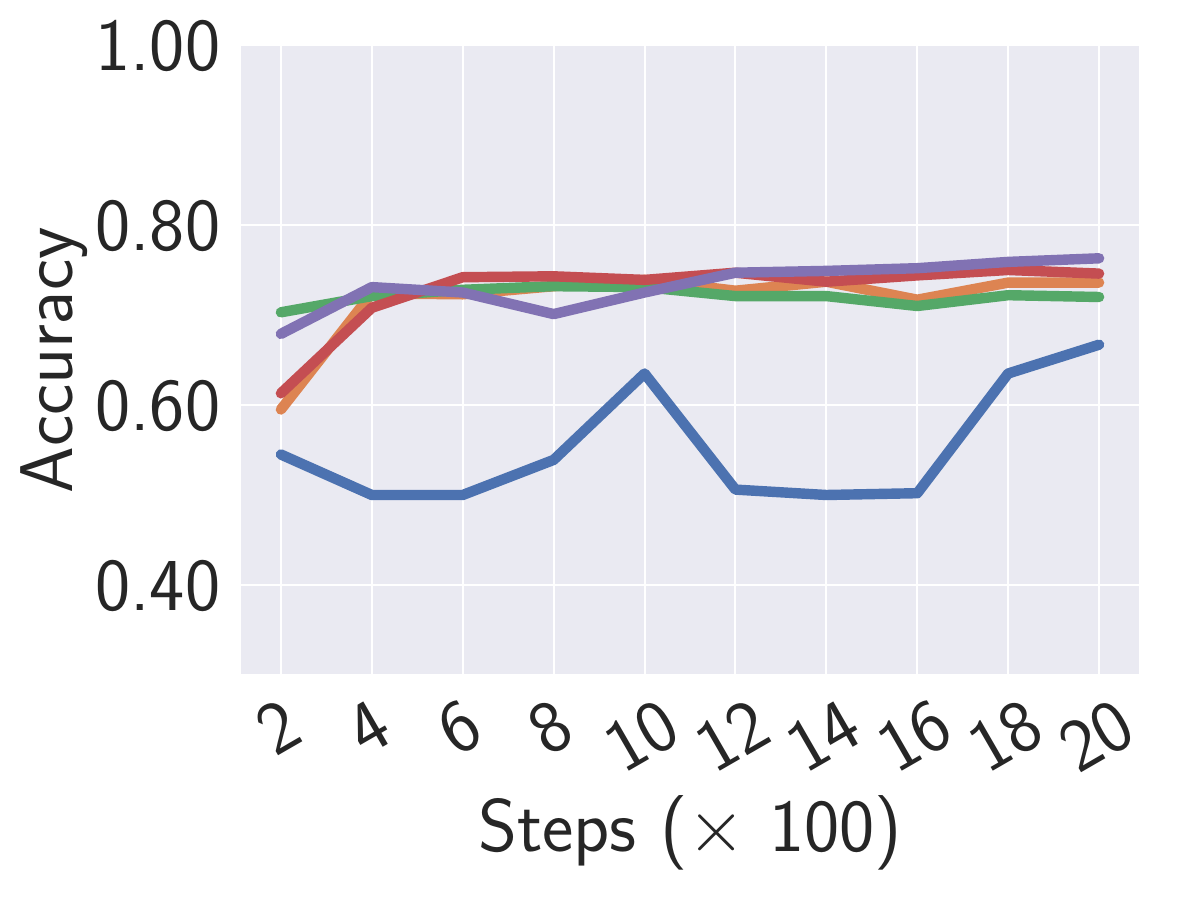}
\caption{MHS}
\label{figure:ablation_fewer_steps_MHS_test_acc}
\end{subfigure}
\caption{Accuracy of Task 1 with different training steps.}
\label{figure:task1_fewer_steps_test_acc}
\end{figure*}

\begin{figure*}[!ht]
\centering
\begin{subfigure}{0.18\linewidth}
\includegraphics[width=\columnwidth]{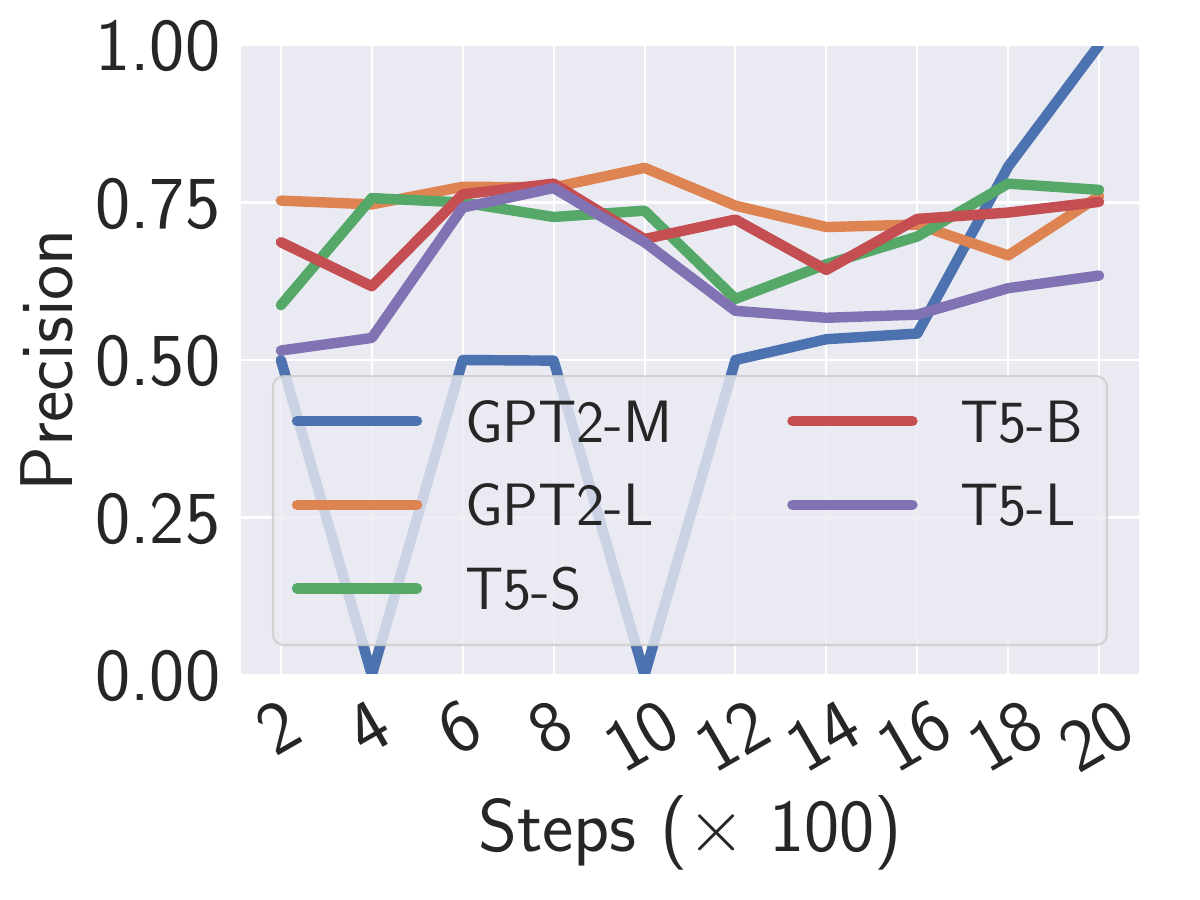}
\caption{HateXplain}
\label{figure:ablation_fewer_steps_HateXplain_precision}
\end{subfigure}
\begin{subfigure}{0.18\linewidth}
\includegraphics[width=\columnwidth]{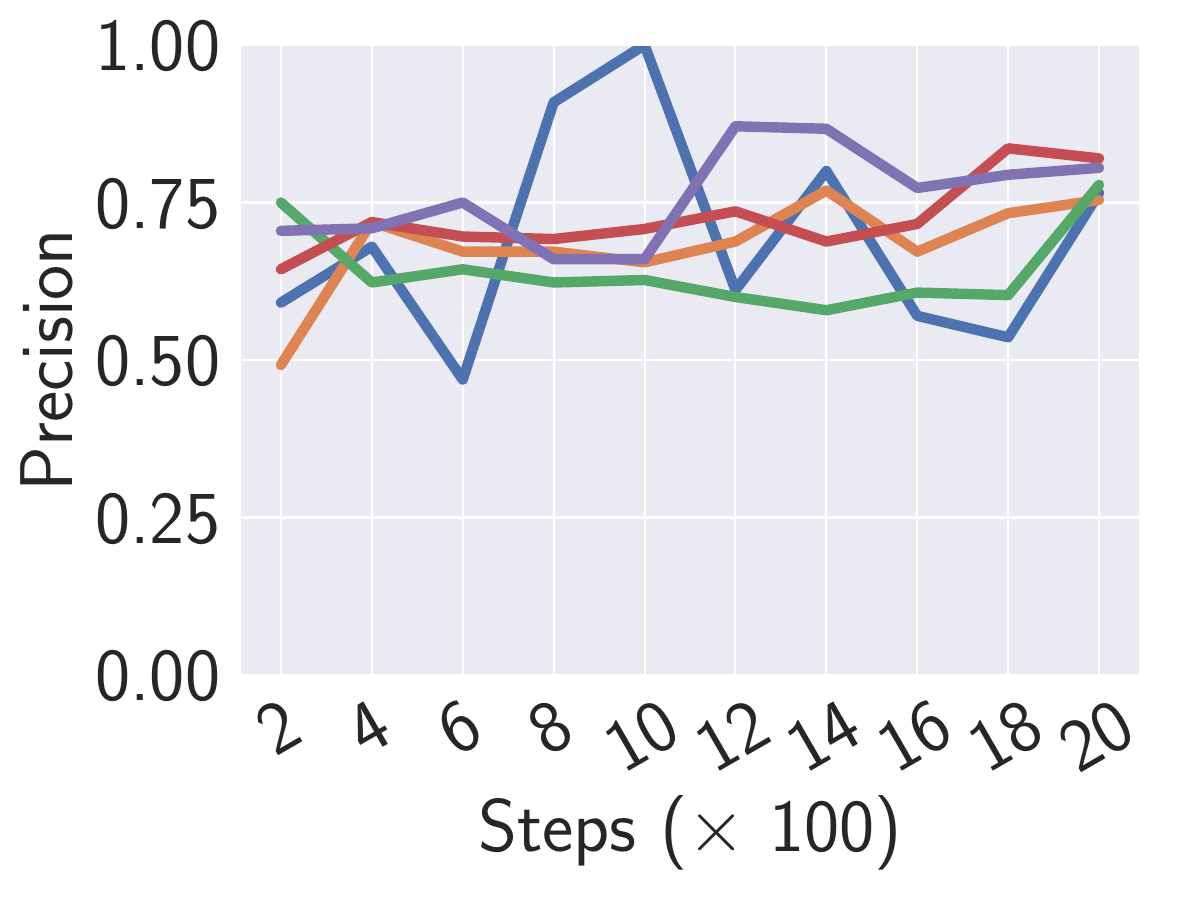}
\caption{USElectionHate20}
\label{figure:ablation_fewer_steps_USElectionHate20_precision}
\end{subfigure}
\begin{subfigure}{0.18\linewidth}
\includegraphics[width=\columnwidth]{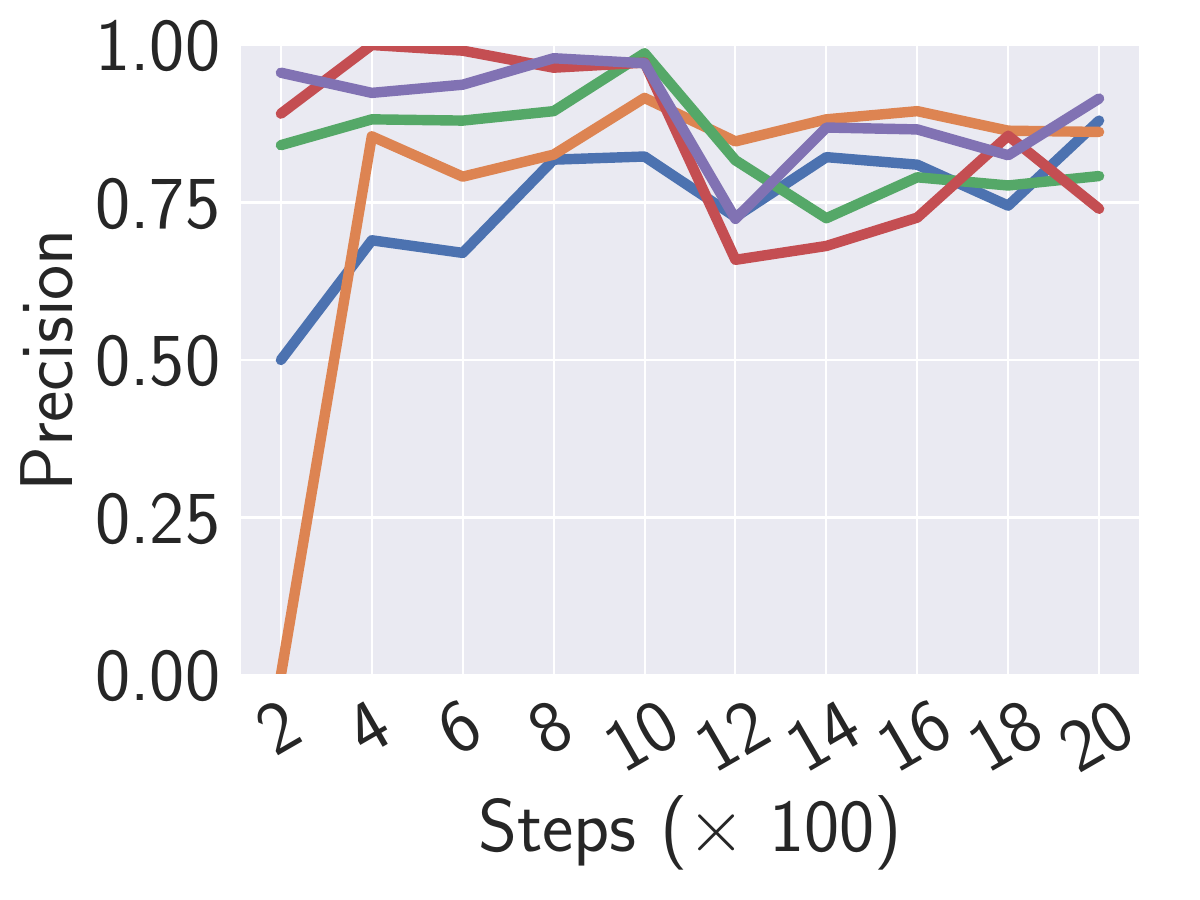}
\caption{HateCheck}
\label{figure:ablation_fewer_steps_HateCheck_precision}
\end{subfigure}
\begin{subfigure}{0.18\linewidth}
\includegraphics[width=\columnwidth]{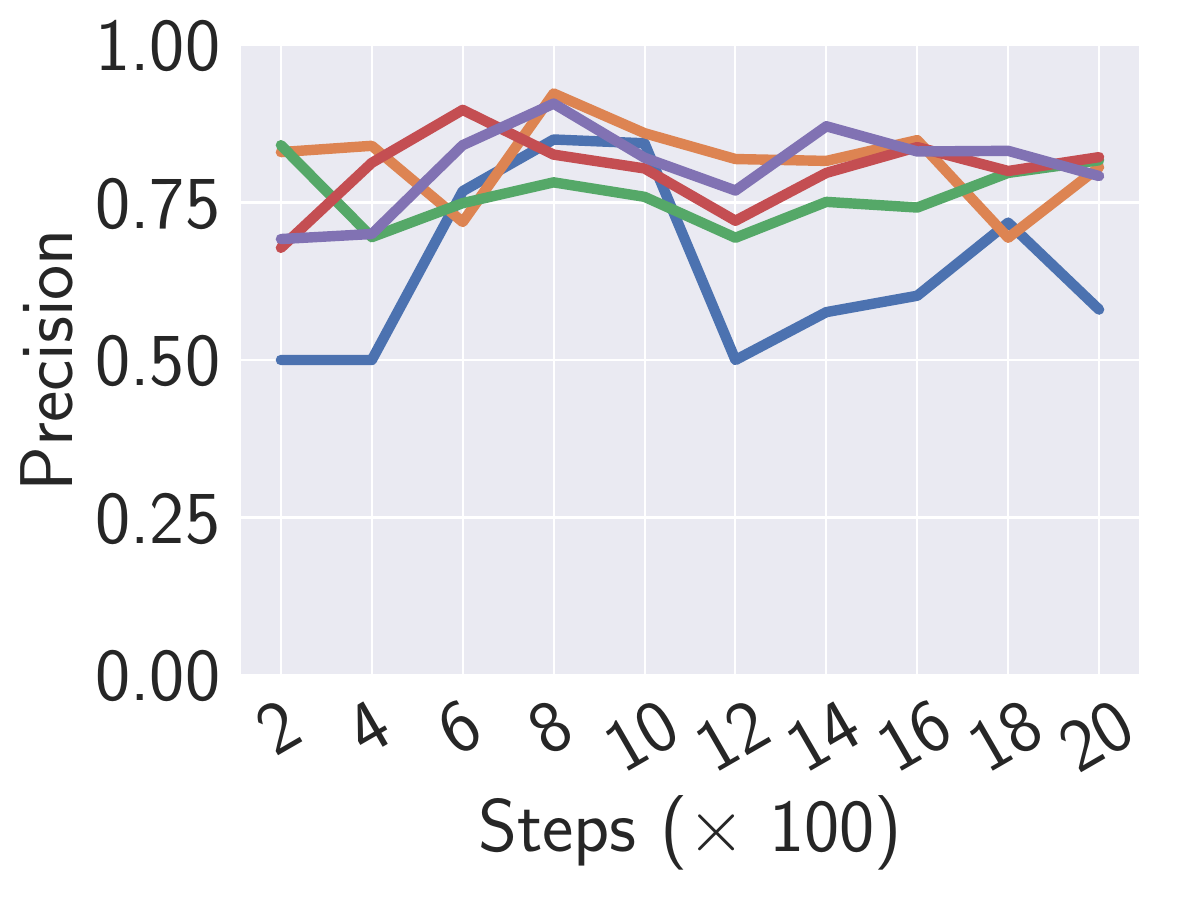}
\caption{SBIC}
\label{figure:ablation_fewer_steps_SBIC_precision}
\end{subfigure}
\begin{subfigure}{0.18\linewidth}
\includegraphics[width=\columnwidth]{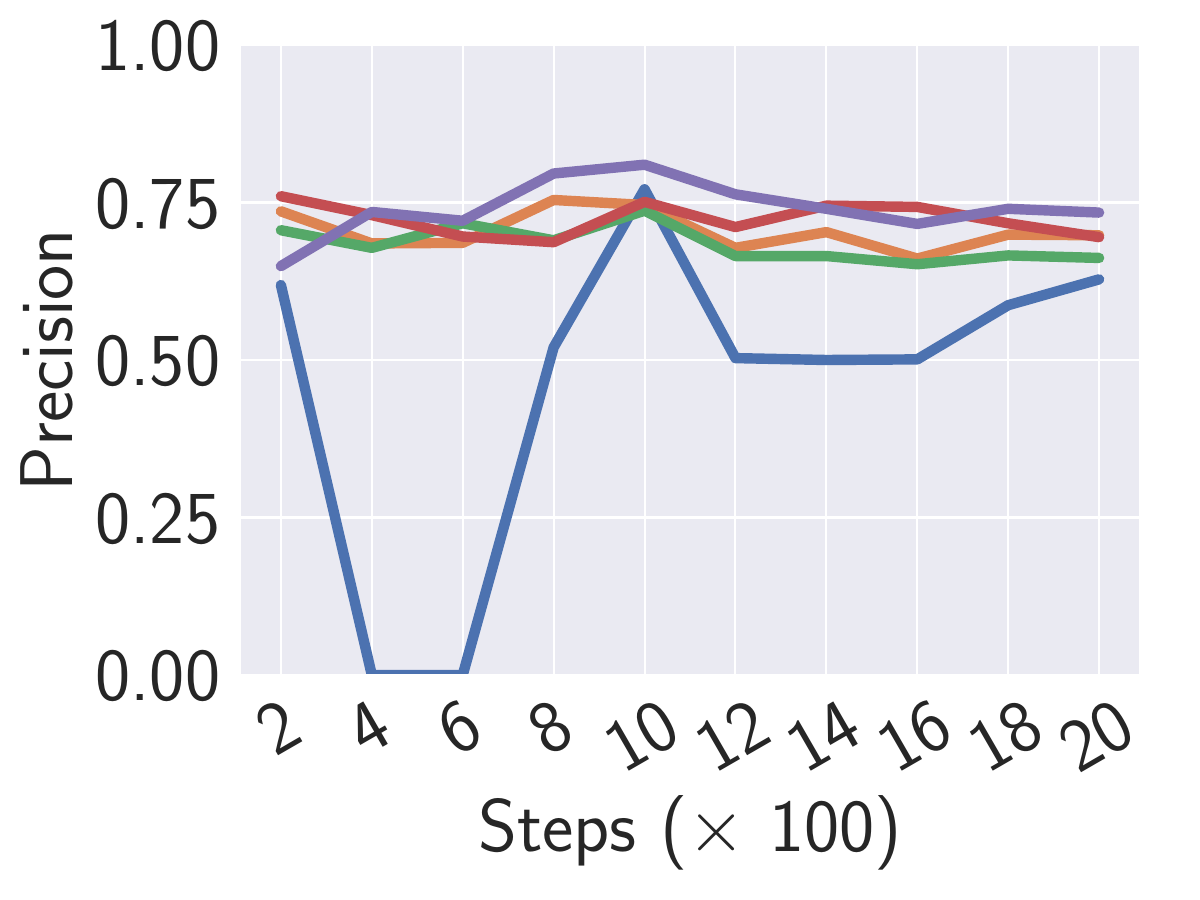}
\caption{MHS}
\label{figure:ablation_fewer_steps_MHS_precision}
\end{subfigure}
\caption{Precision of Task 1 with different training steps.}
\label{figure:task1_fewer_steps_precision}
\end{figure*}

\begin{figure*}[!ht]
\centering
\begin{subfigure}{0.18\linewidth}
\includegraphics[width=\columnwidth]{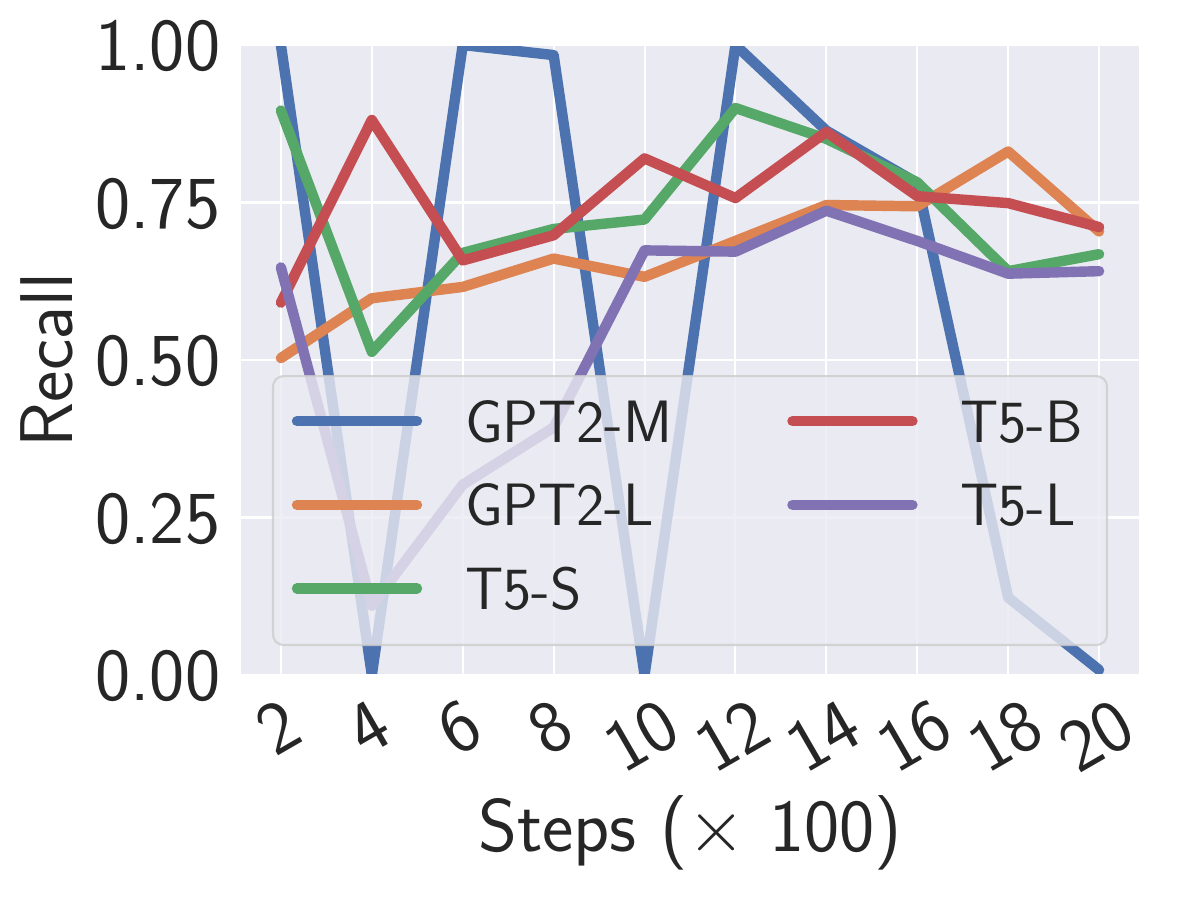}
\caption{HateXplain}
\label{figure:ablation_fewer_steps_HateXplain_recall}
\end{subfigure}
\begin{subfigure}{0.18\linewidth}
\includegraphics[width=\columnwidth]{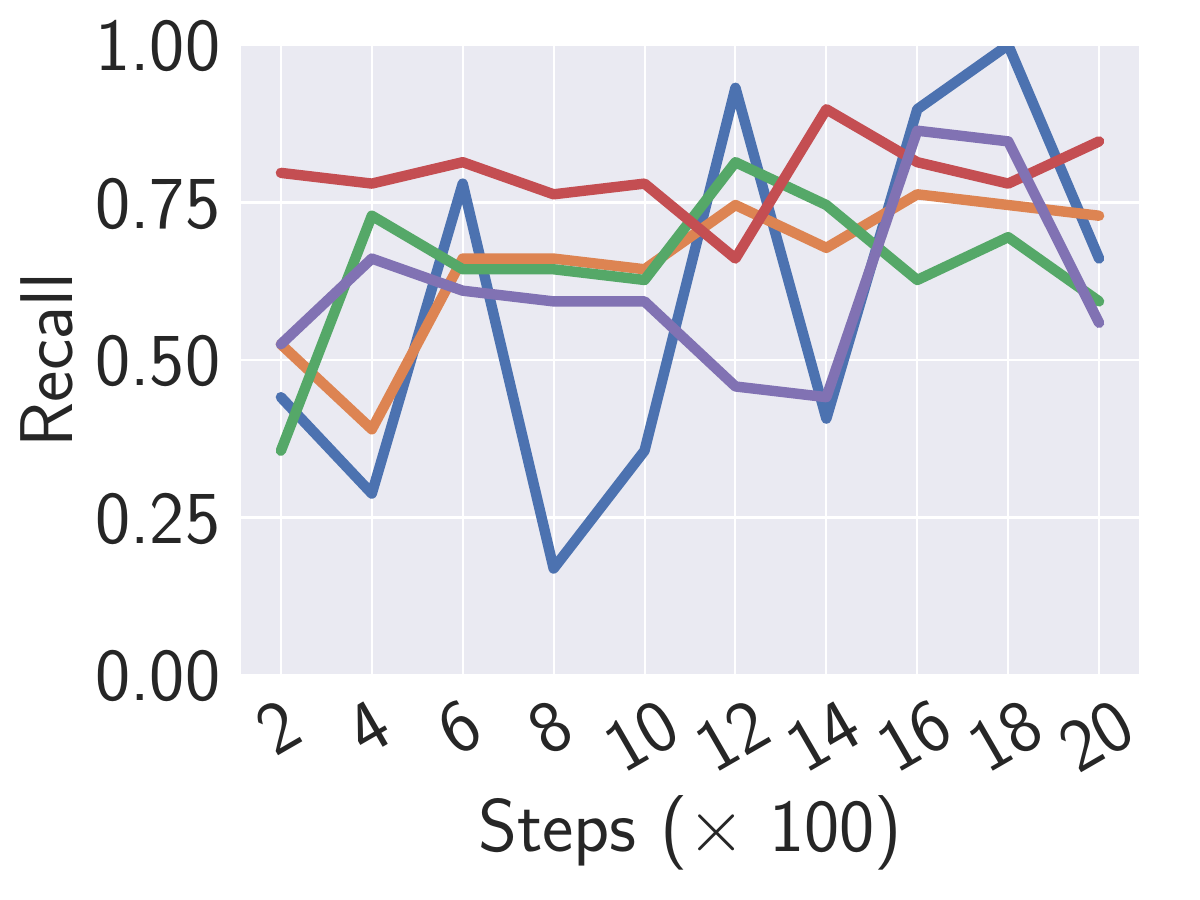}
\caption{USElectionHate20}
\label{figure:ablation_fewer_steps_USElectionHate20_recall}
\end{subfigure}
\begin{subfigure}{0.18\linewidth}
\includegraphics[width=\columnwidth]{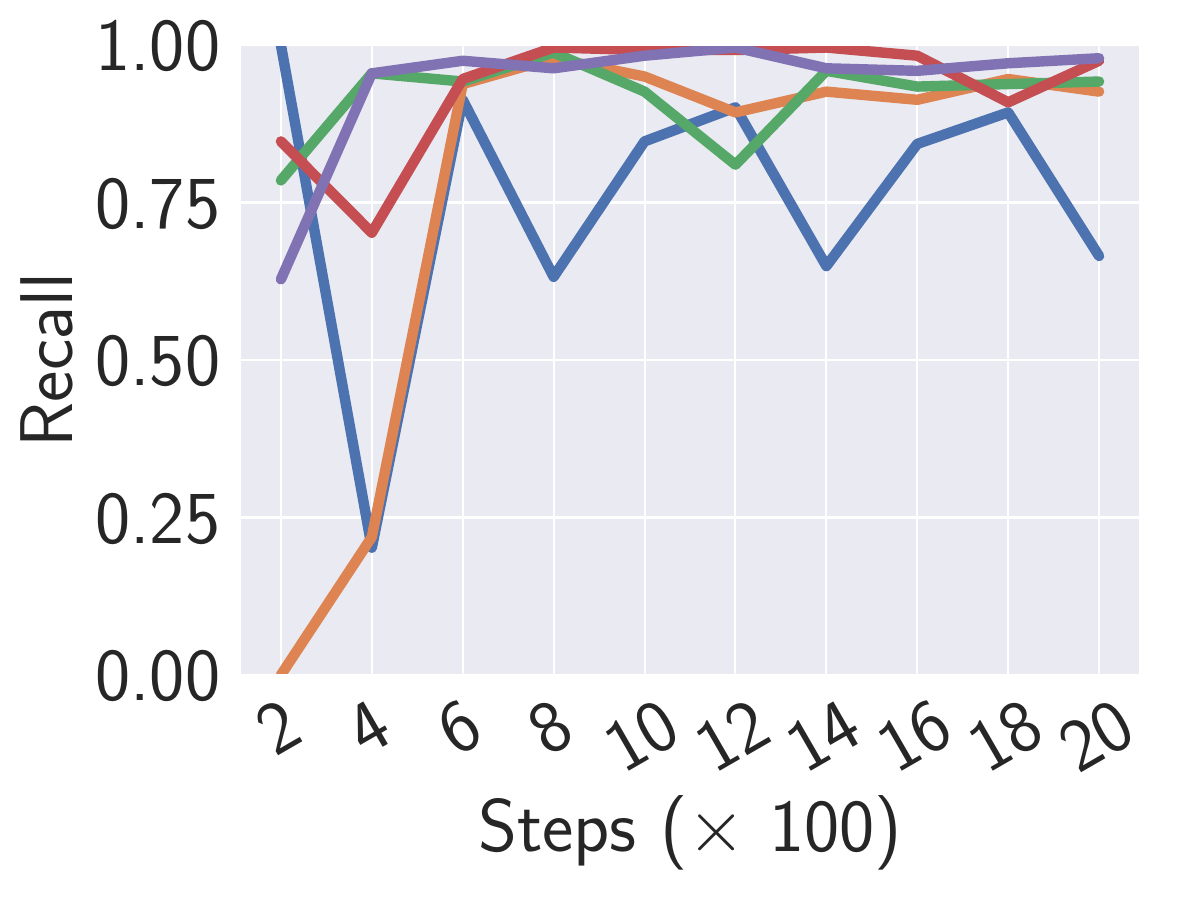}
\caption{HateCheck}
\label{figure:ablation_fewer_steps_HateCheck_recall}
\end{subfigure}
\begin{subfigure}{0.18\linewidth}
\includegraphics[width=\columnwidth]{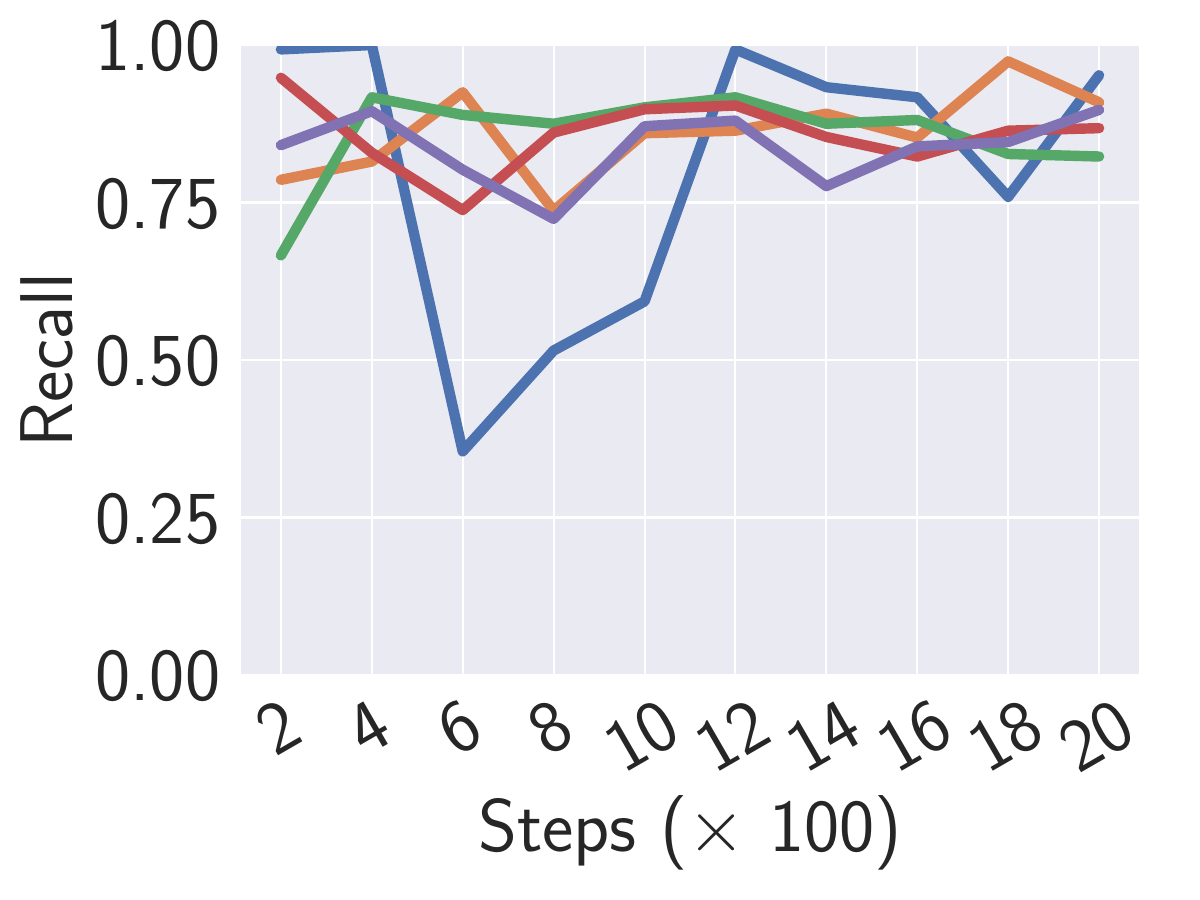}
\caption{SBIC}
\label{figure:ablation_fewer_steps_SBIC_recall}
\end{subfigure}
\begin{subfigure}{0.18\linewidth}
\includegraphics[width=\columnwidth]{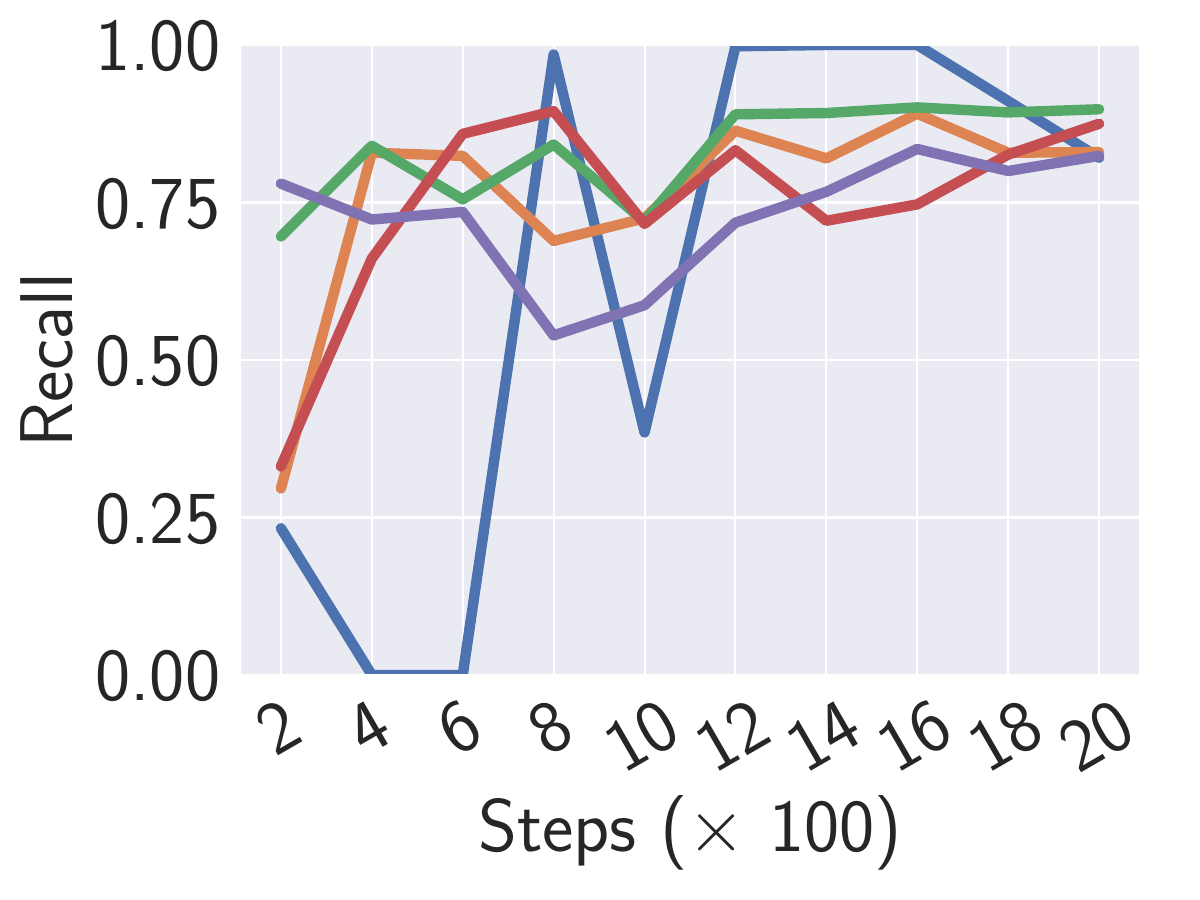}
\caption{MHS}
\label{figure:ablation_fewer_steps_MHS_recall}
\end{subfigure}
\caption{Recall of Task 1 with different training steps.}
\label{figure:task1_fewer_steps_recall}
\end{figure*}

\mypara{Fewer Training Steps}
\autoref{figure:task1_fewer_steps_test_acc}, \autoref{figure:task1_fewer_steps_precision}, and \autoref{figure:task1_fewer_steps_recall} show the accuracy, precision, and recall of Task 1 with fewer training steps

\mypara{Fewer Training Samples}
\autoref{table:task1_ablation_500_training_samples_accuracy_pr} shows the accuracy, precision, and recall of Task 1 with fewer training samples.

\mypara{Prompt Transferability}
\autoref{table:task1_performance_transfer_t5-base_accuracy}, \autoref{table:task1_performance_transfer_t5-base_precision}, and \autoref{table:task1_performance_transfer_t5-base_recall} shows the accuracy, precision, and recall of Task 1 when the training dataset is different from the transfer dataset.
Here the model architecture is T5-B.

\begin{table*}[!ht]
\centering
\caption{Accuracy, precision, and recall of Task 1 with manual prompt.}
\tabcolsep 1.8pt
\scalebox{0.9}{
\begin{tabular}{l | ccccc | ccccc | ccccc}
\toprule
{\textbf{Dataset}} & \multicolumn{5}{c}{\textbf{Accuracy}} & \multicolumn{5}{c}{\textbf{Precision}} & \multicolumn{5}{c}{\textbf{Recall}} \\
 & \textbf{GPT2-M} & \textbf{GPT2-L} & \textbf{T5-S} & \textbf{T5-B} & \textbf{T5-L} & \textbf{GPT2-M} & \textbf{GPT2-L} & \textbf{T5-S} & \textbf{T5-B} & \textbf{T5-L} & \textbf{GPT2-M} & \textbf{GPT2-L} & \textbf{T5-S} & \textbf{T5-B} & \textbf{T5-L} \\
\midrule
\textbf{HateXplain} & 0.482 & 0.475 & 0.491 & 0.485 & 0.504 & 0.356 & 0.362 & 0.351 & 0.464 & 0.502 & 0.045 & 0.066 & 0.022 & 0.196 & 0.994 \\
\textbf{USElectionHate20} & 0.449 & 0.508 & 0.500 & 0.500 & 0.517 & 0.125 & 0.524 & 0.500 & 0.500 & 0.510 & 0.017 & 0.186 & 0.136 & 0.203 & 0.831\\
\textbf{HateCheck} & 0.510 & 0.506 & 0.508 & 0.498 & 0.587 & 0.778 & 0.800 & 0.526 & 0.333 & 0.595 & 0.029 & 0.017 & 0.169 & 0.004 & 0.545 \\
\textbf{SBIC.v2} & 0.533 & 0.516 & 0.477 & 0.454 & 0.530 & 0.630 & 0.618 & 0.370 & 0.265 & 0.521 & 0.160 & 0.086 & 0.065 & 0.052 & 0.764\\
\textbf{MHS} & 0.470 & 0.488 & 0.452 & 0.512 & 0.561 & 0.366 & 0.442 & 0.372 & 0.562 & 0.538 & 0.082 & 0.095 & 0.141 & 0.113 & 0.867 \\
\midrule
\textbf{Avg.} & 0.489 & 0.499 & 0.486 & 0.490 & 0.540 & 0.451 & 0.549 & 0.424 & 0.425 & 0.533 & 0.067 & 0.090 & 0.107 & 0.114 & 0.800\\
\bottomrule
\end{tabular}
}
\label{table:task1_ablation_manual_prompt_accuracy_pr}
\end{table*}

\begin{table*}[!ht]
\centering
\caption{Accuracy, precision, and recall of Task 1 with 500 training samples on each dataset.}
\scalebox{0.9}{
\begin{tabular}{l | ccc | ccc | ccc}
\toprule
{\textbf{Dataset}} & \multicolumn{3}{c}{\textbf{Accuracy}} & \multicolumn{3}{c}{\textbf{Precision}} & \multicolumn{3}{c}{\textbf{Recall}} \\
& \textbf{T5-S} & \textbf{T5-B} & \textbf{T5-L}  & \textbf{T5-S} & \textbf{T5-B} & \textbf{T5-L}  & \textbf{T5-S} & \textbf{T5-B} & \textbf{T5-L} \\
\midrule
\textbf{HateXplain} & 0.633 & 0.627 & 0.664 & 0.641 & 0.603 & 0.672 & 0.608 & 0.744 & 0.638 \\
\textbf{HateCheck} & 0.862 & 0.890 & 0.587 & 0.843 & 0.849 & 0.562 & 0.888 & 0.950 & 0.781\\
\textbf{SBIC.v2} & 0.757 & 0.770 & 0.719 & 0.728 & 0.741 & 0.659 & 0.823 & 0.829 & 0.908 \\
\textbf{MHS} & 0.650 & 0.678 & 0.668 & 0.643 & 0.661 & 0.694 & 0.676 & 0.731 & 0.602 \\
\midrule
\textbf{Avg.} & 0.726 & 0.741 & 0.659 & 0.714 & 0.714 & 0.647 & 0.749 & 0.813 & 0.732 \\
\bottomrule
\end{tabular}
}
\label{table:task1_ablation_500_training_samples_accuracy_pr}
\end{table*}

\section{Statistical Test}
\label{appendix:statistical_test}

To investigate whether the performance difference is significant, we perform the paired $t$-test on the predictions of the best baseline and the best prompt tuning model in Task 1.
The results are shown in \autoref{table:task1_statistical_test}.
We observe that, on all datasets, the $p$-value is less than 0.01, which indicates that the performance is indeed significantly different.
On Task 2, we compare the performance between SPAN-BERT and PT (T5-L) with the Mann-Whitney U test.
The $p$-value (0.936) indicates that the performance of SPAN-BERT and PT (T5-L) are similar.
Note that we do not perform the statistical test on Task 3 as it has multiple metrics to evaluate the performance.

\begin{table}[ht]
\centering
\caption{Paired $t$-test on Task 1 performance.}
\tabcolsep 1.5pt
\scalebox{0.9}{
\begin{tabular}{l | cccc}
\toprule
{\textbf{Dataset}} & \textbf{Baseline ($F_1$)} & \textbf{Prompt Tuning ($F_1$)} & \textbf{$p$-value} \\
\midrule
\textbf{HateXplain} & 0.703 & 0.731 & 2.33e-51 \\
\textbf{USElectionHate20} & 0.506 & 0.833 & 1.42e-9 \\
\textbf{HateCheck} & 0.784 & 0.946 & 8.07e-8 \\
\textbf{SBIC.v2} & 0.669 & 0.854 & 0.00 \\
\textbf{MHS} & 0.790 & 0.776 & 7.93e-50 \\
\bottomrule
\end{tabular}
}
\label{table:task1_statistical_test}
\end{table}

\section{Task 1 Performance with Dynamic Threshold}
\label{appendix:dynamic_threshold}

The performance with the dynamic threshold (rather than 0.5) for Perspective API is shown in \autoref{table:dynamic_threshold}. 
We observe that prompt tuning still outperforms Perspective API in most of the cases.

\begin{table}[ht]
\centering
\caption{Task 1 performance with dynamic threshold.}
\tabcolsep 0.5pt
\scalebox{0.9}{
\begin{tabular}{l | cccc}
\toprule
{\textbf{Dataset}} & \textbf{Baseline} & \textbf{Baseline (Threshold)} & \textbf{Prompt Tuning} \\
\midrule
\textbf{HateXplain} & 0.703 & 0.714 (0.415) & 0.731 \\
\textbf{USElectionHate20} & 0.506 & 0.762 (0.230) & 0.833 \\
\textbf{HateCheck} & 0.784 & 0.790 (0.445) & 0.946 \\
\textbf{SBIC.v2} & 0.669 & 0.782 (0.167) & 0.854 \\
\textbf{MHS} & 0.790 & 0.790 (0.498) & 0.776 \\
\bottomrule
\end{tabular}
}
\label{table:dynamic_threshold}
\end{table}

\begin{table*}[ht]
\centering
\caption{Accuracy of Task 1. 
The best results of each dataset are highlighted in bold.}
\scalebox{0.9}{
\begin{tabular}{l |ccc|ccccc}
\toprule
\multirow{2}{*}{\textbf{Dataset}}   & \multicolumn{3}{c|}{\textbf{Baselines}} & \multicolumn{5}{c}{\textbf{Prompt Tuning}}  \\
                                    & \textbf{Perspective} & \textbf{ToxicBERT} & \textbf{UnRoBERTa} & \textbf{GPT2-M} & \textbf{GPT2-L} & \textbf{T5-S} & \textbf{T5-B} & \textbf{T5-L} \\
\midrule
\textbf{HateXplain} & 0.668 & 0.628 & 0.625 & 0.504 & 0.741 & 0.734 & \textbf{0.738} & 0.635 \\
\textbf{USElectionHate20} & 0.653 & 0.644 & 0.610 & 0.729 & 0.746 & 0.712 & \textbf{0.831} & 0.712 \\
\textbf{HateCheck} & 0.750 & 0.620 & 0.616 & 0.787 & 0.888 & 0.847 & 0.816 & \textbf{0.944} \\
\textbf{SBIC.v2} & 0.696 & 0.635 & 0.635 & 0.631 & \textbf{0.845} & 0.819 & 0.840 & 0.831 \\
\textbf{MHS} & 0.760 & 0.736 & 0.747 & 0.667 & 0.736 & 0.720 & 0.746 & \textbf{0.763} \\
\midrule
\textbf{Avg.} & 0.705 & 0.653 & 0.647 & 0.664 & 0.791 & 0.766 & \textbf{0.794} & 0.777 \\
\bottomrule
\end{tabular}
}
\label{table:task1_performance_acc}
\end{table*}

\begin{table*}[ht]
\centering
\caption{Precision of Task 1.}
\scalebox{0.9}{
\begin{tabular}{l |ccc|ccccc}
\toprule
\multirow{2}{*}{\textbf{Dataset}}   & \multicolumn{3}{c|}{\textbf{Baselines}} & \multicolumn{5}{c}{\textbf{Prompt Tuning}}  \\
                                    & \textbf{Perspective} & \textbf{ToxicBERT} & \textbf{UnRoBERTa} & \textbf{GPT2-M} & \textbf{GPT2-L} & \textbf{T5-S} & \textbf{T5-B} & \textbf{T5-L} \\
\midrule
\textbf{HateXplain} & 0.635 & 0.610 & 0.611 & 1.000 & 0.760 & 0.770 & 0.751 & 0.634 \\
\textbf{USElectionHate20} & 0.875 & 0.870 & 0.810 & 0.765 & 0.754 & 0.778 & 0.820 & 0.805 \\
\textbf{HateCheck} & 0.691 & 0.592 & 0.586 & 0.880 & 0.862 & 0.792 & 0.740 & 0.915 \\
\textbf{SBIC.v2} & 0.734 & 0.682 & 0.683 & 0.580 & 0.806 & 0.817 & 0.822 & 0.792 \\
\textbf{MHS} & 0.703 & 0.686 & 0.697 & 0.628 & 0.698 & 0.662 & 0.695 & 0.734 \\
\midrule
\textbf{Avg.} & 0.728 & 0.688 & 0.677 & 0.771 & 0.776 & 0.764 & 0.766 & 0.776 \\
\bottomrule
\end{tabular}
}
\label{table:task1_performance_precision}
\end{table*}

\begin{table*}[ht]
\centering
\caption{Recall of Task 1.}
\scalebox{0.9}{
\begin{tabular}{l |ccc|ccccc}
\toprule
\multirow{2}{*}{\textbf{Dataset}}   & \multicolumn{3}{c|}{\textbf{Baselines}} & \multicolumn{5}{c}{\textbf{Prompt Tuning}}  \\
                                    & \textbf{Perspective} & \textbf{ToxicBERT} & \textbf{UnRoBERTa} & \textbf{GPT2-M} & \textbf{GPT2-L} & \textbf{T5-S} & \textbf{T5-B} & \textbf{T5-L} \\
\midrule
\textbf{HateXplain} & 0.787 & 0.712 & 0.690 & 0.008 & 0.704 & 0.668 & 0.711 & 0.641 \\
\textbf{USElectionHate20} & 0.356 & 0.339 & 0.288 & 0.661 & 0.729 & 0.593 & 0.847 & 0.559 \\
\textbf{HateCheck} & 0.905 & 0.773 & 0.785 & 0.665 & 0.926 & 0.942 & 0.975 & 0.979 \\
\textbf{SBIC.v2} & 0.614 & 0.506 & 0.505 & 0.952 & 0.909 & 0.823 & 0.868 & 0.897 \\
\textbf{MHS} & 0.902 & 0.873 & 0.872 & 0.821 & 0.830 & 0.898 & 0.875 & 0.824 \\
\midrule
\textbf{Avg.} & 0.713 & 0.641 & 0.628 & 0.621 & 0.820 & 0.785 & 0.855 & 0.780 \\
\bottomrule
\end{tabular}
}
\label{table:task1_performance_recall}
\end{table*}

\begin{table*}[!ht]
\centering
\caption{Accuracy of Task 1 when the training dataset is different from the transfer dataset.}
\scalebox{0.9}{
\begin{tabular}{l |ccccc}
\toprule
\multirow{2}{*}{\textbf{Training Dataset}}   & \multicolumn{5}{c}{\textbf{Transfer Dataset}} \\
 & \textbf{HateXplain} & \textbf{USElectionHate20} & \textbf{HateCheck} & \textbf{SBIC} & \textbf{MHS} \\
\midrule
\textbf{HateXplain} & - & 0.627 & 0.556 & 0.579 & 0.703 \\
\textbf{USElectionHate20} & 0.629 & - & 0.574 & 0.552 & 0.708 \\
\textbf{HateCheck} & 0.507 & 0.559 & - & 0.603 & 0.552 \\
\textbf{SBIC.v2} & 0.553 & 0.551 & 0.552 & - & 0.588 \\
\textbf{MHS} & 0.654 & 0.669 & 0.603 & 0.585 & - \\
\bottomrule
\end{tabular}
}
\label{table:task1_performance_transfer_t5-base_accuracy}
\end{table*}

\begin{table*}[!ht]
\centering
\caption{Precision of Task 1 when the training dataset is different from the transfer dataset.}
\scalebox{0.9}{
\begin{tabular}{l |ccccc}
\toprule
\multirow{2}{*}{\textbf{Training Dataset}}   & \multicolumn{5}{c}{\textbf{Transfer Dataset}} \\
 & \textbf{HateXplain} & \textbf{USElectionHate20} & \textbf{HateCheck} & \textbf{SBIC} & \textbf{MHS} \\
\midrule
\textbf{HateXplain} & - & 0.778 & 0.634 & 0.675 & 0.726 \\
\textbf{USElectionHate20} & 0.615 & - & 0.622 & 0.571 & 0.674 \\
\textbf{HateCheck} & 0.506 & 0.733 & - & 0.646 & 0.546 \\
\textbf{SBIC.v2} & 0.536 & 0.600 & 0.534 & - & 0.563 \\
\textbf{MHS} & 0.622 & 0.794 & 0.600 & 0.618 & - \\
\bottomrule
\end{tabular}
}
\label{table:task1_performance_transfer_t5-base_precision}
\end{table*}

\begin{table*}[!ht]
\centering
\caption{Recall of Task 1 when the training dataset is different from the transfer dataset.}
\scalebox{0.9}{
\begin{tabular}{l |ccccc}
\toprule
\multirow{2}{*}{\textbf{Training Dataset}}   & \multicolumn{5}{c}{\textbf{Transfer Dataset}} \\
 & \textbf{HateXplain} & \textbf{USElectionHate20} & \textbf{HateCheck} & \textbf{SBIC} & \textbf{MHS} \\
\midrule
\textbf{HateXplain} & - & 0.356 & 0.264 & 0.304 & 0.654 \\
\textbf{USElectionHate20} & 0.689 & - & 0.380 & 0.422 & 0.803 \\
\textbf{HateCheck} & 0.586 & 0.186 & - & 0.456 & 0.616 \\
\textbf{SBIC.v2} & 0.786 & 0.305 & 0.818 & - & 0.783 \\
\textbf{MHS} & 0.785 & 0.458 & 0.620 & 0.446 & - \\
\bottomrule
\end{tabular}
}
\label{table:task1_performance_transfer_t5-base_recall}
\end{table*}

\end{document}